\documentclass[11pt,letterpaper]{article}

\usepackage{latexsym,multirow}
\usepackage{amssymb,amsmath, bm}
\usepackage{bigints}
\usepackage{bbm} 
\usepackage{graphicx,subfigure}
\usepackage[color,all,import,arrow]{xy}
\usepackage{enumerate}
\usepackage{verbatim}
\usepackage{mathtools}

\usepackage{placeins}

\usepackage{natbib}

\usepackage[pdftex, bookmarksopen=true, bookmarksnumbered=true,
pdfstartview=FitH, breaklinks=true, urlbordercolor={0 1 0}, citebordercolor={0 0 1}]{hyperref}
\usepackage{colortbl}
\usepackage{subfigure}

\usepackage{dcolumn}
\newcolumntype{.}{D{.}{.}{-1}}
\newcolumntype{d}[1]{D{.}{.}{#1}}

\usepackage{theorem}
\theoremstyle{plain}
\theoremheaderfont{\scshape}

\usepackage{caption}
\usepackage[capposition=top]{floatrow}
\newcommand{\mynote}[1]{{\floatfoot{\textit{Notes:} #1}}}

\usepackage{longtable}
\usepackage{booktabs}

\usepackage{tikz}
\usepackage{inputenc}
\usetikzlibrary{shapes,arrows,trees,fit,positioning,shapes.misc}
\usetikzlibrary{bayesnet}

\usepackage[ruled,vlined]{algorithm2e}

\newcommand{\iid}{\stackrel{\rm i.i.d.}{\sim}}
\newcommand{\indep}{\stackrel{\rm indep.}{\sim}}

\usepackage{rotating}

\usepackage{arydshln}

\usepackage[compact]{titlesec}

\usepackage{setspace}
\usepackage[bottom]{footmisc}
\usepackage{appendix}

\allowdisplaybreaks

\newcommand\spacingset[1]{\renewcommand{\baselinestretch}%
{#1}\small\normalsize}

\newcommand{\blind}{0}

\newcommand{\Beta}{\textsf{Beta}}

\newcommand{\Bern}{\textsf{Bernoulli}}

\newcommand{\Normal}{\mathcal{N}}
\newcommand{\Gammad}{\textsf{Gamma}}
\newcommand{\logit}{\text{logit}}

\newcommand{\Dir}{\textsf{Dirichlet}}

\newcommand{\Cat}{\textsf{Categorical}}

\newcommand{\wt}{\widetilde}

\newcommand{\E}{\mathbb{E}}
\newcommand{\bX}{\mathbf{X}}
\newcommand{\bx}{\mathbf{x}}
\newcommand{\bs}{\mathbf{s}}

\newcommand{\bP}{\mathbf{P}}

\newcommand{\bw}{\mathbf{w}}

\newcommand{\boldeta}{\bm{\eta}}

\newcommand{\bh}{\bm{h}}

\newcommand{\bz}{\bm{z}}

\newcommand{\cW}{\mathcal{W}}

\newcommand{\cV}{\mathcal{V}}

\newcommand{\btheta}{\boldsymbol{\theta}}
\newcommand{\bbeta}{\boldsymbol{\beta}}
\newcommand{\balpha}{\boldsymbol{\alpha}}

\newcommand{\bSigma}{\boldsymbol{\Sigma}}
\newcommand{\bgamma}{\boldsymbol{\gamma}}
\newcommand{\blambda}{\boldsymbol{\lambda}}
\newcommand{\bphi}{\boldsymbol{\phi}}

\newcommand{\bpi}{\boldsymbol{\pi}}

\newcommand{\keyATM}{\textsf{keyATM}}

\newcommand{\LDA}{\textsf{LDA}}
\newcommand{\wLDA}{\textsf{wLDA}}
\newcommand{\STM}{\textsf{STM}}
\newcommand{\kprime}{k^{'}}

\newcommand{\zdel}{\mathbf{z}^{- di}}

\begin{document}

\newcommand{\tit}{Keyword Assisted Topic Models}
%

\spacingset{1}

\if0\blind

{\title{\bf\tit\thanks{The proposed methodology is implemented via an
      open-source software package \keyATM, which is available at
      \url{https://cran.r-project.org/package=keyATM}.  We thank Doug
      Rice and Yutaka Shinada for sharing their data, Luwei Ying,
      Jacob Montgomery, Brandon Stewart for sharing us their experience
       of setting up validation exercises, and Soichiro Yamauchi for
      advice on methodological and computational issues.
      We also thank Soubhik Barari, Matthew Blackwell, Max
      Goplerud, Andy Halterman, Masataka Harada, Hiroto Katsumata,
      Gary King, Dean Knox, Shiro Kuriwaki, Will Lowe, Luke Miratrix,
      Hirofumi Miwa, Daichi Mochihashi, Santiago Olivella, Yon Soo
      Park, Reed Rasband, Hunter Rendleman, Sarah Mohamed, Yuki
      Shiraito, Tyler Simko, and Diana Stanescu, as well as seminar
      participants at the Institute for Quantitative Social Science
      Applied Statistics Workshop, the Japanese Society for
      Quantitative Political Science 2020 Winter Meeting,
      International Methods Colloquium, Annual Conference of the
      Society for Political Methodology (2020), and Annual Conference
      of the American Political Science Association (2020) for helpful
      discussions and comments on this project. Lastly, we thank the editors
      and our three anonymous reviewers for providing us with additional
      comments.}}

  \author{Shusei Eshima\thanks{Graduate Student, Department of
      Government, Harvard University. 1737 Cambridge Street, Institute
      for Quantitative Social Science, Cambridge MA 02138. Email:
      \href{mailto:shuseieshima@g.harvard.edu}{shuseieshima@g.harvard.edu}
      URL:
      \href{https://shusei-e.github.io/}{https://shusei-e.github.io/}
    } \and Kosuke Imai\thanks{Professor, Department of
      Government and Department of Statistics, Harvard University.
      1737 Cambridge Street, Institute for Quantitative Social
      Science, Cambridge MA 02138.  Email:
      \href{mailto:imai@harvard.edu}{imai@harvard.edu} URL:
      \href{https://imai.fas.harvard.edu}{https://imai.fas.harvard.edu}}
    \and Tomoya Sasaki\thanks{Graduate Student, Department of
      Political Science, Massachusetts Institute of Technology. 77
      Massachusetts Avenue E53, Cambridge MA 02142. Email:
      \href{mailto:tomoyas@mit.edu}{tomoyas@mit.edu}
      URL: \href{https://tomoya-sasaki.github.io/}{https://tomoya-sasaki.github.io/}}
      }

   \date{
     Forthcoming in \textit{American Journal of Political Science} \\ \vspace{5mm}
     This Draft: \today \\ \vspace{1.2mm}
     First Draft: March 29, 2020
  }
}

\fi

\if1\blind
\title{\bf \tit}
\fi

\maketitle

\pdfbookmark[1]{Title Page}{Title Page}

\thispagestyle{empty}
\setcounter{page}{0}

\clearpage

\pdfbookmark[1]{Abstract}{Abstract}
\thispagestyle{empty}
\setcounter{page}{0}

\begin{abstract}

  In recent years, fully automated content analysis based on
  probabilistic topic models has become popular among social
  scientists because of their scalability.  The unsupervised nature of
  the models makes them suitable for \textit{exploring} topics in a
  corpus without prior knowledge.  However, researchers find that
  these models often fail to \textit{measure} specific concepts of
  substantive interest by inadvertently creating multiple topics with
  similar content and combining distinct themes into a single topic.
  In this paper, we empirically demonstrate that providing a small
  number of keywords can substantially enhance the measurement
  performance of topic models. An important advantage of the proposed
  keyword assisted topic model (\keyATM) is that the specification of
  keywords requires researchers to label topics prior to fitting a
  model to the data. This contrasts with a widespread practice of
  post-hoc topic interpretation and adjustments that compromises the
  objectivity of empirical findings. In our application, we find that
  \keyATM{} provides more interpretable results, has better document
  classification performance, and is less sensitive to the number of
  topics than the standard topic models.  Finally, we show that
  \keyATM{} can also incorporate covariates and model time trends. An
  open-source software package is available for implementing the
  proposed methodology. 

\vspace{.20in}
\noindent {\bf Verification Materials:}
The data and materials required to verify the computational reproducibility of the results,
procedures and analyses in this article are available on
the American Journal of Political Science Dataverse within
the Harvard Dataverse Network, at:\\ https://doi.org/10.7910/DVN/RKNNVL

\vspace{.25in}
\end{abstract}


\clearpage
\spacingset{1.83}

\section{Introduction}
\label{sec:intro}

Textual data represent the most fundamental way of recording and
preserving human communication and activities.  Social scientists have
analyzed texts to measure a variety of substantive concepts such as
political ideology and policy positions
\citep{Laver2003,Otjes2021}.  A typical process would
require researchers to read and manually classify relevant documents
into different categories based on a codebook prepared specifically
for measuring substantive concepts of interest
\citep[e.g.,][]{Bauer2000}.  Given the lack of scalability of this
traditional approach, social scientists are increasingly relying on
fully automated content analysis based on machine learning models
\citep{Grimmer2013}.  In particular, probabilistic topic models have
been widely used to uncover the content of documents and to explore
the relations between discovered topics and meta-information such as
author characteristics
\citep[e.g.,][]{Blei2003,Grimmer2010,Roberts2016}.

Unfortunately, while topic models can \textit{explore} themes of a
corpus \citep[e.g.,][]{Roberts2014}, they do not necessarily
\textit{measure} specific concepts of substantive interest.  Although
researchers have also relied upon topic models for measurement
purposes \citep[e.g., ][]{Grimmer2013ajps,Bagozzi2018,Blaydes2018,%
  Barbera2019,Dietrich2019,Martin2019}, they acknowledge that these
fully automated models often inadvertently create multiple topics with
similar content and combine different themes into a single topic
\citep{Chang2009,Newman2011,Morstatter2016}. These undesirable
features may impede obtaining a clear interpretation of topics and
adequate measurements of substantive concepts.  This mismatch is not
surprising because these models do not directly incorporate the
information about topics of interest.  Researchers would not know
whether a model yields topics that properly measure substantive
concepts until they fit the model. For this reason,
scholars have emphasized the importance of human validation
\citep{Grimmer2013,Ying2021}.

Another unsatisfactory characteristic of the current approaches is
that researchers must interpret and label estimated topics after model
fitting.  This task is crucial especially when topic models are used
for measurement. Researchers must make a post-hoc decision about
drawing a connection between estimated topics and substantive concepts
of interest.  Together with commonly used post-hoc adjustments of
topics such as topic merging or word reweighting \citep{Bischof2012},
this widespread practice can compromise the scientific objectivity of
empirical findings. Finally, the empirical results obtained under
probabilistic topic models are known to be sensitive to the number of
topics \citep{Boyd-Graber2014,Roberts2016Book}.

In this paper, we propose the keyword assisted topic models (\keyATM)
that allow researchers to label topics via the specification of
keywords \textit{prior to} model fitting. This semi-supervised
approach avoids post-hoc interpretation and adjustments of topics
because researchers can use keywords to specify the substantive
concepts to be measured.  Unlike the popular models such as the latent
Dirichlet allocation \citep[\LDA,][]{Blei2003} and the structural
topic model \citep[\STM,][]{Roberts2016}, the \keyATM{} methodology is
not an unsupervised topic model that only uses unlabeled data. Rather,
it is a semi-supervised topic model that combines a small amount of
information with a large amount of unlabeled
data. \label{fn:semisupervised} As opposed to a supervised approach,
\keyATM{} only requires a small number of keywords for each concept of
interest and does not necessitate manual labeling of many documents.
We emphasize that \keyATM{} is most useful when a researcher is
interested in measuring specific topics.  If the goal is to explore
topics, the existing topic models such as LDA might be more
appropriate.

We empirically demonstrate that providing topic models with a small
number of keywords substantially improves their performance and better
serves the purpose of measurement. We assess the performance of topics
models both qualitatively, by examining the most frequent words for
each estimated topic and employing crowd-sourced validation
\citep{Ying2021}, and quantitatively, by comparing the document
classification with human coding and conducting a crowdsource
validation study.

The proposed \keyATM{} methodology builds upon the model originally
introduced by \citet{Jagarlamudi2012} by making the following
improvements.  First, \keyATM{} can have topics with no
keyword. Researchers may not be able to prepare keywords for all
topics that potentially exist in a corpus \citep{King2017}.  Thus, we
leave room for \textit{exploring} a corpus with these additional
no-keyword topics.  Second, researchers can characterize
document-topic distribution with meta-information to study how
estimated topics vary as a function of document-level covariates and
over time.  Third, \keyATM{} is a fully Bayesian approach and
estimates hyper-parameters to improve the model performance
\citep{Wallach2009}.

\paragraph{Related Methods}
Since the pioneering work of \citet{Blei2003}, numerous researchers
have worked on improving topic interpretability \citep{Blei2012}. We
neither claim that \keyATM{} is better than other existing
approaches nor attempt to provide a comprehensive review of this large
literature here.  Instead, we briefly compare \keyATM{} with the
existing closely related models.  Most importantly, the base \keyATM{}
adds the aforementioned improvements to the model of
\citet{Jagarlamudi2012}.  \citet{Li2019} propose a model similar to
the base \keyATM{} under the assumption that each document has a
single keyword topic and some topics with no keyword.  In contrast,
\keyATM{} allows each document to belong to multiple keyword
topics.

Some researchers have also used human inputs to restrict the parameters of
topic models. For example, \citet{Chemudugunta2008b} incorporate
keywords into a topic model.  However, unlike \keyATM, their model
assumes that pre-specified keywords cannot belong to other topics and
each topic with such keywords cannot have other keywords.
\citet{Hu2011} propose a topic model where researchers refine the
discovered topics by iteratively adding constraints such that
certain sets of words are forced to appear together in the same
topic. \keyATM{} does not place such a strict constraint and also avoids
an iterative refinement process.

Other researchers have incorporated substantive knowledge by placing
an informative prior distribution over topic-word distributions.
\citet{Lu2011} modify the values of prior parameters for keywords,
essentially inflating their frequency.  \citet{Fan2019} employ a
similar strategy when constructing informative priors using a
combination of TF-IDF and domain knowledge. While these
  models require researchers to specify the particular values of hyper
  prior parameters that directly control the importance of keywords,
  our approach imposes a restriction on the structure of prior
  distribution so that the models are allowed to learn from the data
  the importance of keywords (Appendix~p.~\pageref{app:keywords-as-prior}
  explains this connection in detail).  \citet{Hansen2013} and
\citet{Wood2017} use external corpus such as Wikipedia to construct
either topic-word distributions or hyper-parameters.

Placing a certain structure on the prior information is a common strategy.
For example, \citet{Newman2011} model the structural
relations among different words using an external data.
\citet{xie2015} incorporate the information about the similarity of
words through a Markov random field, while \citet{Andrzejewski2009}
specify a set of words that have a similar probability within a
certain topic through a Dirichlet Forest prior distribution.


In contrast to these existing approaches, \keyATM{} directly
incorporates a small number of keywords into a topic-word
distribution.  We believe that this simplicity of \keyATM{} is
particularly appealing for social scientists.


\section{The Base \keyATM}
\label{sec:base}

We begin by describing the base \keyATM, which we will extend in
various ways. The base \keyATM{} improves the model of
\citet{Jagarlamudi2012} by allowing some topics to have no keyword and
estimating hyper-parameters for better empirical performance. Our
application demonstrates that this base \keyATM{} yields results
superior to those of \LDA{} qualitatively and quantitatively.

\subsection{Model}
\label{sec:model-base}

Suppose that we have a total of $D$ documents and each document $d$
has $N_d$ words.  These documents contain a total of $V$ unique words.
Let $w_{di}$ represent the $i$th word in document $d$ where
$\cW_d = \{w_{d1}, w_{d2}, \ldots, w_{dN_d}\}$ represents the set of
all words used in document $d$.  We are interested in identifying the
topics that underlie each document.  We consider two types of topics:
topics with keywords, which are of primary interest to researchers and
are referred to as {\it keyword topics}, and topics without keywords,
which we call {\it no-keyword topics}.  Suppose that we have a total
of $K$ topics and the first $\wt{K}$ of them are keyword topics, i.e.,
$\wt{K} \leq K$.  For each keyword topic $k$, researchers provide a
set of $L_k$ keywords, which is denoted by
$\cV_k=\{v_{k1}, v_{k2}, \dots, v_{k L_k}\}$.  Note that the same
keywords may be used for different keyword topics and keywords are a
part of total $V$ unique words.

Our model is based on the following data generation process.
For each word $i$ in document
$d$, we first draw the latent topic variable
$z_{di} \in \{1,2,\ldots,K\}$ from the topic distribution of the
document,
\begin{equation*}
  z_{di} \ \indep \ \Cat(\btheta_d)
\end{equation*}
where $\btheta_d$ is a $K$-dimensional vector of topic probabilities
for document $d$ with $\sum_{k=1}^K \theta_{dk}=1$.  This
document-topic distribution $\btheta_d$ characterizes the relative
proportion of each topic for document $d$.

If the sampled topic is one of the no-keyword topics, then we draw the
word $w_{di}$ from the corresponding word distribution of the topic,
\begin{equation*}
  w_{di} \mid z_{di} = k \ \indep \ \Cat(\bphi_{k}) \quad
  {\rm for} \ k \in  \{\wt{K}+1, \wt{K}+2,\ldots, K\}
\end{equation*}
where $\bphi_{k}$ is a $V$-dimensional vector of word probabilities
for topic $k$ with $\sum_{v=1}^V \phi_{kv} = 1$.  This probability
vector represents the relative frequency of each word within topic
$k$.

On the other hand, if the sampled topic has keywords, we draw a
Bernoulli random variable $s_{di}$ with success probability $\pi_k$
for word $i$ in document $d$.  If this variable is equal to 1, then
word $w_{di}$ is drawn from the set of keywords for the topic based on
probability vector $\wt{\bphi}_k$.  In contrast, if $s_{di}$ is equal
to 0, then we sample the word from the standard topic-word
distribution of the topic $\bphi_k$.  Therefore, we have,
\begin{eqnarray*}
  s_{di} \mid z_{di} = k & \indep & \Bern(\pi_{k}) \quad \text{for } k \in \{1, 2, \ldots, \wt{K} \} \\
  w_{di} \mid s_{di}, z_{di} = k  & \indep &  \begin{cases} \Cat(\bphi_{k}) & {\rm if} \ s_{di} = 0 \\
  \Cat(\wt{\bphi}_{k})  & \text{if} \ s_{di} = 1  \end{cases} \quad {\rm for} \
                                k \in  \{1, 2, \ldots, \wt{K}\}
\end{eqnarray*}
where $\pi_k$ represents the probability of sampling from the set of
keywords, and
$\wt{\bphi}_{k}$ is a $V$ dimensional vector of word
probabilities for the set of keywords of topic $k$, i.e.,
$\mathcal{V}_{k}$.
Thus, $L_k$ of $V$ elements in $\wt{\bphi}_k$ have
positive values and the others are $0$.  We use the following prior
distributions,
\begin{eqnarray}
  \pi_k & \iid & \Beta(\gamma_1, \gamma_2) \quad \text{for} \ k =
                   1,2,\ldots,\wt{K} \nonumber\\
  \bphi_k & \iid & \Dir(\bbeta) \quad \text{for} \ k =
  1,2,\ldots,K \nonumber \\ 
  \wt{\bphi}_k & \iid & \Dir(\wt{\bbeta}) \quad \text{for} \ k =
  1,2,\ldots, \wt{K} \label{eq:tildephi_prior}\\
  \btheta_d & \iid & \Dir(\balpha) \quad \text{for} \ d = 1,2,\ldots,D \label{eq:theta_base}\\
   \alpha_k & \indep & \begin{cases} \Gammad(\tilde\eta_1, \tilde\eta_2)  &  \text{for} \ k =
                    1,2,\ldots,\wt{K}  \\
                    \Gammad(\eta_1, \eta_2)  & \text{for} \ k =
                    \wt{K} + 1,\wt{K}+2,\ldots,K\end{cases} \label{eq:alpha_base}
\end{eqnarray}
where with slight abuse of notation we specify the prior
  distribution for $\wt{\bphi}_k$, with the constant hyper-parameter value placed
  only on keyword elements of the topic.

In typical applications, the choice of hyper-parameters matters little
so long as the amount of data is sufficiently large.\footnote{The
  default values are: $\gamma_1 = \gamma_2 = 1$, $\beta = 0.01$,
  $ \tilde{\beta} = 0.1$, $\eta_1 = 2$, $\eta_2 = 1$, and
  $\tilde{\eta}_1 = \tilde{\eta}_2 = 1$. But, these
    values can be adjusted reflecting one's prior knowledge.\label{fn:default_prior}} The only exception is the prior
for $\pi_k$, which controls the influence of keywords.  We use the
uniform prior distribution for $\pi_k$, i.e., $\gamma_1=\gamma_2=1$ as
a non-informative prior, which we find works well across a variety of
applications.

As shown above, \keyATM{} is based on a mixture of two distributions,
one with positive probabilities only for keywords and the other with
positive probabilities for all words.  This mixture structure makes
the prior means for the frequency of user-selected keywords given a
topic greater than those of non-keywords in the same topic.  In
addition, the prior variance is also larger for the frequency of
keywords given a topic than for non-keywords.  This encourages
\keyATM{} to place greater importance on keywords {\it a priori} while
allowing the model to learn from the data about the precise degree to
which keywords matter for a given topic.  Because keyword topics are
distinct, \keyATM{} is less prone to the label switching problem.

\subsection{Sampling Algorithm}
\label{sec:sampling-base}

We next describe the sampling algorithm.  To improve the empirical
performance of the model, we use term weights that help prevent highly
frequent words from dominating the resulting topics \citep[][hereafter
\wLDA{}]{Wilson2010}.  We use a collapsed Gibbs sampling algorithm to
sample from the posterior distribution by integrating out the
variables $(\btheta,\bphi, \wt{\bphi}, \bpi)$ \citep{Griffiths2004}.
This yields a Markov chain of $(\bz, \bs, \balpha)$ and helps
address the identifiability problem regarding $\bphi_{kv}$,
$\wt{\bphi}_{kv}$, and $\pi_k$.

From the expression of the collapsed posterior distribution,
it is straightforward to derive the
conditional posterior distribution of each parameter.  First, the
sampling distribution of topic assignment for each word $i$ in
document $d$ is given by,
\begin{align}
  & \Pr(z_{di}=k \mid \zdel, \bw, \bs, \balpha, \bbeta,
    \wt{\bbeta}, \bgamma) \nonumber\\
\propto &\quad
\begin{cases} %
  \dfrac{ \beta_v + n_{k v}^{- di} }{   \sum_v \beta_v +  n_{k}^{- di}} \cdot %
  \dfrac{ n^{- di}_{k} + \gamma_2 }{ \tilde{n}_{k}^{- di} + \gamma_1 + n^{- di}_{k} + \gamma_2 } \cdot %
\left(n_{d{k}}^{- di} + \alpha_{k}  \right)  & \ {\rm if \ } s_{di} = 0, \\
\dfrac{ \tilde{\beta}_v + \tilde{n}_{k v}^{- di}    }{ \sum_{v \in \cV_k} \tilde{\beta}_v + \tilde{n}_{k}^{- di}  } \cdot%
\dfrac{ \tilde{n}^{ - di}_{k} + \gamma_1 }{ \tilde{n}^{- di}_{k} + \gamma_1 + n^{- di}_{k} + \gamma_2 } \cdot %
\left(n_{d{k}}^{- di} + \alpha_{k}  \right) & \ {\rm if \ } s_{di} = 1,
\end{cases}  \label{eq:sample-z-base}
\end{align}
where $n_{k}^{-di}$ ($\tilde{n}_{k}^{-di}$) represents the number of
words (keywords) in the documents assigned to topic (keyword topic)
$k$ excluding the $i$th word of document $d$.  Similarly,
$n_{kv}^{-di}$ ($\tilde{n}_{kv}^{-di}$) denotes the number of times
word (keyword) $v$ is assigned to topic (keyword topic) $k$ again
excluding the $i$th word of document $d$, and $n_{dk}^{-di}$
represents the number of times word $v$ is assigned to topic $k$ in
document $d$ excluding the $i$th word of document $d$.

Next, we sample $s_{di}$ from the following conditional posterior
distribution,
\begin{align*}
  \Pr(s_{di}  = s \mid \bs^{- di}, \bz, \bw,  \bbeta,  \wt{\bbeta}, \bgamma)
  & \ \propto \
\begin{cases}
	\dfrac{\beta_v + n_{z_{di} v}^{- di}    }{ \sum_v \beta_v + n_{z_{di}}^{ - di} } \cdot %
  ( n^{- di}_{z_{di}} + \gamma_2 )  %
	& {\rm if} \quad s  = 0, \\
\dfrac{ \tilde{\beta}_v + \tilde{n}_{z_{di} v}^{- di}    }{   \sum_{v \in \cV_{z_{di}}} \tilde{\beta}_v + \tilde{n}_{z_{di}}^{- di}  } \cdot%
  (\tilde{n}^{- di}_{z_{di}} + \gamma_1 ) %
	&  {\rm if} \quad  s = 1.
\end{cases}
\end{align*}
Finally, the conditional posterior distribution of $\alpha_k$ is given
by,
\begin{align}
  p(\alpha_k \mid \balpha_{-[k]}, \bs, \bz, \bw, \wt{\boldsymbol{\eta}}) \ \propto \   \dfrac{ \Gamma\left(\sum_{k=1}^K \alpha_k \right) \prod_{d=1}^D \Gamma \left(n_{dk} + \alpha_k \right)}{\Gamma(\alpha_k) \prod_{d=1}^D \Gamma \left( \sum_{k=1}^K  n_{dk} + \alpha_k \right)} \cdot
  \alpha_{k}^{\tilde\eta_{1} - 1} \exp(- \tilde\eta_{2} \alpha_{k} ), \label{eq:alpha_base_sample}
\end{align}
for $k=1,2,\ldots,\wt{K}$.  For $k=\wt{K}+1,\ldots,K$, the conditional
distribution is identical except that $\tilde\eta_1$ and
$\tilde\eta_2$ are replaced with $\eta_1$ and $\eta_2$.  We use an
unbounded slice sampler to efficiently sample from a large parameter
space \citep{Mochihashi2020}.

As mentioned earlier, we apply the weighting method of
\wLDA{} when computing these word counts in the collapsed
Gibbs sampler so that frequently occurring words do not overwhelm
other meaningful words.  Based on the information theory,
\citet{Wilson2010} propose a weighting scheme based on $- \log_2 p(v)$
where $p(v)$ is estimated using an observed frequency of term $v$. The
weight for a term $v$ is defined as,
\begin{align}
    m(v) & \ = \  -\log_2 \frac{\sum_{d=1}^D \sum_{i=1}^{N_d} \mathbbm{1}(w_{di} = v) }{\sum_{d=1}^D N_d}. 
\end{align}
Then, the weighted word counts used in the collapsed Gibbs sampler are
given by,
\begin{align*}
  n_{kv} & \ = \ m(v) \sum_{d=1}^D \sum_{i=1}^{N_d} \mathbbm{1}(w_{di}=v) \mathbbm{1}(s_{di}=0)\mathbbm{1}(z_{di}=k), \\
  \tilde{n}_{kv} & \ = \ m(v) \sum_{d=1}^D \sum_{i=1}^{N_d} \mathbbm{1}(w_{di}=v) \mathbbm{1}(s_{di}=1)\mathbbm{1}(z_{di}=k),\\
  n_{dk} & \ = \ \sum_{i=1}^{N_d} m(w_{di}) \mathbbm{1}(z_{di}=k)
\end{align*}
where $m(v) = 1$ for all $v$ corresponds to the unweighted sampler.

\subsection{Model Interpretation}\label{subsec:base-interpret}

To interpret the fitted \keyATM, we focus on two quantities of
interest. The topic-word distribution represents the relative frequency of
words for each topic, characterizing the topic content. The document-topic
distribution represents the proportions of topics for each document,
reflecting the main themes of the document.

We obtain a single topic-word distribution $\bphi^\ast_k$ for
keyword topics by
combining both $\bphi_k$ and $\wt{\bphi}_k$ according to the following
mixture structure assumed under the model for each word $v$ of topic
$k$,
\begin{align}
   \phi^\ast_{kv}
  &= (1 - \pi_k) \phi_{kv} + \pi_k \wt{\phi}_{kv}. \label{eq:base-phi}
\end{align}
Since both $\bphi_k$ and $\wt{\bphi}_k$ are marginalized out, we
compute the marginal posterior mean as our estimate of topic-word
distribution,
\begin{align}
 & \E [\phi^\ast_{kv} \mid \bw]\nonumber \\
 & = \E\Big\{ \E [\phi^\ast_{kv}
       \mid \beta_v,
     \wt{\beta}_v, \bgamma, \bs, \bz, \bw] \Bigm\vert \bw\Big\} \nonumber\\
 &=
    \begin{cases}
     \E\left[\dfrac{ n_{k} + \gamma_2 }{
      \tilde{n}_{k} + \gamma_1 + n_{k} + \gamma_2 } \cdot  \dfrac{ \beta_v + n_{k v} }{ \sum_{v'} \beta_{v'} +
       n_{k}} + \dfrac{ \tilde{n}_{k} + \gamma_1
        }{ \tilde{n}_{k} + \gamma_1 + n_{k} + \gamma_2 }
        \cdot \dfrac{\tilde{\beta}_v + \tilde{n}_{k v}
        }{  \sum_{v' \in \cV_k} \tilde{\beta}_{v'} + \tilde{n}_{k v} } \
        \Biggl | \ \bw \right]
      & \text{if } v \in \mathcal{V}_{k} \\
     \E\left[\dfrac{ n_{k} + \gamma_2 }{
      \tilde{n}_{k} + \gamma_1 + n_{k} + \gamma_2 } \cdot  \dfrac{ \beta_v + n_{k v} }{ \sum_{v'} \beta_{v'} + n_k{k}}  \
        \Biggl | \ \bw \right]
      & \text{if } v \not\in \mathcal{V}_{k}
    \end{cases}
\label{eq:base_Ephi}
\end{align}
where $n_k = \sum_{v=1}^V n_{kv}$,
$\tilde{n}_k = \sum_{v=1}^V \tilde{n}_{kv}$ and the second equality
follows from the conditional independence relations assumed under the
model.
Similarly, although the document-topic distribution
$\theta_{dk}$ is also marginalized out, we compute its marginal
posterior,
\begin{equation}
  \E[\theta_{dk} \mid \bw] \ = \ \E \Big\{ \E[\theta_{dk} \mid \alpha_k,
    \bz, \bw] \Bigm\vert \bw \Big\} \ = \ \E \bigg[\frac{\alpha_k +
   n_{dk}}{\sum_{\kprime=1}^{K} \alpha_{\kprime} + n_{d \kprime} } \biggm\vert
  \bw \bigg],\label{eq:base_Etheta}
\end{equation}
for each document $d$ and topic $k$.

\subsection{Empirical Evaluation}
\label{subsec:base-empirical}

We assess the empirical performance of the base \keyATM{} by analyzing
the texts of Congressional bills, using labels and keywords compiled
by the Comparative Agenda Project
(CAP).\footnote{\url{https://www.comparativeagendas.net} Last accessed
  on December 10, 2019.}  We show that \keyATM{} yields more
interpretable topic-word distributions than \wLDA.  Recall that the
only difference between \keyATM{} and \wLDA{} is the existence of
keyword topics.  In addition, topic-word distributions obtained from
\keyATM{} are more consistent with the human-coded labels given by the
CAP.  Finally, we validate the topic classification of these bills
against the corresponding human coding obtained from the Congressional
Bills Project
(CBP).\footnote{\url{http://www.congressionalbills.org/codebooks.html}
  Last accessed on December 10, 2019. \label{ft:bill-source}} We find
that \keyATM{} outperforms \wLDA, illustrating the improved quality of
estimated document-topic distributions.  A greater correspondence
between the human coding and \keyATM{} outputs suggests the advantage
of \keyATM{} when used for measuring specific topics of interest.

\subsubsection{Data and Setup}
\label{subsec:base-data}

We analyze the Congressional bills that were subject to floor votes
during the 101st to 114th Sessions.\footnote{These sessions are chosen
  for the availability of data.}  These bills are identified via
Voteview\footnote{\url{https://voteview.com} Last accessed on December
  10, 2019.} and their texts are obtained from
\texttt{congress.gov}.\footnote{\url{https://www.congress.gov/} Last
  accessed on December 10, 2019.}  There are a total of 4,421 such
bills with an average of 316 bills per session.  We preprocess the raw
texts by first removing stop words via the \textsf{R} package
\texttt{quanteda} \citep{Benoit2018}, then pruning words that appear
less than 11 times in the corpus, and lemmatizing the remaining words
via the \textsf{Python} library \texttt{NLTK} \citep{Bird2009}.%
\footnote{See Appendix~p.~\pageref{app:base-preprocessing} for details.}
After preprocessing, we have on average 5,537 words per
bill and 7,776 unique words in the entire corpus. The maximum document
length is 152,624 and the minimum is 26.

\begin{table}[t]
   \spacingset{1}
\centering
\begin{tabular}{lrrl}
  \hline \hline
Topic label & Count & Percentage & Most frequent keywords\\
  \hline
Government operations & 864 & 19.54 & administrative capital collection \\
  Public lands & 464 & 10.50 & land resource water \\
  Defense & 433 & 9.79 & security military operation \\
  Domestic commerce & 392 & 8.87 & cost security management \\
  Law \& crime & 274 & 6.20 & code family court \\
  Health & 272 & 6.15 & cost health payment \\
  International affairs & 207 & 4.68 & committee foreign develop \\
  Transportation & 191 & 4.32 & construction transportation air \\
  Macroeconomics & 177 & 4.00 & cost interest budget \\
  Environment & 163 & 3.69 & resource water protection \\
  Education & 138 & 3.12 & education area loan \\
  Energy & 132 & 2.99 & energy vehicle conservation \\
  Technology & 131 & 2.96 & transfer research technology \\
  Labor & 111 & 2.51 & employee benefit standard \\
  Foreign trade & 110 & 2.49 & agreement foreign international \\
  Civil rights & 102 & 2.31 & information contract right \\
  Social welfare &  73 & 1.65 & assistance child care \\
  Agriculture &  68 & 1.54 & product food market \\
  Housing &  65 & 1.47 & housing community family \\
  Immigration &  52 & 1.18 & immigration refugee citizenship \\
  Culture &   2 & 0.05 & cultural culture \\
   \hline \hline
\end{tabular}
\caption{Frequency of each topic and its most frequent
    keywords}
\mynote{The table presents the label of each topic from the
Comparative Agendas Project codebook, the number and proportion of
the bills classified by the human coders of the Congressional Bills
Project for each topic, and three most frequent keywords associated
with each topic.  Note that the \textit{Culture} topic only has two
keywords and the same keywords may appear for different topics.}
\label{tab:base-keywords}
\end{table}

These bills are ideal for our empirical evaluation because the CBP
uses human coders to assign a primary policy topic that follows CAP to
each bill, enabling us to validate the automated classification of
topic models against the manual coding.\footnote{Master Codebook. The
  Policy Agendas Project at the University of Texas at Austin,
  2019. Available at
  \url{https://www.comparativeagendas.net/pages/master-codebook} Last
  accessed on December 10, 2019.}  We derive keywords of each topic
from the brief description provided by the CAP.  We make this process
as automatic as possible to reduce the subjectivity of our empirical
validation (see Appendix~p.~\pageref{app:base-preprocessing}).
Appendix (pp.~\pageref{app:base-different-keywords}--\pageref{app:base-random-keywords})
demonstrates that our empirical results shown below are robust to
selection of different keywords.
Table~\ref{tab:base-keywords} presents the 21 CAP topics, the number
and proportion of the bills assigned by the CBP human coders to each
topic, and their most frequent keywords (see
Table~\ref{table:app-base-keywords} on
Appendix~p.~\pageref{app:base-keywords-selection} for the full list of
keywords).

We fit \keyATM{} and \wLDA{} to this corpus.  For both models, we use
a total of $K=\widetilde{K}=21$ topics and do not include any
additional topics because the CAP topics are designed to encompass all
possible issues in this corpus.  This setting means that for \keyATM,
all topics have some keywords.  We use the default prior specification
of the \keyATM{} package (see footnote~\ref{fn:default_prior}).  For
\wLDA, we use the exactly same implementation and specification as
\keyATM{} with the exception of using no keyword, i.e., $\pi_k = 0$
for all $k$.  We run five independent Markov chains with different
random starting values obtained from the prior distribution of each
model.  We run the MCMC algorithms for 3,000 iterations and obtain the
posterior means of $\phi^\ast_{kv}$ and $\theta_{dk}$ using
Equations~\eqref{eq:base_Ephi}~and~\eqref{eq:base_Etheta},
respectively (Appendix~p.~\pageref{app:base-convergence} analyzes the
convergence).

\subsubsection{Topic Interpretability}
\label{subsec:base-qualitative-comparison}

We begin by examining the interpretability of the resulting topics.
We focus on the topic-word distributions and show that words with high
probabilities given a topic are consistent with the topic's label.
For \keyATM, there is no need to label topics with pre-specified
keywords after model fitting.  In contrast, \wLDA{} requires the
post-hoc labeling of the resulting topics.  Here, we determine the
topic labels such that the document classification performance of
\wLDA{} is maximized (see Appendix~p.~\pageref{app:topicmatch}).  This leads
to a less favorable empirical evaluation for \keyATM.  Below, we show
that even in this setting, \keyATM{} significantly outperforms \wLDA.

\begin{table}[!t]
\centering
\spacingset{1.1}
\begin{tabular}{ll|ll|ll}
   \hline \hline
   \multicolumn{2}{c|}{Labor} & \multicolumn{2}{c|}{Transportation} &
   \multicolumn{2}{c}{Foreign trade}\\ \hline
   \keyATM{} & \wLDA{} & \keyATM{} & \wLDA{} & \keyATM{} & \wLDA{}\\ \hline
\textbf{employee} & apply & \textbf{transportation} & transportation & product$^*$ & air \\
  \textbf{benefit} & tax & \textbf{highway} & highway & \textbf{trade} & vessel \\
  individual & amendment & safety & safety & change & airport \\
  rate & end & carrier & vehicle & \textbf{agreement} & transportation \\
  \textbf{compensation} & taxable & \textbf{air} & carrier & good & aviation \\
  period & respect & code$^*$ & motor & tobacco$^*$ & administrator \\
  code$^*$ & period & system & system & head & aircraft \\
  payment$^*$ & individual & vehicle$^*$ & strike & article & carrier \\
  determine & case & \textbf{airport} & rail & free & administration \\
  agreement$^*$ & relate & motor & code & chapter & coast \\
  \hline \hline
  \multicolumn{2}{c|}{Immigration} & \multicolumn{2}{c|}{Law \& crime} &
  \multicolumn{2}{c}{Government operations}\\ \hline
  \keyATM{} & \wLDA{} & \keyATM{} & \wLDA{} & \keyATM{} & \wLDA{}\\ \hline
security$^*$ & alien & intelligence$^*$ & security & expense & congress \\
  alien & attorney & attorney & information & appropriation & house \\
  \textbf{immigration} & child & \textbf{crime} & intelligence & remain & senate \\
  homeland$^*$ & crime & \textbf{court} & homeland & authorize & office \\
  border$^*$ & immigration & \textbf{enforcement} & committee & necessary & committee \\
  status & grant & \textbf{criminal} & director & transfer$^*$ & commission \\
  nationality & enforcement & \textbf{code} & system & expend & representative \\
  describe & person & offense & foreign & exceed & congressional \\
  individual & court & person & government & office & strike \\
  employer$^*$ & offense & \textbf{justice} & office & activity & bill \\
   \hline
  \hline
   \end{tabular}
   \caption{Comparison of top ten words for six selected
       topics between \keyATM{} and \wLDA{}}
  \label{tab:base-topwords}
  \mynote{The table shows the ten
     words with the highest estimated probability for each topic under
     each model. For \keyATM, the pre-specified keywords for each
     topic appear in bold letters, whereas the asterisks indicate the
     keywords specified for another topic. Both models use term
     weights described in Section~\ref{sec:sampling-base}.}
 \end{table}

 Table~\ref{tab:base-topwords} presents ten words with the highest
 estimated probabilities for six selected topics under each model (see
 Table~\ref{tab:app-base-topwords} on Appendix~p.~\pageref{app:base-topwords}
 for the remaining 15 topics).  For \keyATM, the keywords of each
 topic appear in bold letters whereas the asterisks indicate the
 keywords from another topic.  Each model's result is based on the
 MCMC draws from one of the five chains that has the median
 performance in terms of the overall area
 under the receiver operating characteristics (AUROC).\footnote{For \wLDA, it is
   difficult to combine multiple chains due to the label switching
   problem.  There is no such problem for \keyATM{} since the topics
   are labeled before model fitting.}

 The results demonstrate several advantages of \keyATM.  First, the
 \textit{Labor} topic of \wLDA{} includes many unrelated terms and
 does not contain any terms related to this topic whereas \keyATM{}
 lists many keywords among the most frequent words for the topic, such
 as ``benefit,'' ``employee,'' and ``compensation.''  Second, \wLDA{}
 does not find meaningful terms for the {\it Foreign trade} topic and
 instead creates two topics (labeled as {\it Transportation} and {\it
   Foreign trade}) whose most frequent terms are related to the {\it
   Transportation} topic.  In contrast, the top words selected by
 \keyATM{} represent the content of the {\it Foreign trade} topic,
 while those for the {\it Transportation} capture the substantive
 meaning of the topic well.  As shown in the full top words table in
 Appendix (p.~\pageref{app:base-topwords}), \wLDA{} fails to create topics
 whose top words contain terms related to {\it Labor} or {\it Foreign
   trade}.

Similarly, \wLDA{} has a difficulty selecting the words that
 represent the \textit{Law \& crime} topic and cannot distinguish it
 from the \textit{Immigration} topic.  Indeed, the
 \textit{Immigration} topic for \wLDA{} includes the keywords of the
 \textit{Law \& crime} topic such as ``crime,''
 ``court,'' and ``enforcement.''  In contrast, \keyATM{} selects many
 keywords among the top ten words for each of these two topics without
 conflating them.  This result is impressive since the bills whose
 primary topic is the \textit{Immigration} topic account only for
 1.18\% of all bills.  Thus, \keyATM{} can measure {\it Law \& crime}
 and {\it Immigration} as two distinct topics while \wLDA{} fails to
 separate them.  Finally, both \keyATM{} and \wLDA{} are unable to
 identify the meaningful words for the \textit{Government operations}
 topic, which is the most frequent topic in our
 corpus. Appendix (p.~\pageref{subsubsec:quality}) explains why both models fail
 to uncover this particular topic.

\subsubsection{Topic Classification}
\label{subsubsec:topic-class-base}

Next, to evaluate the quality of topic-document distributions, we
compare the automated classification of \keyATM{} and \wLDA{} with
human coding. The proximity between estimated topic-document
distributions and human coding implies better measurement.
Specifically, we compare the estimated topic-document distribution,
$\hat{\theta}_{dk}$
given in Equation~\eqref{eq:base_Etheta}, with
the primary policy topic assigned by the CBP human coders.  While the
topic models allow each document to belong to multiple topics, the CBP
selects only one primary topic for each bill.  Despite this
difference, we independently evaluate the classification performance
of \keyATM{} against that of \wLDA{} via the ROC curves based on
$\hat{\theta}_{dk}$ for each topic $k$. As noted earlier, our
evaluation setting favors \wLDA{} because \wLDA{} topics are matched
with the CBP topics by maximizing its classification performance.

\begin{figure}[!t]
  \spacingset{1}
  \includegraphics[width=\linewidth]{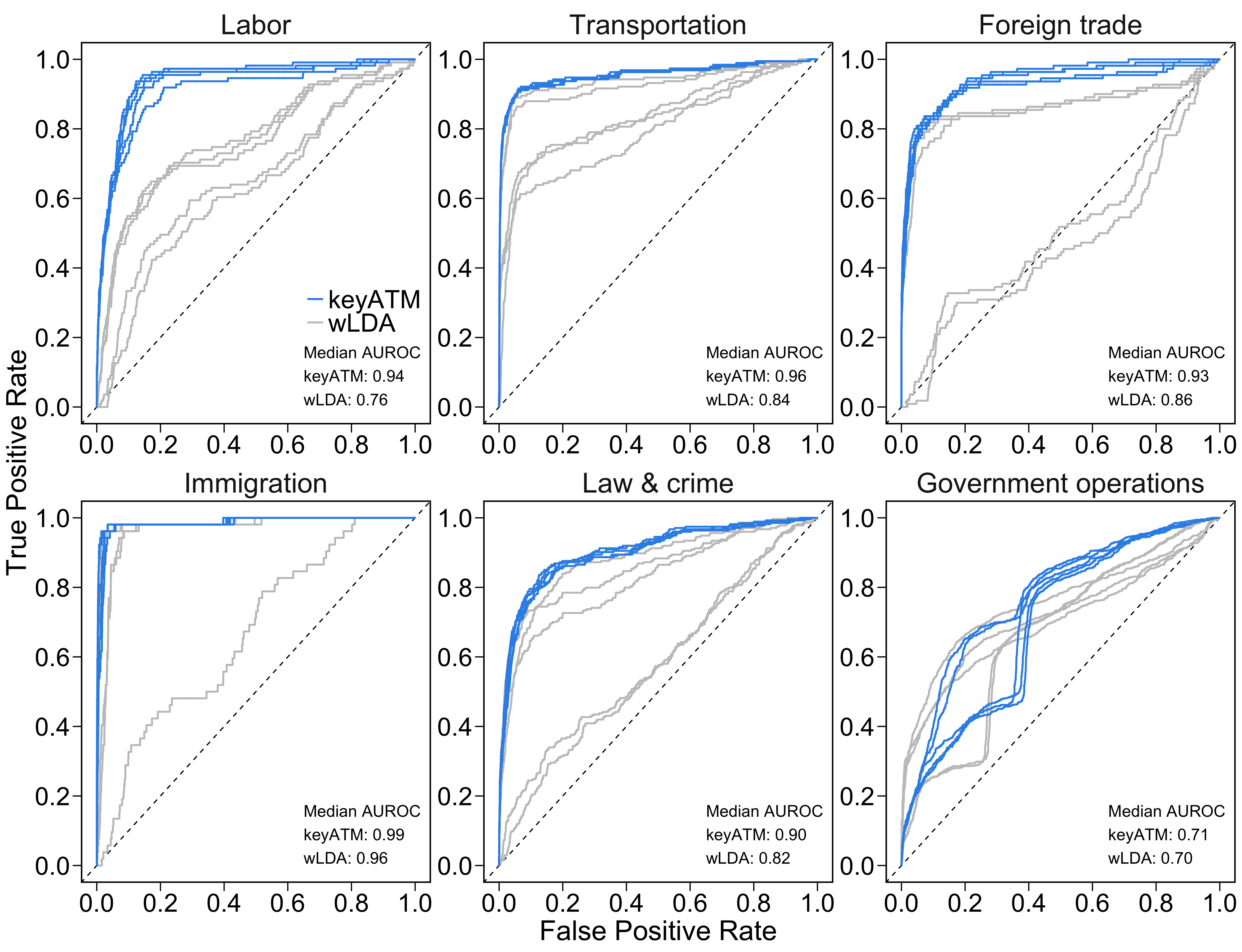}
  \caption{Comparison of the ROC curves between \keyATM{}
      and \wLDA{} for six selected topics}
 \label{tab:base-roc}
  \mynote{Each line represents the
    ROC curve from one of the five Markov chains with different
    starting values for \keyATM{} (blue lines) and \wLDA{} (gray
    lines). The median AUROC indicates the median value of AUROC among
    five chains for each model.  The plots show that \keyATM{} has a
    better topic classification performance than \wLDA{} with the
    exception of the ``Government operations'' topic.  The results of
    \keyATM{} are also less sensitive to the starting values.}
\end{figure}

Figure~\ref{tab:base-roc} presents the ROC curves for the same six
selected topics as those shown in Table~\ref{tab:base-topwords} (see
Appendix~p.~\pageref{app:base-roc} for the other topics).  Each line
represents the ROC curve based on one of the five Markov chains with
different starting values for \keyATM{} (blue lines) and \wLDA{} (gray
lines).  We find that \keyATM{} outperforms \wLDA{} except for the
\textit{Government operations} topic.  The results are consistent with
the qualitative evaluation based on Table~\ref{tab:base-topwords}.
For example, the poor performance of both models for the {\it
  Government operations} is not surprising given that their top words
are not informative about the topic content
(Appendix~p.~\pageref{subsubsec:quality} explains this underperformance).  When compared to \wLDA,
\keyATM{} has a much better classification performance for the {\it
  Labor}, {\it Transportation}, {\it Foreign trade}, and {\it Law \&
  crime} topics, where its topic interpretability is superior.
Finally, the ROC curves for \keyATM{} are less sensitive to different
starting values than those for \wLDA{} with the exception of the {\it
  Government operations} topic.

\section{The Covariate \keyATM}\label{sec:covariate}
Next, we extend the base \keyATM{} by incorporating covariates for the
document-topic distribution.  The inclusion of covariates is useful
as social scientists often have meta-information about documents
(e.g., authorship).  We adopt the Dirichlet-Multinomial regression
framework of \citet{Mimno2008} rather than the logistic normal regression
approach of the \STM{} in \citet{Roberts2016}
so that the collapsed Gibbs sampler strategy for the base \keyATM{} can be
used.

\subsection{Model}

Suppose that we have an $M$ dimensional covariate $\bx_d$ (including
an intercept) for each document $d$.  We model the document-topic
distribution using these covariates in the following fashion (in place
of Equations~\eqref{eq:theta_base}~and~\eqref{eq:alpha_base}),
\begin{eqnarray*}
  \btheta_{d}   &\indep & \Dir(\exp(\blambda^\top \bx_{d})) \\
  \lambda_{mk}  &\iid &  \Normal (\mu, \sigma^2)
\end{eqnarray*}
for each $d = 1,2,\ldots,D$ where $\blambda$ is an $M \times K$ matrix
of coefficients and $\lambda_{mk}$ is the $(m, k)$ element of
$\blambda$.  The sampling algorithm and the model interpretation are
straightforward extension of the base \keyATM{}
(Appendix~p.~\pageref{app:sampling-cov}).\footnote{We do not
  directly model the correlation across topics.  Although we have
  explored alternative modeling strategies including the Logistic-Normal approach of \citet{Roberts2016}, we find that the
  proposed models generally perform well without directly modeling the
  correlation structure.\label{fn:logistic-normal}}

\subsection{Empirical Evaluation}\label{subsec:cov-empirical}

We evaluate the empirical performance of the covariate \keyATM{}
against that of \STM{} using the Japanese election manifesto data
\citep{Catalinac2015}. Analyzing manifestos of Japan's Liberal
Democratic Party (LDP) candidates, the author finds that the 1994
electoral reform is associated with a relative increase in the topics
about programmatic policies and a decline in the topics about pork
barrel. Since the manifestos come from eight elections and the author
focuses on LDP candidates, we include the election-year and LDP
dummies as the covariates. We find that the covariate \keyATM{} yields
more interpretable topics and its results are less sensitive to the
total number of topics chosen by researchers than \STM{}.

\subsubsection{Data and Setup}

We analyze a total of 7,497 manifestos \citep{Shinada2006}. \citet{Catalinac2015}
preprocessed the data by tokenizing Japanese sentences, removed
punctuations and stop words, and cleaned up the documents based on an
author-defined dictionary. We use the document-term matrix from the
original study so that the preprocessing steps remain identical.
In Japanese elections, every registered political
candidate is given a fixed amount of space in a government
publication, in which their manifesto can be printed.  This document,
containing the manifestos of all candidates, is then distributed to
all registered voters.  After preprocessing, the average number of
words is about 177 (the maximum is 543 and the minimum is 4), whereas
the number of unique terms is 2,832.

These manifestos cover 3,303 unique candidates who ran in
the eight consecutive elections held between 1986 and 2009.  Because
Japanese electoral campaigns are heavily restricted, the manifestos
represent one of the few ways, in which candidates communicate their
policy goals to voters.

\subsubsection{Keyword Construction}
\label{subsubsec:keyword-construction}

Unlike the validation study presented in
Section~\ref{subsec:base-empirical}, we do not have human-coded topics
for this data set.  In the original article, the author applies \LDA{}
with 69 topics and label all the topics after fitting the model by
carefully examining the 15 most frequent words for each topic. The
author, then, discusses how the estimated topic proportions change
after the 1994 electoral reform.  To apply the covariate \keyATM{}, we
must develop a set of keywords for topics. Unfortunately, we cannot
merely use the most frequent words identified by \citet{Catalinac2015}
with \LDA{} as the keywords because that would imply analyzing the
same data as the one used to derive keywords.

To address this problem, we independently construct keywords using the
questionnaires of the UTokyo-Asahi Surveys (UTAS), which is a
collaborative project between the University of Tokyo and the Asahi
Shimbun, a major national newspaper.
Table~\ref{tab:cov-keywords} presents the resulting 16 topics and
their keywords (two pork barrel and 14 programmatic policy topics).
Since most UTAS questions consist of a single sentence, we typically
choose nouns that represent each topic's substantive meanings (Appendix~p.\pageref{app:CovKeywords} explains details of keyword construction).

\begin{table}[!t]
   \spacingset{1}
\centering
\begin{tabular}{lll}
  \hline
  Type         & Topic label               & Keywords\\
  \hline
  Pork barrel  & Public works              & employment, public, works \\
               & Road construction         & road, budget \\
  \hdashline
  Programmatic & Regional devolution       & rural area, devolve, merger \\
               & Tax                       & consumption, tax, tax increase \\
               & Economic recovery         & economic climate, measure, fiscal policy, deficit \\
               & Global economy            & trade, investment, industry \\
               & Alternation of government & government, alternation \\
               & Constitution              & constitution \\
               & Party                     & party, political party \\
               & Postal privatization      & postal, privatize \\
               & Inclusive society         & women, participate, civilian \\
               & Social welfare            & society, welfare \\
               & Pension                   & pension \\
               & Education                 & education \\
               & Environment               & environment, protection \\
               & Security                  & defense, foreign policy, self defense \\
   \hline
\end{tabular}
\caption{Keywords for each topic}
\label{tab:cov-keywords}
\mynote{The left and middle columns
  show the types of policies and topic labels assigned by
  \citet{Catalinac2015}. The corresponding keywords in the right
  column are obtained from the UTokyo-Asahi Surveys (UTAS). This
  results in the removal of five policy areas (sightseeing, regional
  revitalization, policy vision, political position, and investing
  more on human capital) that do not appear in the UTAS.}
\end{table}

Finally, we fit the covariate \keyATM{} and \STM, using seven
election-year indicator variables and another indicator for the LDP
candidates.  We examine the degree to which the results are sensitive
to model specification by varying the total number of topics.
Specifically, in addition to the 16 keyword topics, we include
different numbers of extra topics with no keyword.  We try 0, 5, 10,
and 15 no-keyword topics. We fit \keyATM{} for 3000 iterations with a
thinning of 10 and the default hyper-parameter values.  We use the
default settings of the \STM{} package.

\subsubsection{Topic Interpretability}

\begin{table}[!t]
\centering
\spacingset{1.1}
\begin{tabular}{ll|ll|ll}
  \hline \hline
  \multicolumn{2}{c|}{Road construction} &  \multicolumn{2}{c|}{Tax} & \multicolumn{2}{c}{Economic recovery}\\
  \hline
  \keyATM & \STM & \keyATM & \STM & \keyATM & \STM  \\
  \hline
  development  & tax  & Japan  & Japan  & reform  & reform \\
   \textbf{road}  & reduce tax  &  \textbf{tax}  & citizen  &  \textbf{measure}  & postal \\
  city  & yen  & citizen  & JCP  & society*  & privatize \\
  construction  & housing  & JCP  & politic  & Japan  & Japan \\
  tracks  & realize  &  \textbf{consumption}  & tax  &  \textbf{economic climate}  & rural area \\
   \textbf{budget}  & daily life  & politic  & consumption  & reassure  & country \\
  realize  & move forward  &  \textbf{tax increase}  & tax increase  & economy  & citizen \\
  promote  & city  & oppose  & oppose  & institution  & safe \\
  move forward  & education  & business  & business  & safe  & government \\
  early  & measure  & protect  & protect  & support  & pension \\
  \hline \hline
  \multicolumn{2}{c|}{Inclusive society} &  \multicolumn{2}{c|}{Education} & \multicolumn{2}{c}{Security}\\
  \hline
  \keyATM & \STM & \keyATM & \STM & \keyATM & \STM  \\
  \hline
  politic  & politic  & politic  & Japan  & Japan  & society \\
   \textbf{civilian}  & reform  & Japan  & person  &  \textbf{foreign policy}  & Japan \\
  society*  & new  & person  & country  & peace  & world \\
   \textbf{participate}  & realize  &  \textbf{children}  & politic  & world  & economy \\
  peace  & citizen  &  \textbf{education}  & necessary  & economy  & environment \\
  welfare*  & government  & country  & problem  & country  & international \\
  aim  & daily life  & make  & children  & citizen  & education \\
  human rights  & rural area  & force  & force  &  \textbf{defense}  & country \\
  realize  & corruption  & have  & have  & safe  & peace \\
  consumption*  & change  & problem  & future  & international  & aim \\
  \hline \hline
   \end{tabular}
   \caption{Comparison of top ten words for six selected
       topics between the covariate \keyATM{} and \STM{}}
  \label{tab:cov-topwords}
    \mynote{The table
     shows the ten words with the highest estimated probabilities for
     each topic under each model. For \keyATM, the pre-specified
     keywords for each topic appear in bold letters whereas the
     asterisks indicate the keywords specified for another topic.}
\end{table}

Table~\ref{tab:cov-topwords} lists the ten most frequent words for each of
the six selected topics.
These topics are chosen since they are easier to understand without
the knowledge of Japanese politics (see
Appendix~p.~\pageref{app:cov-topwords} for the results of the other topics).
We match each topic of \STM{} with that of \keyATM{} by applying the
Hungarian algorithm to the estimated topic-word distributions so that
the overall similarity between the results of the two models is
maximized.

We find that the covariate \keyATM{} produces more interpretable
topics, judged by these ten most frequent words, than \STM.  For
example, \keyATM{} identifies, for the \textit{Road construction}
topic, the terms such as ``development,'' ``construction,'' and
``track'' as well as two assigned keywords, ``road'' and ``budget.''
This makes sense given that developing infrastructures such as road and
railway tracks is considered as one of the most popular pork barrel
policies in Japan.  For the \textit{Education} topic, \STM{} does not
include ``education,'' whereas \keyATM{} includes two selected
keywords, ``children'' and ``education.''  Finally, for the
\textit{Security} topic, \keyATM{} lists terms such as ``peace,''
``safe,'' and ``international,'' while the terms selected by \STM{}
broadly cover international economy and politics.

\subsubsection{Topic Discrimination}

\begin{figure}[!t]
  \spacingset{1}
  \centering
\includegraphics[width=\linewidth]{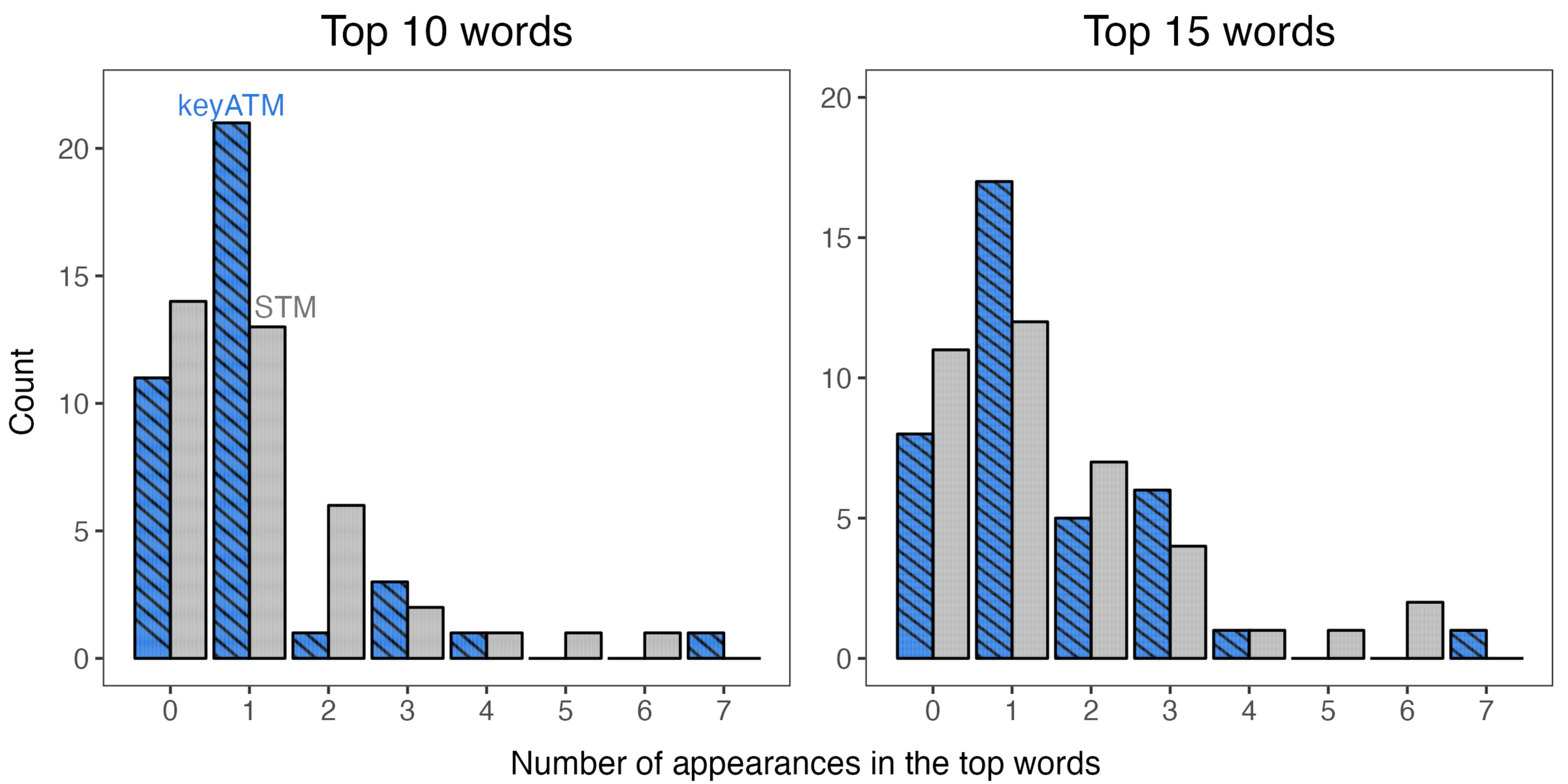}
\caption{Exclusivity of keywords across topics}
\label{fig:cov-spread}
\mynote{Left (right)
  bar plot shows the number of times that each of the 38 keywords
  appears as the 10 most frequent (15 most frequent) words of keyword
  topics.  These words are less likely to be shared across topics
  under the covariate \keyATM{} (blue shaded bars) than under the
  \STM{} (gray bars).}
\end{figure}

Good topic models should yield topics distinct from one another, which
means that we would like different topics to have different words
representing them.  The bar plots in Figure~\ref{fig:cov-spread}
present the number of times that each of the 38 keywords appears in
the top 10 (left panel) or 15 (right panel) words of keyword topics.  As
expected, the covariate \keyATM{} (blue shaded bars) assigns the same
keywords to fewer topics than \STM{} (gray bars).  In particular, more
keywords appear as the most frequent terms only for a single topic
under \keyATM{} than under \STM.

\subsubsection{Covariate Effects}\label{sec:cov-effects}

One key hypothesis of \citet{Catalinac2015} is that after the 1994
electoral reform, LDP candidates adopted electoral strategies to
pursue more programmatic policies.  The author tests this hypothesis
by plotting the estimated topic proportions for each election year.
Here, we take advantage of the fact that the covariate \keyATM{} and
\STM{} can directly incorporate covariates. The quantities of interest
are the election-year proportions of the pork barrel and programmatic
topics for LDP politicians. Specifically, we first compute, for each
topic, the posterior mean of document-topic probability for LDP
manifestos within each election year by using
Equation~\eqref{eq:cov_Etheta} in Appendix (p.~\pageref{app:sampling-cov}) with
the appropriate values of covariates.  We then compute the sum of
these posterior mean proportions for each policy type as an estimate
of the total proportion. In addition, we examine the sensitivity of
these results to the choice of the total number of no-keyword topics
for both \keyATM{} and \STM.

\begin{figure}[!t]
  \spacingset{1}
  \centering
\includegraphics[width=\linewidth]{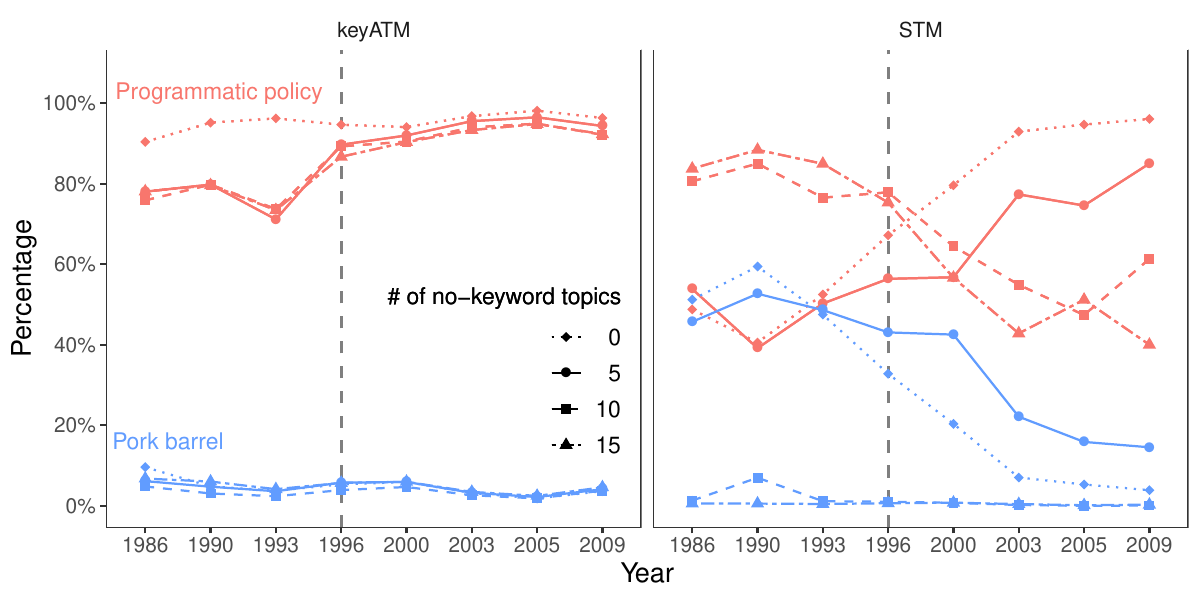}
\caption{Programmatic policy topics increase right after the
    1994 electoral reform}
\label{fig:cov-amy_fix}
\mynote{The results based on the covariate
  \keyATM{} (left panel) shows that the estimated proportion of
  programmatic policy topics increased in the 1996 election right
  after the election reform.  The results are not sensitive to the
  number of topics except when there is no additional no-keyword
  topic.  Note that all topics in \STM{} are referred to
    as no-keyword topics since they do not have pre-assigned
    keywords. The total number of topics is the same between two models.
  The results based on \STM{} vary substantially across
  different numbers of topics.}
\end{figure}

Figure~\ref{fig:cov-amy_fix} plots the sum of estimated topic
proportions corresponding to pork barrel (blue) and programmatic
policies (red), for the LDP candidates.
All topics in \STM{} are referred to as
  no-keyword topics since they do not have pre-assigned keywords.
  The total number of topics is the same between two models.
The plot omits credible
intervals because they are too narrow to be visible.  Consistent with
the original analysis, we generally find that in the first election
after the 1994 electoral reform, the proportion of programmatic policy
topics substantially increased whereas the proportion of pork barrel
topics remain virtually unchanged.  For the covariate \keyATM, this
finding is consistent across all model specifications except the model
without no-keyword topics (dotted lines with solid diamonds).  This
model without any no-keyword topic is not credible in this application
because these keywords taken from UTAS do not cover the entire
contents of manifestos.  In contrast, the performance of \STM{} is
much more sensitive to the total number of topics.  The change after
the electoral reform is also less stark when compared to the covariate
\keyATM{}.
In sum, the covariate \keyATM{} yields more reasonable and robust
results than \STM.

\section{The Dynamic \keyATM}
\label{sec:dynamic}

The second extension we consider is the dynamic modeling of
document-topic distributions.  Researchers are often interested in
investigating how the prevalence of topics changes over time
\citep{Clarke2020}.  Building on the collapsed Gibbs sampler used for
the base \keyATM, we apply the Hidden Markov Model (HMM).  HMM has
been used to introduce time dynamic components in various applications
(e.g., \citet{quinn2010,park:12,Knox2021,Olivella2022}; see
Appendix~p.~\pageref{app:dynamic-literature} for details).  An alternative
modeling strategy is to use time fixed effects either in the covariate
\keyATM{} or in a regression model fitted to the output from the base
\keyATM.  Unlike these approaches, the dynamic \keyATM{} incorporates
time ordering and smoothly models time trend while properly accounting
for uncertainty.

\subsection{Model}

Suppose that we have a total of $T$ time periods and each document $d$
belongs to one of these time periods, $t[d] \in \{1,2,\ldots,T\}$.
The HMM is based on the idea that each time period belongs to one of
the latent discrete states.  Assume that we have a total of $R$ such
states and use $h_t \in \{1,2,\ldots,R\}$ to denote the latent state
for time $t$.  Following \cite{Chib1998}, we only allow for one-step
forward transition, implying that the probability of transition from
state $r$ to state $r^\prime$, i.e.,
$p_{r r^\prime} = \Pr(h_{t+1} = r^\prime \mid h_{t} = r)$, is equal to
zero unless $r^\prime = r + 1$.  This assumption considerably
simplifies the estimation without sacrificing model fit so long as we
have a sufficiently large number of states.  The resulting Markov
transition probability matrix is given by,
\begin{align*}
  \bP & \ = \ \left(
    \begin{array}{cccccc}
      p_{11} & p_{12} & 0 & \cdots & 0 & 0 \\
      0 & p_{22} & p_{23} & \cdots  & 0 & 0\\
      \vdots & \vdots & \vdots & \vdots  & \vdots & \vdots \\
      0 & 0 & 0 & \cdots & p_{R-1,R-1}  & p_{R-1,R} \\
      0 & 0 & 0 & \cdots & 0  & 1
    \end{array}
                            \right).
\end{align*}
The prior distribution for the probability of no transition is
uniform, i.e., $p_{r r} \iid {\sf Uniform}(0, 1)$ for
$r=1,2,\ldots,R$.  Finally, the dynamic \keyATM{} allows the topic
proportion $\btheta_d$ to evolve over time by letting $\balpha$ to
vary across the latent states. Modeling $\balpha$ instead of
$\btheta_d$ makes \keyATM{} less sensitive to the short-term
temporal variation.
Thus, instead of
Equation~\eqref{eq:alpha_base} we have,
\begin{eqnarray}
  \alpha_{rk} & \iid & \Gammad(\eta_1, \eta_2)  \quad \text{for} \ r = 1,2,\ldots,R \ \text{and} \ k = 1,2,\ldots,K
\end{eqnarray}
The sampling algorithm and the model interpretation are
straightforward extension of the base \keyATM{} (Appendix~p.~\pageref{app:sampling-dyn}).

\subsection{Empirical Evaluation} \label{subsec:dyn-empirical}

In this section, we empirically evaluate the performance of the
dynamic \keyATM{} by analyzing the corpus of the United States Supreme
Court opinions from the Supreme Court Database (SCD) project.%
\footnote{\url{http://scdb.wustl.edu/} Accessed
  December 10, 2019} Like the Congressional bill data set analyzed in
Section~\ref{subsec:base-empirical}, the key advantage of this data
set is that human coders have identified the primary topic of each
opinion and each topic comes with keywords.  The only difference
between the dynamic \keyATM{} and \wLDA{} is the existence of keyword
topics.  We show that the dynamic \keyATM{} yields more interpretable
topics and better classification performance than the dynamic \wLDA{}.
Moreover, the time trend of topic prevalence estimated with \keyATM{}
is closer to the human coding than that of \wLDA{} without keywords.

\subsubsection{Data and Setup}

We analyze a total of 17,245 Supreme Court opinions written between
1946 and 2012, with an average of 265 opinions per year \citep{Rice2017}.%
\if0\blind
\footnote{The authors thank Douglas Rice for generously sharing the
text data of the Supreme Court opinions. \label{ft:court-source}}
\fi
We preprocess these texts using the same procedure used in
Section~\ref{subsec:base-data},
yielding a corpus with an average of 1,298 words per document and
a total of 9,608 unique words. The maximum number of words for
a document is 30,767, while the minimum is 1.

\begin{table}[!t]
\centering \spacingset{1}
\begin{tabular}{lrrl}
  \hline \hline
Topic label & Count & Percentage & Most frequent keywords \\
  \hline
Criminal procedure & 4268 & 24.75 & right rule trial evidence justice \\
  Economic activity & 3062 & 17.76 & federal right claim evidence power \\
  Civil rights & 2855 & 16.56 & right public provision party constitutional \\
  Judicial power & 1964 & 11.39 & federal right district rule claim \\
  First amendment & 1795 & 10.41 & amendment first public party employee \\
  Due process & 738 & 4.28 & right defendant constitutional employee process \\
  Federalism & 720 & 4.18 & federal tax regulation property support \\
  Unions & 664 & 3.85 & right employee standard union member \\
  Federal taxation & 529 & 3.07 & federal claim provision tax business \\
  Privacy & 290 & 1.68 & right regulation information freedom privacy \\
  Attorneys & 188 & 1.09 & employee attorney official bar speech \\
  Interstate relations & 119 & 0.69 & property interstate dispute foreign conflict \\
  Miscellaneous &  50 & 0.29 & congress authority legislative executive veto \\
  Private action &   3 & 0.02 & evidence property procedure contract civil \\
       \hline \hline
\end{tabular}
\caption{Frequency of each topic and its most common
    keywords}
\label{tab:dyn-keywords}
\mynote{The table presents the label of each topic from the
  Supreme Court Database (SCD) codebook, the number and proportion of
  the opinions assigned to each topic by the SCD human coders, and
  five most frequent keywords associated with each topic.  Note that
  the same keywords may appear for different topics.}
\end{table}

The SCD project used human coders to identify the primary issue area
for each opinion.%
\footnote{Scholars who study judicial politics have used this issue
  code \citep{Rice2017}.} According to the project website, there are
a total of 278 issues across 14 broader categories.  We use the
aggregate 14 categories as our keyword topics, i.e.,
$\widetilde{K}=14$.  We obtain the keywords from the issue
descriptions provided by the SCD project.  We apply the same
preprocessing procedure used in Section \ref{subsec:base-data}.
Appendix (p.~\pageref{app:dynamic-keywords}) provides further details about the
keyword construction.  Table~\ref{tab:dyn-keywords} presents these 14
topics from the SCD project, the number and proportion of the opinions
classified to each topic by the SCD human coders, and their five most
frequent keywords (see Table~\ref{app:dynamic-keywords-list} on
Appendix p.~\pageref{app:dynamic-keywords} for the full list of keywords).

We fit the dynamic \keyATM{} and \wLDA{} to this corpus.  Since the
SCD topic categories are supposed to be comprehensive, we do not
include any additional topics that do not have keywords, i.e.,
$K=\widetilde{K}=14$.  Therefore, all topics have some keywords for
\keyATM.  For the HMM specification, we use a total of 5 states, i.e.,
$R=5$, because 5 states performed the best in term of the commonly
used perplexity measure.  For the hyper-parameters, we use the default
values provided by the \keyATM{} package. Finally, the implementation
for the dynamic \wLDA{} is identical to that of the dynamic \keyATM{}
with the exception of setting $\pi = 0$ (i.e., no keyword).

As in Section~\ref{subsec:base-empirical}, we run five independent
Markov chains for 3,000 iterations for each model with different
starting values independently sampled from the prior distribution. We
compute the posterior means of $\phi^\ast_{kv}$ and $\theta_{dk}$ using
Equations~\eqref{eq:base_Ephi}~and~\eqref{eq:dyn-Etheta}.  After
fitting the models, we match the resulting topics from the dynamic
\wLDA{} with the SCD topics by maximizing its classification
performance (see Section~\ref{subsec:base-qualitative-comparison}).
There is no need to apply this procedure to the dynamic \keyATM{}
because the topic labels are determined when specifying keywords
before fitting the model.  Thus, our empirical evaluation provides the
least (most) favorable setting for the dynamic \keyATM{} (\wLDA).

\subsubsection{Topic Interpretability}

\begin{table}[!t]
\centering \spacingset{1.1}
\begin{tabular}{ll|ll|ll}
  \hline
  \hline \multicolumn{2}{c|}{Criminal procedure} & \multicolumn{2}{c|}{First amendment} & \multicolumn{2}{c}{Unions}\\ \hline
  \keyATM{} & \wLDA{} & \keyATM{} & \wLDA{} & \keyATM{} & \wLDA{}\\ \hline
\textbf{trial} & trial & \textbf{public} & public & \textbf{employee} & employee \\
  \textbf{jury} & jury & \textbf{amendment} & first & \textbf{union} & union \\
  defendant$^*$ & petitioner & \textbf{first} & speech & board & labor \\
  \textbf{evidence} & evidence & government & amendment & \textbf{labor} & employer \\
  \textbf{criminal} & defendant & may & interest & \textbf{employer} & board \\
  \textbf{sentence} & counsel & interest & party & agreement & agreement \\
  petitioner & right & \textbf{speech} & may & employment$^*$ & contract \\
  judge & rule & right$^*$ & right & contract$^*$ & employment \\
  conviction & make & can & political & \textbf{work} & bargaining \\
  \textbf{counsel} & judge & \textbf{religious} & government & \textbf{bargaining} & work \\
      \hline  \multicolumn{2}{c|}{Federal taxation} & \multicolumn{2}{c|}{Civil rights} & \multicolumn{2}{c}{Privacy}\\ \hline
      \keyATM{} & \wLDA{} & \keyATM{} & \wLDA{} & \keyATM{} & \wLDA{}\\ \hline
\textbf{tax} & tax & district$^*$ & school & search$^*$ & child \\
  property$^*$ & property & \textbf{school} & district & officer & benefit \\
  pay & income & \textbf{discrimination} & religious & police & interest \\
  income & pay & election$^*$ & discrimination & amendment$^*$ & medical \\
  payment & bank & \textbf{equal} & county & arrest & plan \\
  interest & interest & county & election & warrant & provide \\
  benefit$^*$ & corporation & vote & vote & fourth & parent \\
  amount & payment & \textbf{plan} & equal & evidence$^*$ & woman \\
  plan$^*$ & amount & one & education & person & may \\
  fund$^*$ & business & race & student & use & statute \\
   \hline
\end{tabular}
\caption{Comparison of ten top words for selected six topics
    between the dynamic \keyATM{} and  dynamic \wLDA}
\label{tab:dynamic-topwords}
\mynote{
  The table shows the ten
  words with the highest estimated probabilities for each topic under
  each model. For the dynamic \keyATM{}, the pre-specified keywords
  for each topic appear in bold letters, whereas the asterisks
  indicate the keywords specified for another topic.}
\end{table}

We first compare the interpretability of the topics obtained from the
dynamic \keyATM{} and \wLDA{}. Table~\ref{tab:dynamic-topwords}
presents the ten words with the highest estimated probabilities
defined in Equation~\eqref{eq:base-phi} for selected six topics (see
Table~\ref{tab:app-dynamic-topwords} on
Appendix~p.~\pageref{app:dynamic-topwords} for the remaining 8 topics).  For
the dynamic \keyATM, the pre-specified keywords appear in bold letters
while the asterisks indicate the keywords specified for another topic.
The results for each model are based on the MCMC draws from one of the
five chains that has the median performance in terms of the overall
AUROC.

We find the resulting topics of the dynamic \keyATM{} are at least as
interpretable as those discovered by the dynamic \wLDA.  For example,
the top ten words selected by \keyATM{} for the \textit{First
  amendment} topic contains the relevant keywords such as ``first,''
``amendment,'' ``speech,'' and ``religious.''  In addition, \keyATM{}
can collect substantively meaningful terms even when only a small
number of keywords appear in top frequent words.  For example, for
\keyATM{}, only one of the 19 keywords, ``tax,'' appears in the list
of the top ten words for the {\it Federal taxation} topic.  And yet,
the other words on the list, such as ``income'' and ``pay,'' are
highly representative of the substantive meaning of the topic.
Finally, both \keyATM{} and \wLDA{} fail to identify the meaningful
terms for the \textit{Privacy} topic.
Appendix (p.~\pageref{app:dynamic-quality}) shows that this is because the
keywords for the \textit{Privacy} topic do not frequently appear in
the opinions assigned to this topic by the SCD project.

\subsubsection{Topic Classification}

Next, we compare the classification performance of the dynamic
\keyATM{} and \wLDA{} with the human coding from the SCD project.  We
apply the same procedure as the one used in
Section~\ref{subsubsec:topic-class-base} and compute the ROC curve and
AUROC based on the estimated topic-document distribution,
$\hat{\btheta}_d$, given in Equation~\eqref{eq:dyn-Etheta}.  As
mentioned earlier, the results are more favorable to \wLDA{} because
we match its topics with the SCD topics by maximizing the AUROC of the
\wLDA.

\begin{figure}[!t]
    \includegraphics[width=\linewidth]{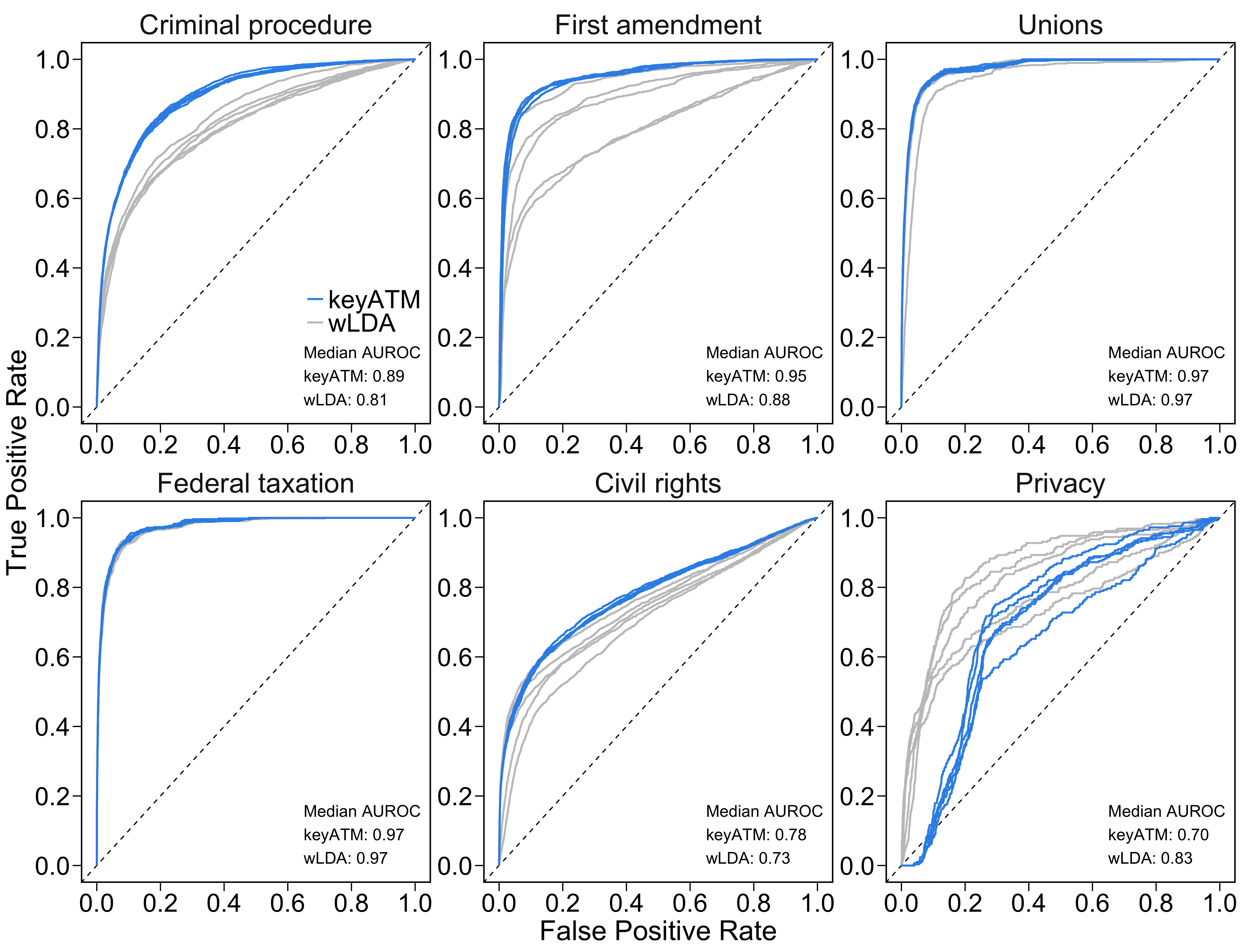}
    \caption{Comparison of the ROC curves between the dynamic
        \keyATM{} and  dynamic \wLDA{} for six selected topics}
    \label{fig:dyn-roc}
    \mynote{Each line
      represents the ROC curve from one of the five Markov chains with
      different starting values for the dynamic \keyATM{} (blue lines)
      and \wLDA{} (gray lines). The plots show that \keyATM{} has a
      better topic classification performance than \wLDA{} with the
      exception of the {\it Privacy} topic. The median AUROC indicates
      the median value of AUROC among five chains for each model.  The
      results of \keyATM{} are also less sensitive to the starting
      values.}
\end{figure}

Figure~\ref{fig:dyn-roc} presents the ROC curves for the same six
selected topics as those shown in Table~\ref{tab:dynamic-topwords}
(see Appendix~p.~\pageref{app:dynamic-roc} for the other topics).  Each line
represents the ROC curve based on one of the five Markov chains for
the dynamic \keyATM{} (blue lines) and \wLDA{} (gray lines) while the
AUROC value is based on the chain with the median performance.  We
find that \keyATM{} outperforms \wLDA{} except for the
\textit{Privacy} topic.  Recall that for this topic, the top words
identified by both models have little substantive relevance.  Lastly,
the ROC curves for \keyATM{} are in general less sensitive to
different starting values than those for \wLDA{}, again, except for
the {\it Privacy} topic.

\subsubsection{Time Trends of Topic Prevalence}\label{sec:dynamic-time-prevalence}

Finally, we compare the time trend of estimated topic prevalence
between each of the two topic models and the SCD human coding.
We first calculate the mean proportion of each topic by
  using all documents that belong to each time point (see
  Appendix~p.~\pageref{app:dyn-model-interpretation} for details). For the
SCD human coding, we compute the proportions of documents that are
assigned to the topic of interest in each year.  Note that the topic
models assign multiple topics to each document, whereas the SCD coding
classifies each document only to one of the 14 topics.  As a result,
these two proportions are not directly
comparable.  Therefore, we use the standardized measure
to focus on relative time trends for comparison (i.e., subtract its
mean from each data point and then divide it by its standard
deviation).


\begin{table}[t]
  \centering
  \begin{tabular}{rcc}
    \hline
    Topic & Dynamic \keyATM & Dynamic \wLDA \\
    \hline
     Criminal procedure & 0.83 & 0.07 \\
     First amendment & 0.82 & 0.64 \\
     Unions & 0.78 & 0.74 \\
     Federal taxation & 0.80 & 0.79 \\
     Civil rights & 0.76 & 0.70 \\
     Privacy & 0.07 & 0.18 \\
     \hline
  \end{tabular}
  \caption{Comparison of the time trends of topic prevalence
      between the dynamic \keyATM{} / \wLDA{} and the SCD human coding
      for six selected topics}
  \label{tab:dyn-trend}
  \mynote{The table shows the correlation
    between the estimated topic prevalence in
    Equation~\eqref{eq:dyn-estimated-trend} and the SCD human coding. To
    compare the estimated topic prevalence and the SCD human coding,
    we use the standardized measure that subtracts its mean from each
    data point and then divide it by its standard deviation.  The
    results show that \keyATM{} exhibits a higher correlation with
    the human coding for most topics than the dynamic \wLDA.}
\end{table}

Table~\ref{tab:dyn-trend} presents the correlation
of the estimated topic prevalence between each of the
two topic models and the SCD human coding
(see Appendix~p.~\pageref{app:dynamic-cor} for the other topics).
We find that \keyATM{} exhibits a higher correlation with the human
coding for most topics than \wLDA{}.
For the {\it Privacy} topic, both \keyATM{} and \wLDA{} only
weakly correlates with the human coding.  This result is not surprising given
the poor performance of \keyATM{} for this topic in terms of both
topic interpretability and classification (see
Appendix~p.~\pageref{app:dynamic-quality}).

\section{Crowdsource Validation}
\label{sec:validation}

We conduct additional validation through crowdsourcing regarding the
results about the superior topic interpretability of \keyATM{} over
\wLDA. We find that \keyATM{} generally improves the interpretability
of estimated topics and their correspondence with labels without
sacrificing topic coherency.

\subsection{The Validation Methodology}\label{sec:validation-methodology}

We follow the validation framework proposed by \citet{Chang2009} and
\citet{Ying2021} to evaluate the resulting topics. First, we measure
the semantic coherency of each topic via Random 4 Word Set Intrusion
\citep[R4WSI, coherency task;][]{Ying2021}.  A worker sees four different word sets
(each word set consists of four words).  Three word sets are randomly
selected from the top words of one topic whereas the other set is
randomly selected from those of a different topic.\footnote{Each word
  is generated from the top twenty words of a topic based on their
  posterior probability.}  The worker is then asked to identify one
word set that is the most unrelated to other three.

Second, we measure the consistency between human labels and topics
through the modified R4WSI coherency-and-label task
  \citep{Ying2021} where we also show each worker a label from the
topic used to generate the three word sets.  We then ask them to
choose an unrelated word set from the four word
sets.  Using all three empirical applications, we apply both methods
to \keyATM{} and its baseline counterpart and compare their
performance (see Appendix~p.~\pageref{app:validation-design} for details).
We do not include Word Intrusion (WI) and Top 8 Word Set
  Intrusion (T8WSI) in our validation exercise because, as
  \citet{Ying2021} note, both are difficult tasks, and the latter is
  particularly known to be sensitive to the choice of displayed
  words.\footnote{Appendix (p.~\pageref{app:validation-alternative})
  shows results from an alternative design.
   \label{fn:validation-T8WSI}}

We recruit English-speaking workers from Amazon Mechanical Turk for
the base and dynamic applications and Japanese-speaking workers from
CrowedWorks for the covariate application, between December 2020 to
January 2021. Recruited workers are directed to a
Qualtrics survey we designed for this validation. Workers who agree to
participate in this task are asked to complete 11 tasks of the same
kind from a single empirical application: five tasks for \keyATM, five tasks
for the baseline model, and one gold-standard task, which is similar to
the other tasks but is much less ambiguous.\footnote{Some workers
  participate in two sets of 11 tasks with two different methods.} We
use the gold standard task to assess worker's attention and remove the
responses of any worker who fails to provide a correct answer.

We use Bayesian hierarchical logistic regressions to analyze the
validation data for each application.\footnote{
  Appendix (p.~\pageref{app:validation-desc-stats}) provides descriptive
  statistics.
} Since the number of observations
is small for any given topic, partial pooling helps improve the
precision of estimates.\footnote{In the case of base models, for
  example, the average number of observations for each topic is 25
  while the total number of observations is 545 for each task.} The
outcome is an indicator variable $Y_i$, which equals one if a worker
correctly answers the task. Our predictor is the indicator variable
$X_i$, which equals one if the task is based on the outputs of
\keyATM. We also include worker-specific and topic-specific random
intercepts as well as random coefficients for each topic to account
for topic heterogeneity. Formally, for each task conducted by worker
$j$, each response $i$ with topic $k$ is modeled as, \begin{equation*}
  \Pr(Y_{i} = 1) = \logit^{-1} \bigg( (\beta_0 + \delta_{0k[i]} +
  \gamma_{j[i]}) + (\beta_1 + \delta_{1k[i]}) X_i \bigg),
\end{equation*}
where $(\delta_{0k}, \delta_{1k})^\top \sim \Normal(0, \bSigma)$ and
$\gamma_j \sim \Normal(0, \sigma)$.  Our quantity of interest is the
difference in the predicted probabilities of correct responses between
\keyATM{} and its baseline counterpart.

\subsection{Findings}
\label{sec:validation-result}

\begin{figure}[p]
    \centering (a) Base Model: Legislative Bills\par\medskip
    \includegraphics[width=0.82\linewidth]{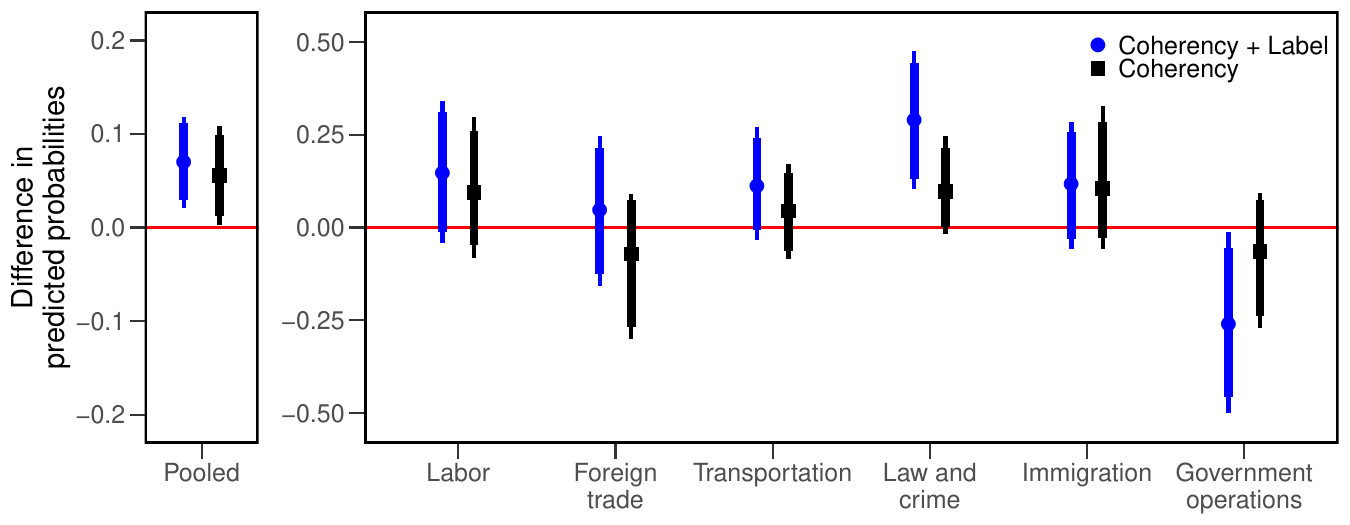}
    \par\medskip
    \centering (b) Covariate Model: Manifestos\par\medskip
    \includegraphics[width=0.82\linewidth]{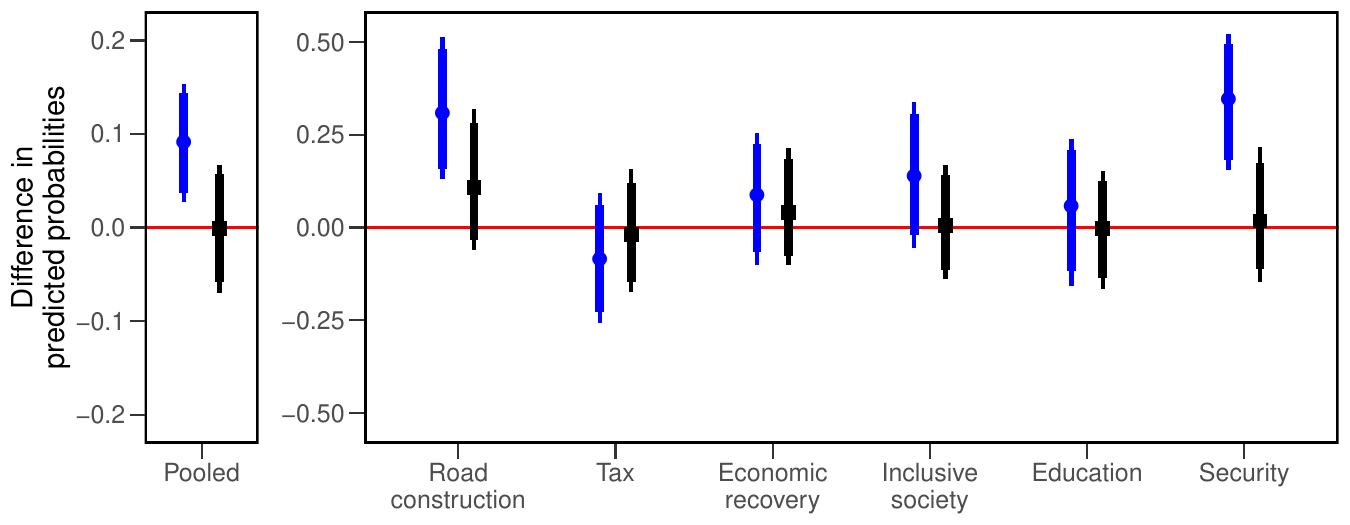}
    \par\medskip
    \centering (c) Dynamic Model: Court Opinions\par\medskip
    \includegraphics[width=0.82\linewidth]{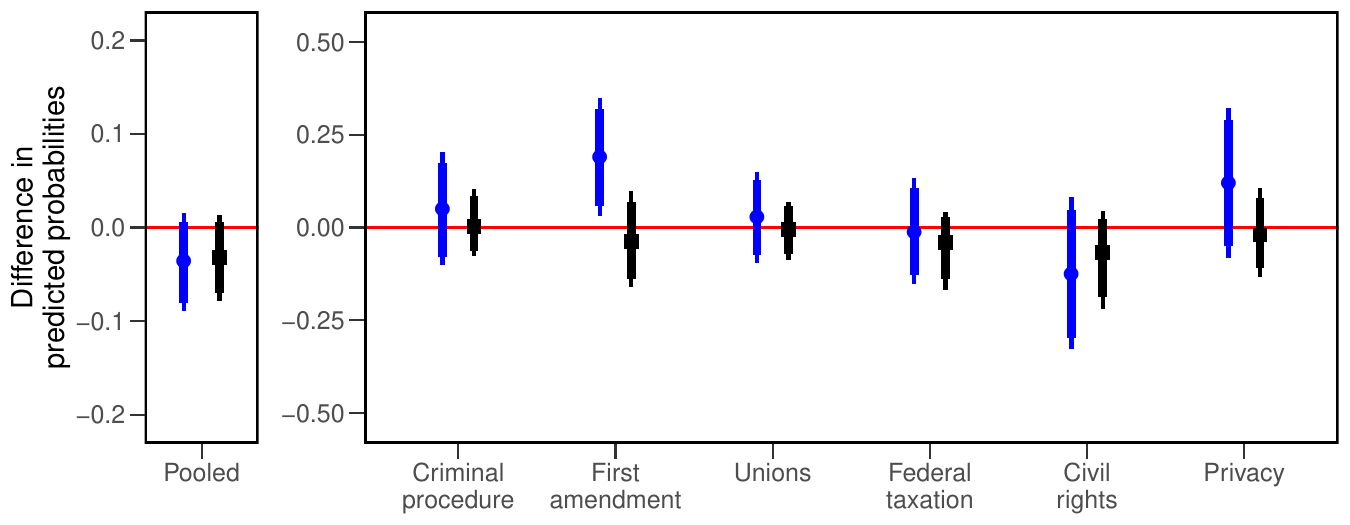}
    \caption{Comparison of the performance of validation
        results between \keyATM{} and its baseline counterpart.}
    \label{fig:validation-main}
    \mynote{Each
      point represents the difference in the predicted probabilities.
      The thick and thin vertical lines indicate the 95\% and 90\%
      credible intervals respectively.  Black and blue
        lines show the results from the coherency task (R4WSI) and the
        coherency-and-label task (modified R4WSI), respectively.  The
      lines in the left panels represent the pooled results whereas
      the lines in the right panels represent the results from six
      topics shown in the previous sections.}
\end{figure}

Figure~\ref{fig:validation-main} presents the main validation results.
We plot the differences in the predicted probabilities based on
Bayesian hierarchical logistic regressions for pooled effects and for
six topics shown in the previous sections (see
Appendix~p.~\pageref{app:validation-topic-by-topic} for the other
topics). First, the pooled estimates show that the base \keyATM{}
performs better in both tasks than \wLDA. The differences in predicted
probabilities are both positive and statistically significant at 95\%
level. Although the topic-specific estimates are less precise, the
base \keyATM{} performs well in the same topics (e.g., \textit{Law and
  crime}) as those identified in
Section~\ref{subsec:base-qualitative-comparison}. On the other hand,
the model performs poorly for the \textit{Government operations},
which is also consistent with the result shown earlier.

Second, the covariate \keyATM{} performs better in the
coherency-and-label task than \STM{}, whereas their performance is
similar in terms of topic coherency. The estimated topics from \keyATM{}
match well with the pre-assigned labels.  Finally, there is no
statistically meaningful difference between the performance of the
dynamic \keyATM{} and \wLDA. This may be because the court opinions
involve technical vocabularies and complex concepts and hence they
pose significant evaluation challenges to crowdsource workers
\citep{Ying2021}. For example, connecting the label \textit{Civil
  rights} to such words as ``district,'' ``school,'' and
``discrimination'' requires some background knowledge.

\section{Keyword Selection}\label{sec:keyword-selection}

The performance of \keyATM{} critically depends on the quality of
keywords.  This section briefly discusses how one might
  select keywords for \keyATM. In
Section~\ref{subsubsec:keyword-construction}, we have shown how to
construct keywords in the context of specific empirical application.
In general, \keyATM{} results in poor topic
  interpretability and classification when selected keywords do not
  frequently occur in one's corpus or do not discriminate their topics
  from others (see Figure~\ref{fig:base-keys} on Appendix~p.~\pageref{fig:base-keys}).
  We also conduct an experiment to
  examine the performance of \keyATM{} by randomly removing some
  keywords.  Our analysis shows that, although \keyATM{} outperforms
  \wLDA{} even when some keywords are removed, the choice of keyword
  matters when the number of keywords is extremely small.  For
  example, the \textit{Immigration} topic in the first application,
   which only contains three
  keywords, performs poorly when the word ``immigration'' is removed
  from the keyword set.  Thus, researchers should examine the relative
  frequency of candidate keywords and choose the ones that
  substantively match with the topics of interest.  See
  Appendix (p.~\pageref{app:base-random-keywords}) for the full set of results.

  Given the importance of keyword selection, future research should
  study how to choose good keywords. For example, \citet{King2017}
  show the promising performance of automated keyword selection
  algorithms, whereas \citet{Watanabe2020} propose a dictionary-making
  procedure based on the average frequency entropy.

\section{Concluding Remarks}

Social scientists have utilized fully automated content analysis based
on probabilistic topic models to measure a variety of concepts from
textual data. To improve the quality of measurement, we propose the
keyword assisted topic models (\keyATM), which require researchers to
label topics of interest before fitting a model.  We have empirically
shown that providing standard models with a small number of keywords
can substantially improve the interpretability and classification
performance of the resulting topics. There are several
  potential extensions of \keyATM.  For example, researchers can
  incorporate a small amount of human-coded labels assigned to each
  document to further improve the performance of \keyATM.  If such
  labels are based on single topics (as is the case for the CBP and
  SCD), this approach yields a mixed-membership version of document
  labels so that each document can belong to multiple topics.

\clearpage
\pdfbookmark[1]{References}{References}
\bibliographystyle{pa}
\bibliography{ref,imai,my}

%
%

\newpage
\spacingset{1.12}
\renewcommand {\theequation} {S\arabic{equation}}
\renewcommand {\thefigure} {S\arabic{figure}}
\renewcommand {\thetable} {S\arabic{table}}

\setcounter{figure}{0}
\setcounter{table}{0}
\setcounter{equation}{0}

\begin{center}
  \Huge{\textbf{Supplementary Information for ``Keyword
  Assisted Topic Models''}}
\end{center}

\floatsetup[figure]{capposition=bottom}
\floatsetup[table]{capposition=bottom}

\begin{appendices}

\section{Graphical Representation of \keyATM}
\label{app:graphs}

\begin{figure}[h!]
  \spacingset{1}
\centering
\begin{tikzpicture}
  \node[latent] (alpha) {$\balpha$};
  \node[latent, left= of alpha] (eta) {$\boldeta$};
  \node[latent, right=3.5 of alpha] (theta) {$\btheta$};
  \node[latent, below=1.5 of theta] (z) {$\bz$};
  \node[latent, left= of z] (s) {$\bs$};
  \node[latent, below= of alpha] (phi-r) {$\bphi$};
  \node[latent, left= of phi-r] (beta) {$\bbeta$};
  \node[latent, below =1.2  of phi-r] (pi) {$\bpi$};
  \node[latent, below= of pi] (phi-s) {$\wt{\bphi}$};
  \node[latent, left= of pi] (gamma) {$\bgamma$};
\node[latent, left= of phi-s] (beta-s) {$\wt{\bbeta}$};
  \node[obs, below=1.5 of z, xshift=-0.9cm] (w) {$w$};
  \edge{alpha}{theta}; \edge{theta}{z}; \edge{z}{w}; \edge{z}{s};
  \edge{gamma}{pi}; \edge{pi}{s};
  \edge{s}{w}; \edge{phi-s}{w}; \edge{phi-r}{w};
  \edge{beta}{phi-r};
  \edge{beta-s}{phi-s};
  \edge{eta}{alpha};
  \plate[inner sep=10pt, xshift=0cm, yshift=0cm]{K}{
    (alpha)(phi-r)
  }{$K$};
  \plate[inner sep=10pt]{Nd}{
    (s)(z)(w)
  }{$N_d$}
  \plate[inner sep=12pt, xshift=-0.08cm]{D}{
    (Nd)(theta)
  }{$D$}
  \plate[inner sep=10pt, xshift=0cm]{Ks}{
    (pi)(phi-s)
  }{$\wt{K}$}
\end{tikzpicture}
\caption{\textbf{Graphical model of the base \keyATM}: The shaded node
  ($w$) represents the observed variable while transparent nodes
  denote latent variables.} \label{fig:graph-base}
\end{figure}

\begin{figure}[h]
\centering
\begin{tikzpicture}
  \node[latent] (lambda) {$\blambda$};
  \node[latent, left = of lambda, yshift = .5cm](mu) {$\mu$};
  \node[latent, left = of lambda, yshift = -.5cm](sigma) {$\sigma$};
  \node[latent, right=3.5 of lambda] (theta) {$\btheta$};
  \node[obs, above= of theta] (x) {$\bx$};
  \node[latent, below=1.5 of theta] (z) {$\bz$};
  \node[latent, left= of z] (s) {$\bs$};
  \node[latent, below= of lambda] (phi-r) {$\bphi$};
  \node[latent, left= of phi-r] (beta) {$\bbeta$};
  \node[latent, below =1.2  of phi-r] (pi) {$\bpi$};
  \node[latent, below= of pi] (phi-s) {$\wt{\bphi}$};
  \node[latent, left= of pi] (gamma) {$\bgamma$};
  \node[latent, left= of phi-s] (beta-s) {$\wt{\bbeta}$};
  \node[obs, below=1.5 of z, xshift=-0.9cm] (w) {$w$};
  \edge{lambda}{theta}; \edge{x}{theta}; \edge{sigma}{lambda}; \edge{mu}{lambda};
  \edge{theta}{z}; \edge{z}{w}; \edge{z}{s};
  \edge{gamma}{pi}; \edge{pi}{s};
  \edge{s}{w}; \edge{phi-s}{w}; \edge{phi-r}{w};
  \edge{beta}{phi-r};
  \edge{beta-s}{phi-s};
  \plate[inner sep=10pt, xshift=0cm, yshift=0cm]{K}{
    (lambda)(phi-r)
  }{$K$};
  \plate[inner sep=10pt]{Nd}{
    (s)(z)(w)
  }{$N_d$}
  \plate[inner sep=12pt, xshift=-0.08cm]{D}{
    (Nd)(theta)(x)
  }{$D$}
  \plate[inner sep=10pt, xshift=0cm]{Ks}{
    (pi)(phi-s)
  }{$\wt{K}$}
\end{tikzpicture}
\caption{\textbf{Graphical model of the covariate \keyATM{}}: The shaded nodes ($w$ and $\bx$) are observed variables while other transparent nodes stand for latent variables.} \label{fig:graph-cov}
\end{figure}

\begin{figure}[!h]
\centering
\begin{tikzpicture}
  \node[latent] (alpha) {$\boldsymbol{\alpha}$};
  \node[latent, left= of alpha] (eta) {$\boldsymbol{\eta}$};
  \node[latent, right=3.5 of alpha] (theta) {$\btheta$};
  \node[latent, above = of theta] (h) {$h_{t[d]}$};
  \node[latent, above = of alpha] (P) {$\mathbf{P}$};
  \node[latent, below=1.5 of theta] (z) {$\bz$};
  \node[latent, left= of z] (s) {$\bs$};
  \node[latent, below=1.3 of alpha] (phi-r) {$\bphi$};
  \node[latent, left= of phi-r] (beta) {$\bbeta$};
  \node[latent, below =1.2  of phi-r] (pi) {$\bpi$};
  \node[latent, below= of pi] (phi-s) {$\wt{\bphi}$};
  \node[latent, left= of pi] (gamma) {$\bgamma$};
\node[latent, left= of phi-s] (beta-s) {$\wt{\bbeta}$};
  \node[obs, below=1.5 of z, xshift=-0.9cm] (w) {$w$};
  \edge{P}{h}; \edge{h}{theta}; \edge{eta}{alpha};
  \edge{alpha}{theta}; \edge{theta}{z}; \edge{z}{w}; \edge{z}{s};
  \edge{gamma}{pi}; \edge{pi}{s};
  \edge{s}{w}; \edge{phi-s}{w}; \edge{phi-r}{w};
  \edge{beta}{phi-r};
  \edge{beta-s}{phi-s};
  \plate[inner sep=8pt, xshift=0cm, yshift=0cm]{R}{
    (alpha)
  }{$R$};
  \plate[inner sep=10pt, xshift=0cm, yshift=0cm]{K}{
    (phi-r)
  }{$K$};
  \plate[inner sep=10pt, xshift=0cm, yshift=0cm]{K}{
    (phi-r)
  }{$K$};
  \plate[inner sep=10pt]{Nd}{
    (s)(z)(w)
  }{$N_d$}
  \plate[inner sep=12pt, xshift=-0.08cm]{D}{
    (Nd)(theta)(h)
  }{$D$}
  \plate[inner sep=10pt, xshift=0cm]{Ks}{
    (pi)(phi-s)
  }{$\wt{K}$}
\end{tikzpicture}
\caption{\textbf{Graphical model of the dynamic \keyATM{}}: The shaded node ($w$) is observed variables while other transparent nodes stand for latent variables.} \label{fig:graph-dynamic}
\end{figure}

\clearpage

\section{Full Posterior Collapsed Distribution of the base \keyATM{}}

The collapsed posterior distribution is,
\begin{align}
  &\iiint p(\bw, \btheta, \bz, \bs, \bphi, \wt{\bphi}, \bpi | \balpha, \bbeta, \wt{\bbeta}, \bgamma, \boldeta) d\bphi d\wt{\bphi} d\bpi d\btheta\\
&= \iint p(\bw | \bz, \bs, \bphi, \wt{\bphi}) p(\wt{\bphi} | \wt{\bbeta}) p(\bphi | \bbeta)  d\wt{\bphi} d\bphi \cdot %
\int p(\bs | \bpi, \bz) p(\bpi | \bgamma) d\bpi \cdot %
\int p(\bz | \btheta) p(\btheta | \balpha) d\btheta \cdot p(\balpha| \boldeta) \\
  &=  \prod_{k=1}^{K} \left[  \dfrac{ \Gamma \left(\sum_{v \in \cV_k} \wt{\beta}_v \right)}{  \prod_{v \in \cV_k} \Gamma\left(\wt{\beta}_v \right)}%
    \dfrac{ \prod_{v \in \cV_k} \Gamma\left(\wt{\beta}_v + \tilde{n}_{kv} \right)}{ \Gamma \left( \sum_{v \in \cV_k} \left( \wt{\beta}_v + \tilde{n}_{kv} \right) \right) }  \cdot  %
\dfrac{  \Gamma \left(\sum_{v=1}^V \beta_v \right)}{ \prod_{v=1}^V \Gamma\left(\beta_v \right)}  %
\dfrac{ \prod_{v=1}^V \Gamma\left(\beta_v + n_{kv} \right)}{ \Gamma \left(\sum_{v=1}^V \left( \beta_v + n_{kv} \right) \right) }  \right] \\[10pt]
  &\quad\times \left( \dfrac{\Gamma(\gamma_1 + \gamma_2)}{\Gamma(\gamma_1) \Gamma(\gamma_2)} \right)^K \prod_{k=1}^{K} \left[ \dfrac{\Gamma \left(\tilde{n}_{k} + \gamma_1 \right) \Gamma \left(n_{k} + \gamma_2 \right)}{\Gamma \left(\tilde{n}_{k} + \gamma_1 + n_{k} + \gamma_2 \right)}  \right] \\[10pt]
	&\quad \times \prod_{d=1}^D \left[  \dfrac{\Gamma\left( \sum_{k=1}^K \alpha_k \right)}{ \prod_{k=1}^K \Gamma(\alpha_k)} \dfrac{ \prod_{k=1}^K \Gamma \left(n_{dk} + \alpha_k \right)}{\Gamma \left( \sum_{k=1}^K \left( n_{dk} + \alpha_k \right) \right)} \right]
	\times \prod_{k=1}^{K} \left\{ \dfrac{\eta_{2}^{\eta_{1}} }{ \Gamma(\eta_{1})}  \cdot \alpha_{k}^{\eta_{1} - 1} \cdot \exp(- \eta_{2} \cdot \alpha_{k} ) \right\} \label{cpd}
\end{align}
where
\begin{align}
  n_{kv} &= \sum_{d=1}^D \sum_{i=1}^{Nd} \mathbbm{1}(w_{di}=v) \mathbbm{1}(s_{di}=0)\mathbbm{1}(z_{di}=k), \\
  \tilde{n}_{kv} &= \sum_{d=1}^D \sum_{i=1}^{Nd} \mathbbm{1}(w_{di}=v) \mathbbm{1}(s_{di}=1)\mathbbm{1}(z_{di}=k),\\
  n_{dk} &= \sum_{i=1}^{Nd} \mathbbm{1}(z_{di}=k).
\end{align}

\section{Sampling Algorithm and Model Interpretation}

\subsection{The Covariate \keyATM{}}\label{app:sampling-cov}

\subsubsection{Sampling Algorithm}

The collapsed Gibbs sampler for the covariate \keyATM{} is also
identical to that of the base \keyATM{} except for a couple of steps.
First, we sample the topic assignment for each word $i$ in document
$d$ from the following conditional posterior distribution that
incorporates the covariates (instead of
Equation~\eqref{eq:sample-z-base}),
\begin{align}
  &\quad \Pr(z_{di}=k \mid \zdel, \bw, \bs, \blambda, \bX, \bbeta,
    \wt{\bbeta}, \bgamma) \nonumber\\
&\quad\propto
\begin{cases} %
  \frac{ \beta_v + n_{k v}^{- di} }{   \sum_v \beta_v +  n_{k}^{- di}} \cdot %
  \frac{ n^{- di}_{k} + \gamma_2 }{ \tilde{n}_{k}^{- di} + \gamma_1 + n^{- di}_{k} + \gamma_2 } \cdot %
  \left(n_{d{k}}^{- di} + \exp(\blambda_{k}^\top \bx_{d})  \right)  & \ {\rm if \ } s_{di} = 0, \\
  \frac{ \tilde{\beta}_v + \tilde{n}_{k v}^{- di}    }{ \sum_{v \in \cV_k} \tilde{\beta}_v + \tilde{n}_{k}^{- di}  } \cdot%
    \frac{ \tilde{n}^{ - di}_{k} + \gamma_1 }{ \tilde{n}^{- di}_{k} + \gamma_1 + n^{- di}_{k} + \gamma_2 } \cdot %
\left(n_{d{k}}^{- di} + \exp(\blambda_{k}^\top \bx_{d})  \right) & \ {\rm if \ } s_{di} = 1.
\end{cases}\label{eq:sample-z-cov}
\end{align}
In addition, we need to sample $\blambda$ instead of $\balpha$ (see
equation~\eqref{eq:alpha_base_sample}). As before, we use the
unbounded slice sampler \citep{Mochihashi2020} and sample each
parameter $\lambda_{mk}$ based on the following conditional posterior
distribution,
\begin{align*}
 p(\lambda_{mk} \mid \blambda_{-[mk]}, \bX, \mu, \sigma^2) \ \propto \
  \prod_{d=1}^D \left[  \frac{\Gamma\left(\sum_{k=1}^K
  \exp(\blambda_{k}^\top \bx_{d}) \right)}{
  \Gamma(\exp(\blambda_{k}^\top \bx_{d}))} \frac{ \Gamma \left(n_{dk}
  + \exp(\blambda_{k}^\top \bx_{d}) \right)}{ \Gamma\left( \sum_{k=1}^K n_{dk} + \exp(\blambda_{k}^\top \bx_{d}) \right)} \right] %
 \exp \left( \frac{-( \lambda_{mk} - \mu)^2}{2 \sigma^2} \right),
\end{align*}
where $\blambda_{-[mk]}$ is $\blambda$ without its $(m,k)$ element.

The model has $K$ topics, $D$ documents, and $M$ covariates. The
covariate matrix is given by $\mathbf{X}$ whose $d$th column is
denoted by $\mathbf{x}_d$, which is an $M \times 1$ dimensional
covariate vector.  In addition, $\boldsymbol{\lambda}$ is an
$M \times K$ dimensional matrix for coefficients.  In the estimation,
we use standardized covariates, which is defined as
\begin{align}
  \tilde{\mathbf{x}}_{m} &= \frac{\mathbf{x}_m - \bar{\mathbf{x}}_m}{\text{SD}(\mathbf{x}_m)}.
\end{align}
where $\text{SD}(\mathbf{x}_m)$ represents the standard deviation of
$\mathbf{x}_m$.  We can write the standardization of $\mathbf{X}$ in
the following matrix form,
\begin{align}
  \tilde{\mathbf{X}} &= \bigg(\mathbf{I}_{D} - \frac{1}{D}\mathbf{1}_{D} \mathbf{1}_{D}^\top \bigg) \mathbf{X} \mathbf{W},
\end{align}
where $\mathbf{I}_{D} - \frac{1}{D} \mathbf{1}_{D} \mathbf{1}_{D}^\top$ is a $D \times D$ matrix to demean $\mathbf{X}$, and $\mathbf{W}$ is a $M \times M$ scaling matrix whose diagonal elements are the inverse of $m^{\text{th}}$ covariate's  standard deviation,
\begin{align}
  \mathbf{W} =
  \begin{bmatrix}
    \frac{1}{\text{SD}(\mathbf{x}_1)} & 0 & \ldots & 0 \\
    0 & \frac{1}{\text{SD}(\mathbf{x}_2)} & \ldots & 0 \\
    \vdots & 0 & \ddots & \vdots \\
    0 & \ldots & 0 & \frac{1}{\text{SD}(\mathbf{x}_M)}
  \end{bmatrix}.
\end{align}

When we use standardized covariates, the model for $\btheta_d$ becomes
$\btheta_{d} \sim \Dir (\exp( \tilde{\blambda}^{\top}
\tilde{\mathbf{x}}_{d} ))$. And yet, we want to compute $\blambda$ for
raw covariates $\mathbf{x}_d$. The linear transformation of
$\mathbf{x}$ will be reflected in $\tilde{\blambda}$, because we do
not transform the assignment counts of $\mathbf{z}$ and the only
difference between the likelihood of $\blambda$ and $\tilde{\blambda}$
is whether or not the covariates are standardized, hence
\begin{align}
  \mathbf{X}\blambda &= \tilde{\mathbf{X}}\tilde{\blambda}.
\end{align}
Now, solve it for $\blambda$ to obtain,
\begin{align}
  &\blambda = (\mathbf{X}^\top \mathbf{X})^{-1} \mathbf{X}^\top \tilde{\mathbf{X}}\tilde{\blambda}.
\end{align}
Therefore, \keyATM{} can store $\tilde{\blambda}$ in each
iteration and rescale it to yield $\blambda$.  Using $\blambda$, we
wish to obtain a posterior predictive distribution of $\theta_d$ given
a new covariate data set $\mathbf{X}=\mathbf{x}_d^\ast$.  Then, the posterior
predictive distribution is,
\begin{align}
 p(\theta_d^\ast \mid \mathbf{x}_d^\ast, \bw)
& \ = \  \int p(\theta_d^\ast\mid \mathbf{x}_{d}^\ast,
                                                 \boldsymbol{\lambda})
                                                 p(\boldsymbol{\lambda}
                                                 \mid \mu, \sigma^2, \bw) d \boldsymbol{\lambda}.
\end{align}
We can compute the posterior predictive distribution of the mean of
$\theta_d^\ast$ given $\bx_d^\ast$
\begin{align}
\int \frac{\exp( \lambda_k^\top \bx_d^\ast) }{\sum_{k^\prime=1}^K
  \exp(\lambda_{k^\prime}^\top \bx_d^\ast)} p(\boldsymbol{\lambda}
                                                 \mid \mu, \sigma^2, \bw) d \boldsymbol{\lambda}.
\end{align}

\subsubsection{Model Interpretation}

The derivation of the topic-word distribution is identical to the one
used for the base \keyATM{} (see Section~\ref{subsec:base-interpret}).
In addition, the covariate \keyATM{} can characterize the relations
between covariates and document-topic distributions, which are
governed by the coefficients $\blambda$. Specifically, we simply
replace $\alpha_k$ with $\exp(\blambda_{k}^\top \bx_d)$ in the
Equation~\eqref{eq:base_Etheta} to obtain the marginal posterior mean
of $\theta_{dk}$,
\begin{equation}
  \E \left (\theta_{dk} \mid  \bx_d, \bw \right) \ = \
  \E \left(\frac{\exp(\blambda_{k}^\top \bx_d) +
      n_{dk}}{\sum_{\kprime=1}^{K} \exp(\blambda_{\kprime}^\top \bx_d)
      + n_{d \kprime} } \,\middle\vert\, \bx_d, \bw \right). \label{eq:cov_Etheta}
\end{equation}
We can also obtain the predicted topic proportion $\theta_{dk}$ by
setting $\bx_d$ to specific values and computing the posterior
distribution of its mean given the new covariate value.

Finally, although such an extension is possible, the covariate
\keyATM{} does not directly model the topic-word distributions.
Nevertheless, it is possible to examine how the topic-word
distributions change across different values of document-level
covariates by simply computing Equation~\eqref{eq:base_Ephi} with
a set of documents that share the same values of
document-level covariates.

\subsection{The Dynamic \keyATM{}}\label{app:sampling-dyn}

\subsubsection{Sampling Algorithm}

For fitting the dynamic \keyATM, we add two steps to the sampling
algorithm described in Section~\ref{sec:sampling-base} (i.e., sampling
of latent states and transition probabilities) and change how to
sample $\balpha$.  To sample the latent state membership $\bh_{1:T}$,
we use a well-known forward-and-backward sampling procedure.  Under
the current setup, joint density of $h$ from time $1$ to $T$ can be
factorized as \citep{Chib1996},
\begin{eqnarray*}
  p(\bh_{1:T} \mid \bz, \balpha, \bP) \ = \ p(h_T \mid \bz_T, \balpha, \bP) \times \cdots \times p(h_t \mid \bh_{t+1:T}, \bz_t, \balpha, \bP) \times \cdots \times p(h_1 \mid \bh_{2:T}, \bz_1, \balpha,  \bP).
\end{eqnarray*}
where $\bz_t$ represents a vector of the topic assignments for all
documents that belong to time period $t$.  Thus, to sample from the
joint distribution, we must sample latent states backwards in time,
starting from $h_T$.  Since we know $h_T = R$ and $h_1 = 1$, we sample
each of the remaining states from the following conditional
distribution,
\begin{eqnarray}
  p(h_{t} \mid \bh_{t+1:T}, \bz_{t},  \balpha, \bP) \ \propto \ p(h_{t} \mid \bz_t,
   \balpha, \bP)  p(h_{t+1} \mid h_{t}, \bP), \label{eq:dynamic-backward}
\end{eqnarray}
where the second term is given in the transition matrix $\bP$.  Each
term in equation~\eqref{eq:dynamic-backward} is obtained using the
following recursion formula \citep[p.227]{Chib1998}, moving forward in
time,
\begin{equation*}
  \Pr(h_{t} = r \mid \bz_{t}, \balpha, \bP ) \ = \ \frac{\Pr(h_{t} = r \mid \bz_{t-1}, \balpha, \bP)  p(\bz_{t} \mid \balpha_{r})}{\sum_{l=r-1}^{r} \Pr(h_{t} = l \mid \bz_{t-1}, \balpha, \bP)  p(\bz_{t} \mid \balpha_{l}) },
\end{equation*}
where
\begin{equation*}
  \Pr(h_{t} = r \mid \bz_{t-1}, \balpha_r, \bP) \ = \ \sum_{l = r - 1}^{r} p_{lr} \cdot \Pr(h_{t-1} = l \mid \bz_{t-1}, \balpha, \bP),
\end{equation*}
which only depends on the information at time $t-1$.  At last, to
compute $p(\bz_t \mid \balpha_r)$, we integrate out $\btheta_d$ with
respect to a state-specific prior $\balpha_r$, yielding a
Dirichlet-Multinomial distribution,
\begin{align*}
   p(\bz_{d} \mid \balpha_{h_{t[d]}},  h_{t[d]})  \ = \ \int  p(\bz_{d}
  \mid \btheta_{d}) p(\btheta_{d} \mid \balpha_{h_{t[d]}}, h_{t[d]}) d \btheta_{d}.
\end{align*}

To sample the transition matrix $\bP$, let $n_{rr}$ be the number of
transitions from state $r$ to state $r$ in the all sequences of latent
state memberships $\bh$. Then,
we have,
\begin{equation*}
  p_{rr} \mid \bh \ \sim \ \Beta(1 + n_{rr}, 1 + n_{r, r+1}),\quad \text{for} \ r = 1,\ldots, R,
\end{equation*}
where $n_{r, r+1} = 1$, because the state always moves forward.

Finally, we sample $\balpha_{r}$ from the following conditional
distribution of $\balpha_{r}$ using the slice sampling procedure
described in \citet{Mochihashi2020},
\begin{eqnarray*}
  p(\alpha_{rk} \mid \alpha_{-[rk]}, \bz, \bh, \boldsymbol{\eta})
  &  \propto &
               \ \alpha_{rk}^{\eta_{1} - 1} \exp(- \eta_{2} \alpha_{rk} ) \prod_{d=1}^{N_d} \left[  \dfrac{ \Gamma\left(\sum_{k=1}^K \alpha_{rk} \right) \Gamma \left(n_{dk} + \alpha_{rk} \right)}{\Gamma(\alpha_{rk}) \Gamma \left( \sum_{k=1}^K  n_{dk} + \alpha_{rk} \right)} \right]^{\mathbbm{1}(h_{t[d]}=r)}
\end{eqnarray*}
for $k=1,2,\ldots,\wt{K}$.  For $k=\wt{K}+1,\ldots,K$, the conditional
distribution is identical except that $\tilde\eta_1$ and
$\tilde\eta_2$ are replaced with $\eta_1$ and $\eta_2$.

\subsubsection{Model Interpretation}\label{app:dyn-model-interpretation}

The topic-word distribution can be computed as before using
Equation~\eqref{eq:base_Ephi}.  To understand the dynamic trend of
topic proportions, we first compute the posterior mean of topic
proportion for each document $d$ as,
\begin{equation}
  \E(\theta_{dk} \mid \bw) \ = \ \E \left[ \dfrac{\alpha_{h_{t[d]}, k} + n_{dk}}{\sum_{k=1}^{K} \alpha_{h_{t[d]}, k} + n_{dk} } \ \biggl | \ \bw  \right], \label{eq:dyn-Etheta}
\end{equation}
for topic $k$.  Then, to examine the resulting time trends, we simply
compute the sample average of this quantity across all documents that
belong to each time point $t$,
\begin{equation}
  \dfrac{1}{N_{t}} \sum_{d \in \mathcal{D}_t} \E(\theta_{d k} \mid \bw) \label{eq:dyn-estimated-trend}
\end{equation}
where $\mathcal{D}_t$ represents the set of documents for time $t$ and
$N_{t}=|\mathcal{D}_t|$ is the number of documents for time $t$.   We
can then plot this quantity to visualize the time trend of prevalence
for each topic.

\subsection{Keywords as Informative
  Prior}\label{app:keywords-as-prior}

The incorporation of keywords in \keyATM{} imposes an
  informative prior on the structure of topic-word distribution.
  Here, we directly connect our approach to the existing methods.
  \citet{Lu2011} add a pseudo counts $C_v$ to a keyword $v$ via,
  $$\bphi_k \sim \Dir(\{ \beta + C_v \}_{v \in V})$$
  where $C_v > 0$ if $v$ is a keyword and $C_v = 0$ otherwise.  Then,
  the mean of the topic-word distribution for a term $v$ in a topic
  $k$ is given by
  $\E [\phi_{k v}] = (\beta + C_v) / (\beta V + \sum_{v'} C_{v'})$.
  Since $C_v = 0$ if $v$ is not a keyword, a keyword of a topic is
  more likely to appear in the topic compared to other topics on
  average.  In contrast, the prior distribution used in
  \keyATM{} 
  implies that
  $\E[\phi_{kv}^{\text{keyATM}}] > \E[\phi_{kv}^{\text{LDA}}]$ if a
  $v$ is a keyword and
  $\E[\phi_{kv}^{\text{keyATM}}] < \E[\phi_{kv}^{\text{LDA}}]$
  otherwise where $\phi_{kv}^{\text{keyATM}}$ and
  $\phi_{kv}^{\text{LDA}}$ represent the topic-word distribution under
  \keyATM{} and \LDA{}, respectively.  Therefore, in terms of the
  expectation of prior probabilities, \keyATM{} is similar to the
  model of \citet{Lu2011}.  In addition, \citet[p.215]{Fan2019} take
  a similar approach by multiplying the prior $\beta_v$ with a scaling
  factor $c$ if $v$ is a keyword.

The important difference between the \keyATM{} and these two
  existing models is that we do not require researchers to specify the
  values of the hyper-parameters, which directly determine the importance of
  keywords. Instead, we place a restriction on the structure of the
  prior distribution by setting no-keyword frequencies exactly to zero
  in the prior distribution of $\tilde\phi_k$ so that the data can
  inform the importance of keywords.

\section{Additional Information for the Base \keyATM}

\subsection{Details of Preprocessing}\label{app:base-preprocessing}

\subsubsection*{Documents}
We preprocess the raw
texts by first removing stop words via the \textsf{R} package
\texttt{quanteda} \citep{Benoit2018}, then pruning words that appear
less than 11 times in the corpus, and lemmatizing the remaining words
via the \textsf{Python} library \texttt{NLTK}
\citep{Bird2009}.
We also use lemmatization.
Removing stop words and low frequent terms, and lemmatizing
  terms help reduce the dimensionality of data
  \citep{Manning2008}. We use the same preprocessing steps for all
  models.


\subsubsection*{CAP Codebook}
First, we remove stopwords and lemmatize the
remaining terms in the same way as done for the bill texts.  We also
remove the words and phrases that have little to do with the
substantive meaning of each topic.  For example, a topic description
always begins with the phrase ``Description: Includes issues
\ldots'', and hence we remove this phrase.  Similarly, we exclude
the phrases, ``\ldots related generally to \ldots'' and ``\ldots
generally related to \ldots,'' which appear in the description of 13
different topics.  Third, we use the same keywords for multiple
topics only if their inclusion can be substantively justified.  For
example, the term ``technology,'' which appears in the description
of several topics, is only used as a keyword for the
\textit{Technology} topic.  Lastly, we limit the number of keywords
to 25 per topic.  We remove terms based on the proportion of
keywords among all terms in the corpus if the topic contains more
than 25 keywords.

\subsection{The Full List of Keywords}\label{app:base-keywords-selection}

\begin{longtable}{p{2cm}p{14cm}}
\hline
Label                 & Keywords\\  \hline
Government operations
& administrative advertising appointment attack auditing branch campaign capital census city coin collection currency mail medal mint nomination post postal registration statistic terrorist victim voter \\ \hdashline
Public lands
& fire flood forest grazing historic indigenous land livestock natural parks recreation resource site staff territorial territorie water \\      \hdashline
Defense
& armed base capability civilian compliance contractor coordination damage equipment foreign homeland installation intelligence material military nuclear operation personnel procurement reserve security services weapon      \\      \hdashline
Domestic commerce
& account accounting bankruptcy business card commerce commercial commodity consumer cost credit disaster finance financial fraud industry insurance investment management mortgage patent promote property relief security     \\      \hdashline
Law and Crime
& abuse border code combat court crime criminal custom cyber drug enforcement family fine judiciary justice juvenile legal penalty police prison release representation sexual terrorism violence  \\      \hdashline
Health                &
abuse alcohol care clinical cost cover coverage disease drug health insurance insurer liability license medical mental pay payment prescription prevention provider rehabilitation supply tobacco treatment \\ \hdashline
International affairs
& aid assessment associate citizen combat committee convention country cross develop directly foreign human international monetary ocean region regional sea target terrorism treaty union world \\  \hdashline
Transportation
& air airport aviation channel construction deployment freight highway infrastructure inland maintenance maritime mass pilot rail railroad ship traffic transportation travel waterway  \\
\hdashline
Macroeconomics        & bank budget budgeting central cost deficit growth industrial inflation interest macroeconomic manufacturing monetary price revitalization tax treasury \\       \hdashline
Environment       &
alternative asbestos chemical climate conservation disposal drinking endanger environment environmental hazardous laboratory performance pollution protection resource solid specie supply toxic waste wastewater water wildlife    \\      \hdashline
Education             & adult area bilingual college education educational elementary excellence handicapped improve language literacy loan mentally need outcome physically primary school schools secondary skill student university vocational \\      \hdashline
Energy     &
alternative biofuel clean coal conservation drilling electrical electricity energy gas gasification gasoline geothermal hydrogen hydropower natural nuclear oil power production renewable shortage spill utility vehicle \\      \hdashline
Technology
& broadcast communication computer cooperation encourage exploration forecast form geological internet publishing radio research satellite science space speed survey technology telecommunication telecommunications telephone television transfer weather \\      \hdashline
Labor                 & bargaining benefit compensation debt employee employer employment fair injury insurance job labor minimum pension protection retirement standard training unemployment union wage work worker workforce youth \\       \hdashline
Foreign trade
& agreement balance barrier competitiveness dispute exchange export foreign import international negotiation private productivity subsidy tariff trade treaty \\      \hdashline
Civil rights          &
abortion age civil contract discrimination ethnic expression franchise freedom gender group information mandatory minority participation party privacy racial religious right rights sex sexual speech voting\\      \hdashline
Social welfare        &
aid alleviate assess assistance association care charity child disability disable elderly family income leave parental physical poverty social volunteer welfare youth \\       \hdashline
Agriculture           & agricultural agriculture animal crop farmer fish fishery food inspection labeling market pest pesticide product rancher seafood subsidy \\      \hdashline
Housing  & affordability community economic family handicap homeless homelessness housing income neighborhood rural urban veteran \\      \hdashline
Immigration           & refugee citizenship immigration\\      \hdashline
Culture               & culture cultural \\ \hline

\caption{\normalsize \textbf{Keywords for the base \keyATM{}}: Keywords used in the base \keyATM{} application.} \label{table:app-base-keywords}
\end{longtable}

\subsection{Top Words}\label{app:base-topwords}
  Table~\ref{tab:app-base-topwords} presents the top ten
  words with the highest estimated probability defined in
  Equation~\eqref{eq:base-phi}.  The results show that \keyATM{} shows
  better performance than \wLDA{} in most topics.

\begin{longtable}{p{2.4cm}p{2.4cm}|p{2.4cm}p{2.4cm}|p{2.4cm}p{2.4cm}}
  \hline   \hline
  \multicolumn{2}{c|}{Government operations} & \multicolumn{2}{c|}{Public lands} & \multicolumn{2}{c}{Defense}\\
  \hline \keyATM{} & \wLDA{} & \keyATM{} & \wLDA{} & \keyATM{} & \wLDA{}\\  \hline
expense & congress & \textbf{land} & land & \textbf{military} & military \\
  appropriation & house & \textbf{water} & management & force & force \\
  remain & senate & area$^*$ & area & member & system \\
  authorize & office & management$^*$ & forest & air$^*$ & support \\
  necessary & committee & river & indian & code$^*$ & authorization \\
  transfer$^*$ & commission & \textbf{resource} & interior & authority & security \\
  expend & representative & \textbf{forest} & resource & authorization & operation \\
  exceed & congressional & authorize & park & \textbf{reserve} & army \\
  office & strike & cost$^*$ & conservation & \textbf{armed} & committee \\
  activity & bill & stat & within & army & air \\
          \hline   \hline
\multicolumn{2}{c|}{Domestic commerce} & \multicolumn{2}{c|}{Law \& crime} & \multicolumn{2}{c}{Health}\\ \hline \keyATM{} & \wLDA{} & \keyATM{} & \wLDA{} & \keyATM{} & \wLDA{}\\ \hline
\textbf{financial} & financial & intelligence$^*$ & security & \textbf{health} & health \\
  institution & institution & attorney & information & \textbf{care} & care \\
  bank$^*$ & bank & \textbf{crime} & intelligence & \textbf{drug} & individual \\
  \textbf{insurance} & company & \textbf{court} & homeland & \textbf{payment} & drug \\
  company & insurance & \textbf{enforcement} & committee & \textbf{medical} & payment \\
  corporation & corporation & \textbf{criminal} & director & individual & medical \\
  board & board & \textbf{code} & system & \textbf{coverage} & describe \\
  \textbf{security} & security & offense & foreign & medicare & respect \\
  \textbf{credit} & credit & person & government & respect & social \\
  commission & commission & \textbf{justice} & office & describe & part \\
      \hline   \hline
\multicolumn{2}{c|}{International affairs} & \multicolumn{2}{c|}{Transportation} & \multicolumn{2}{c}{Macroeconomics}\\ \hline \keyATM{} & \wLDA{} & \keyATM{} & \wLDA{} & \keyATM{} & \wLDA{}\\ \hline
assistance$^*$ & assistance & \textbf{transportation} & transportation & apply & transfer \\
  \textbf{foreign} & foreign & \textbf{highway} & highway & \textbf{tax} & appropriation \\
  \textbf{country} & country & safety & safety & amendment & emergency \\
  \textbf{international} & international & carrier & vehicle & end & operation \\
  government & president & \textbf{air} & carrier & taxable & military \\
  president & government & code$^*$ & motor & strike & construction \\
  development & committee & system & system & relate & procurement \\
  \textbf{committee} & development & vehicle$^*$ & strike & income$^*$ & remain \\
  export$^*$ & export & \textbf{airport} & rail & respect & maintenance \\
  organization & organization & motor & code & case & budget \\
    \hline \hline
\multicolumn{2}{c|}{Environment} & \multicolumn{2}{c|}{Education} & \multicolumn{2}{c}{Energy}\\ \hline \keyATM{} & \wLDA{} & \keyATM{} & \wLDA{} & \keyATM{} & \wLDA{}\\ \hline
committee$^*$ & water & \textbf{education} & school & \textbf{energy} & energy \\
  submit & river & \textbf{school} & education & fuel & fuel \\
  review & cost & \textbf{educational} & educational & \textbf{gas} & gas \\
  later & stat & \textbf{student} & student & facility & change \\
  administrator & authorize & local & child & \textbf{vehicle} & facility \\
  require & carry & institution & local & \textbf{oil} & new \\
  requirement & non & grant & grant & electric & vehicle \\
  develop$^*$ & development & part & part & standard$^*$ & electric \\
  information$^*$ & resource & high & activity & administrator & production \\
  management$^*$ & study & eligible & eligible & \textbf{power} & administrator \\
      \hline \hline
  \multicolumn{2}{c|}{Technology} & \multicolumn{2}{c|}{Labor} & \multicolumn{2}{c}{Foreign trade}\\ \hline \keyATM{} & \wLDA{} & \keyATM{} & \wLDA{} & \keyATM{} & \wLDA{}\\ \hline
\textbf{research} & research & \textbf{employee} & apply & product$^*$ & air \\
  \textbf{technology} & technology & \textbf{benefit} & tax & \textbf{trade} & vessel \\
  business$^*$ & development & individual & amendment & change & airport \\
  information$^*$ & establish & rate & end & \textbf{agreement} & transportation \\
  director & committee & \textbf{compensation} & taxable & good & aviation \\
  system & activity & period & respect & tobacco$^*$ & administrator \\
  small & administrator & code$^*$ & period & head & aircraft \\
  \textbf{science} & system & payment$^*$ & individual & article & carrier \\
  development & information & determine & case & free & administration \\
  center & carry & agreement$^*$ & relate & chapter & coast \\
        \hline   \hline
  \multicolumn{2}{c|}{Civil rights} & \multicolumn{2}{c|}{Social welfare} & \multicolumn{2}{c}{Agriculture}\\ \hline \keyATM{} & \wLDA{} & \keyATM{} & \wLDA{} & \keyATM{} & \wLDA{}\\ \hline
person & person & \textbf{child} & expense & \textbf{food} & food \\
  action & action & \textbf{assistance} & appropriation & \textbf{agricultural} & agricultural \\
  \textbf{information} & regulation & individual & authorize & loan$^*$ & loan \\
  order & require & grant & remain & \textbf{agriculture} & agriculture \\
  court$^*$ & information & \textbf{family} & necessary & farm & payment \\
  regulation & court & \textbf{disability} & expend & producer & farm \\
  commission & order & strike & office & payment$^*$ & producer \\
  require & rule & indian & exceed & rural$^*$ & crop \\
  rule & commission & receive & transfer & \textbf{crop} & rural \\
  provision & review & support & activity & commodity$^*$ & commodity \\
  \hline \hline
  \multicolumn{2}{c|}{Housing} & \multicolumn{2}{c|}{Immigration} & \multicolumn{2}{c}{Culture}\\ \hline \keyATM{} & \wLDA{} & \keyATM{} & \wLDA{} & \keyATM{} & \wLDA{}\\ \hline
\textbf{housing} & housing & security$^*$ & alien & congress & member \\
  loan$^*$ & assistance & alien & attorney & house & strike \\
  assistance$^*$ & loan & \textbf{immigration} & child & senate & code \\
  development & development & homeland$^*$ & crime & office & force \\
  \textbf{community} & family & border$^*$ & immigration & committee$^*$ & pay \\
  mortgage$^*$ & mortgage & status & grant & representative & military \\
  \textbf{family} & community & nationality & enforcement & member & officer \\
  \textbf{income} & insurance & describe & person & strike & authorize \\
  insurance$^*$ & income & individual & court & government & duty \\
  unit & unit & employer$^*$ & offense & congressional & reserve \\
      \hline
    \hline
    \caption{\normalsize \textbf{Top words of the base \keyATM.} The
  table shows the top ten words with the highest estimated probability for each topic under each model. For \keyATM, the pre-specified keywords for each topic appear in bold letters whereas the asterisks indicate the keywords specified for another topic.}
\label{tab:app-base-topwords}
\end{longtable}

\subsection{ROC Curves}\label{app:base-roc}

\begin{figure}[!ht]
\centering
\includegraphics[width= .9\linewidth]{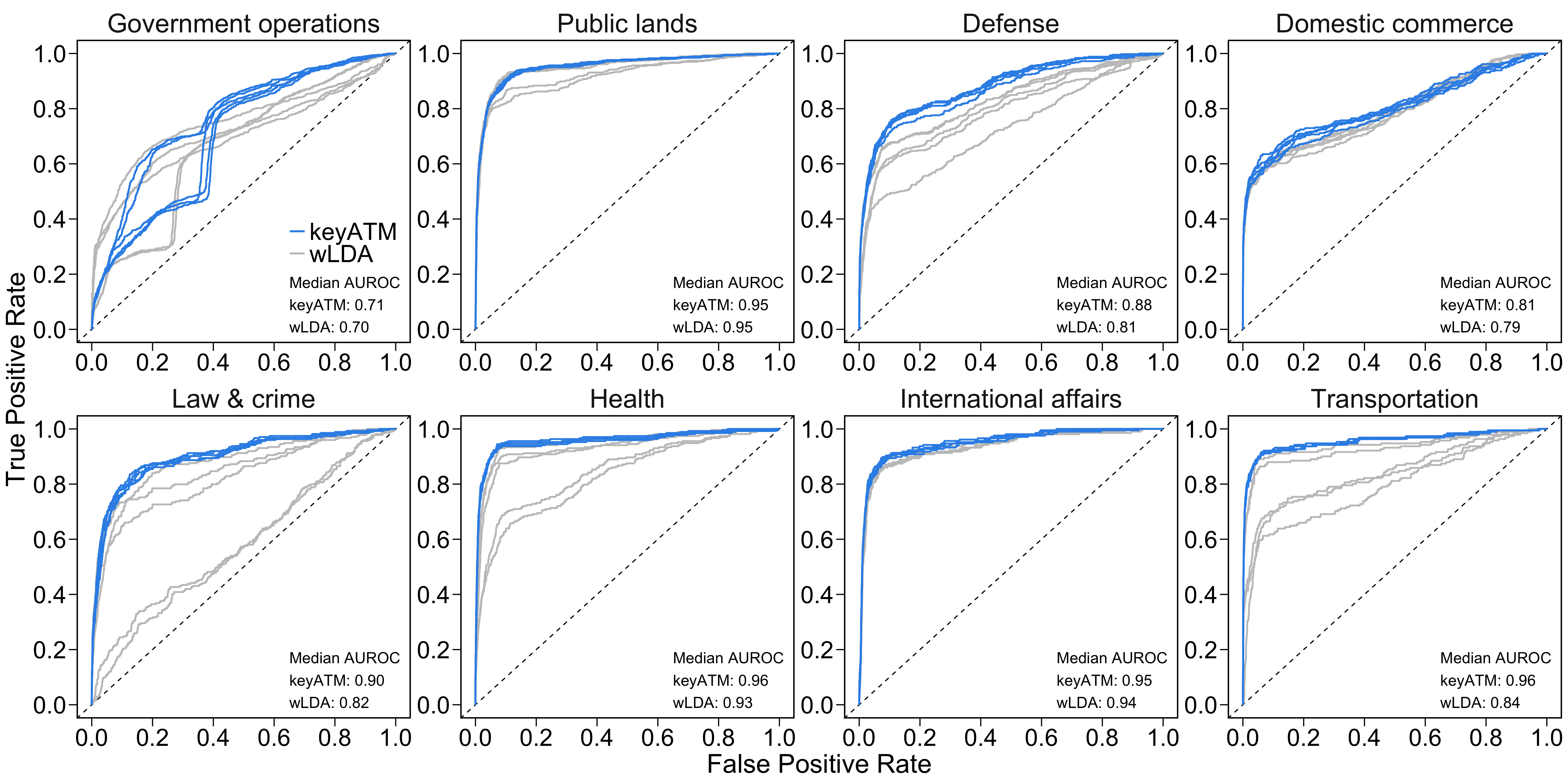}
%
\includegraphics[width= .9\linewidth]{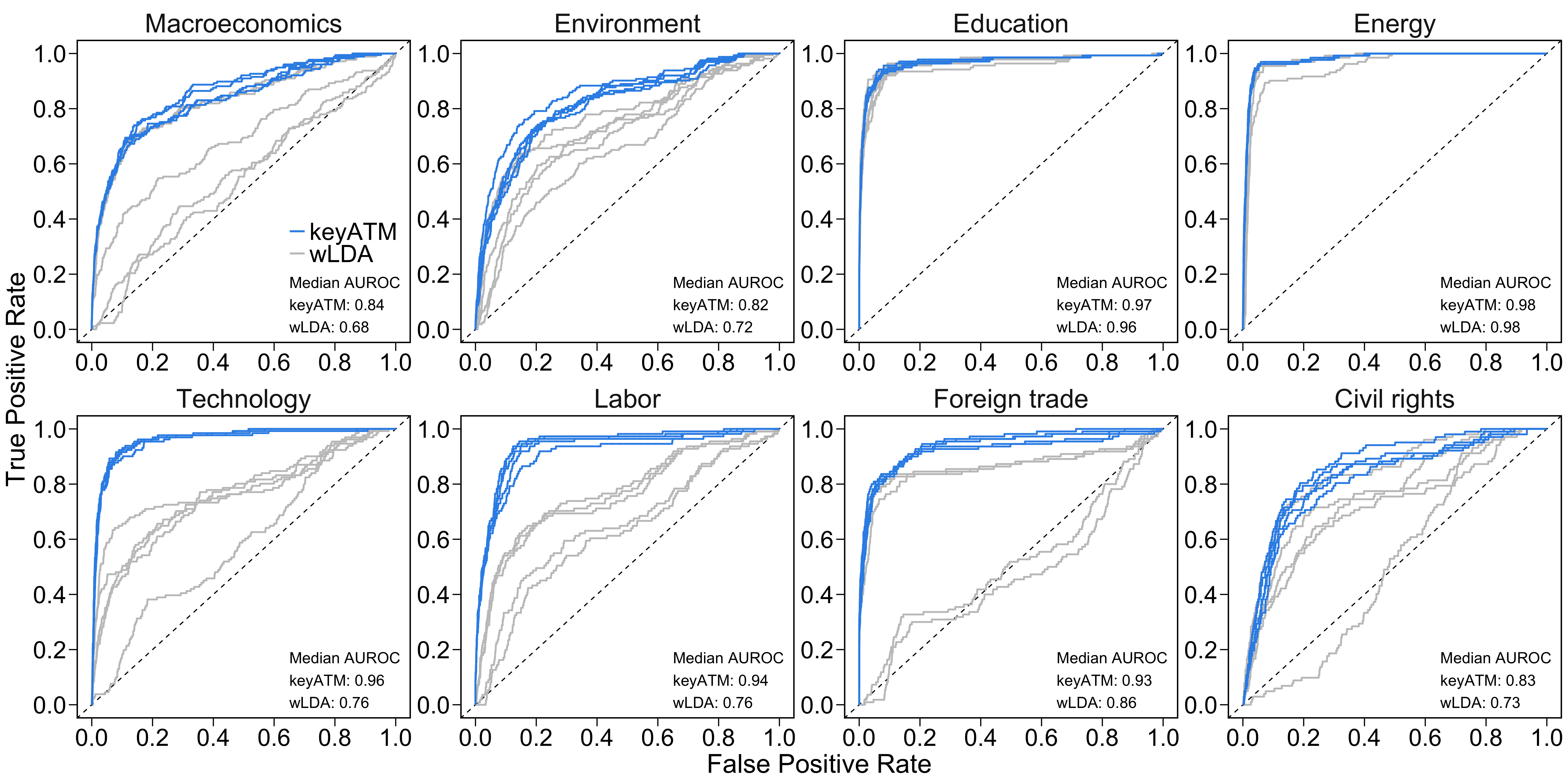}
\includegraphics[width= .65\linewidth]{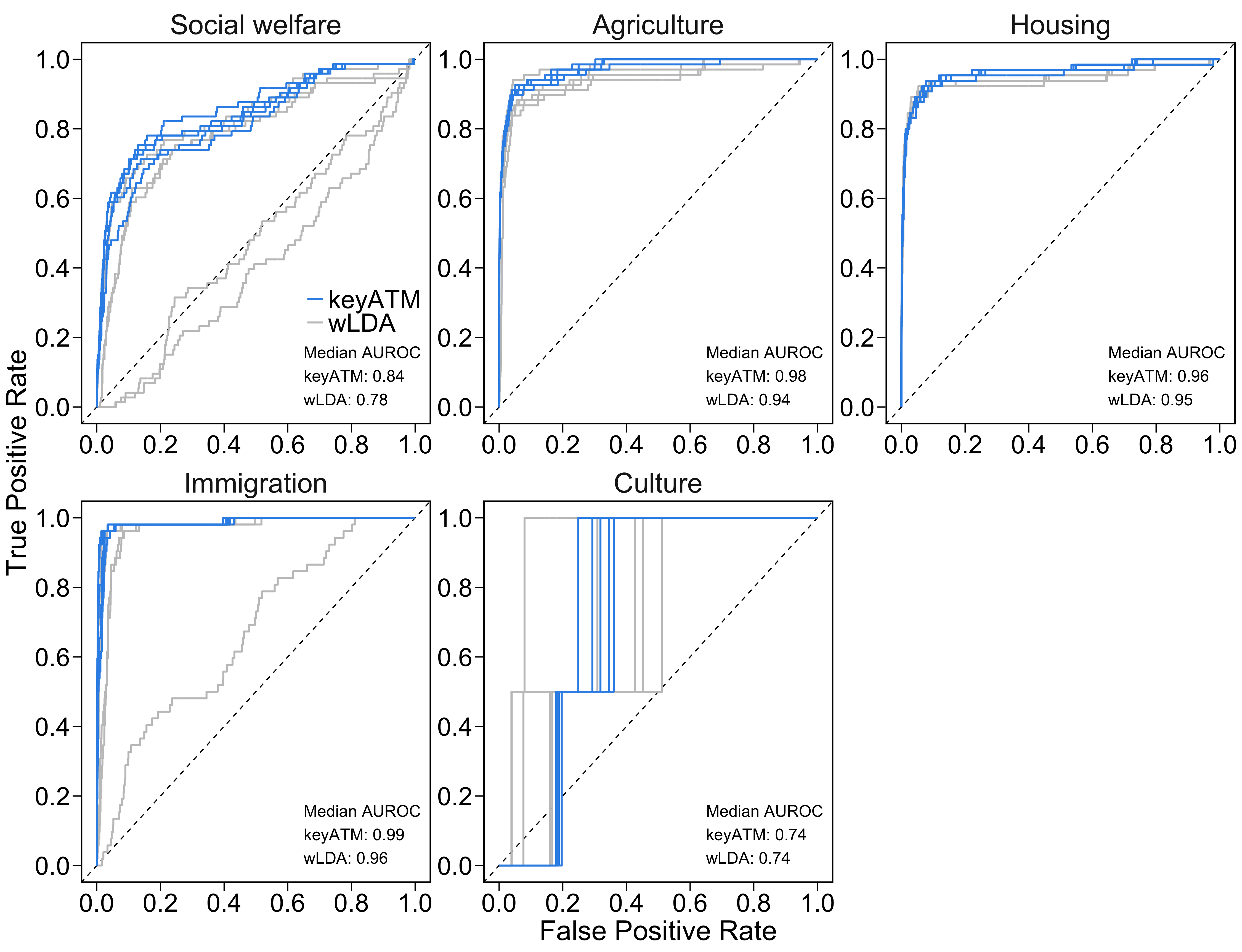}
\caption{\textbf{Comparison of the ROC curves between \keyATM{} and \wLDA.}}
\label{app:base-roc-fig}
\end{figure}

Figure~\ref{app:base-roc-fig} shows the ROC curves of all topics.
Each line represents the ROC curve from one of the five
  Markov chains with different starting values for \keyATM{} (blue
  lines) and \wLDA{} (grey lines).  The median AUROC indicates the
  median value of AUROC among five chains for each model.  \keyATM{}
  performs at least as well as \wLDA{} in all topics.

\subsection{Pooled Results}

\begin{figure}[!htb]
  \centering
  \includegraphics[width=0.9 \linewidth]{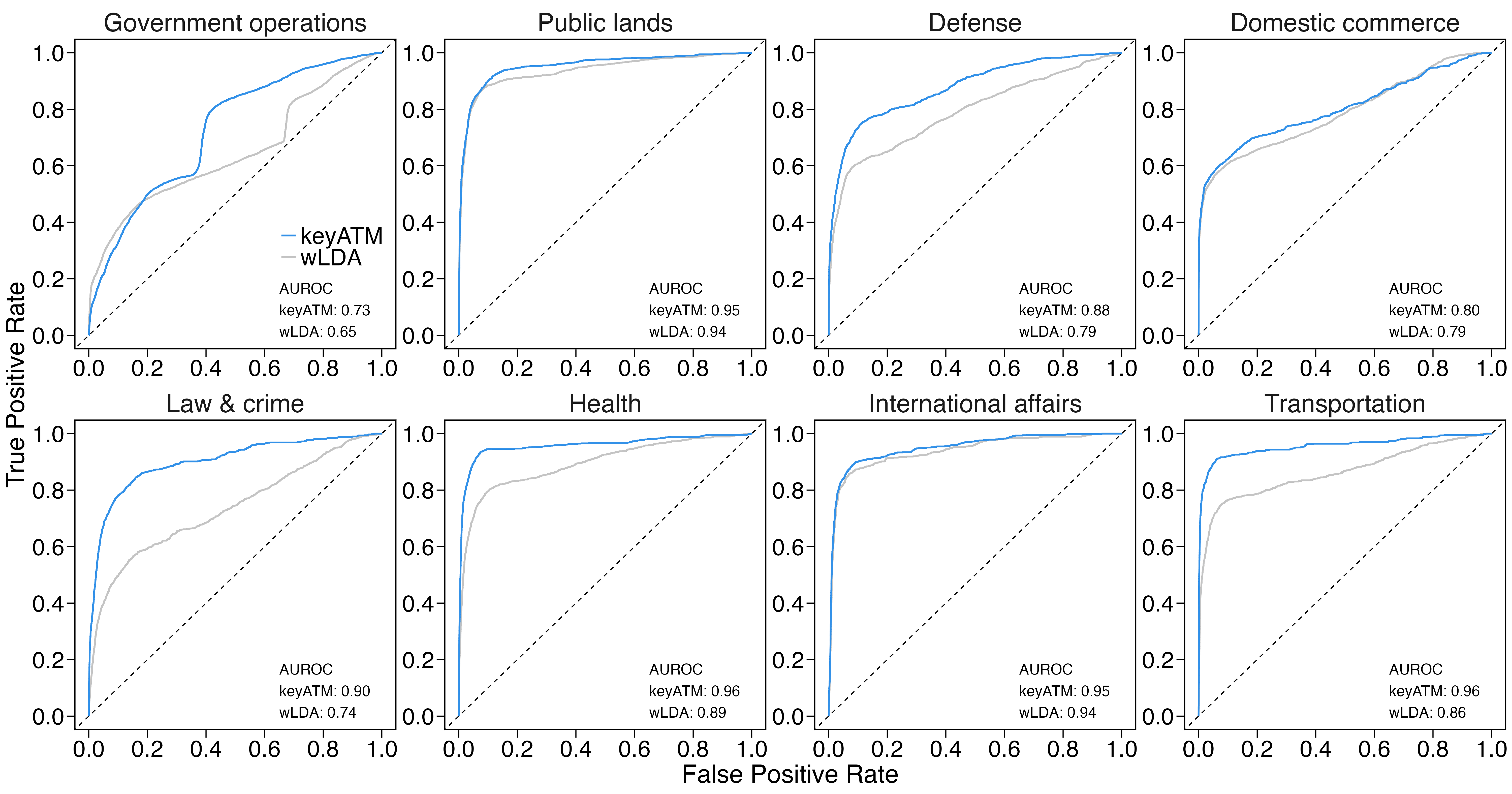}
  \centering
  \includegraphics[width=0.9 \linewidth]{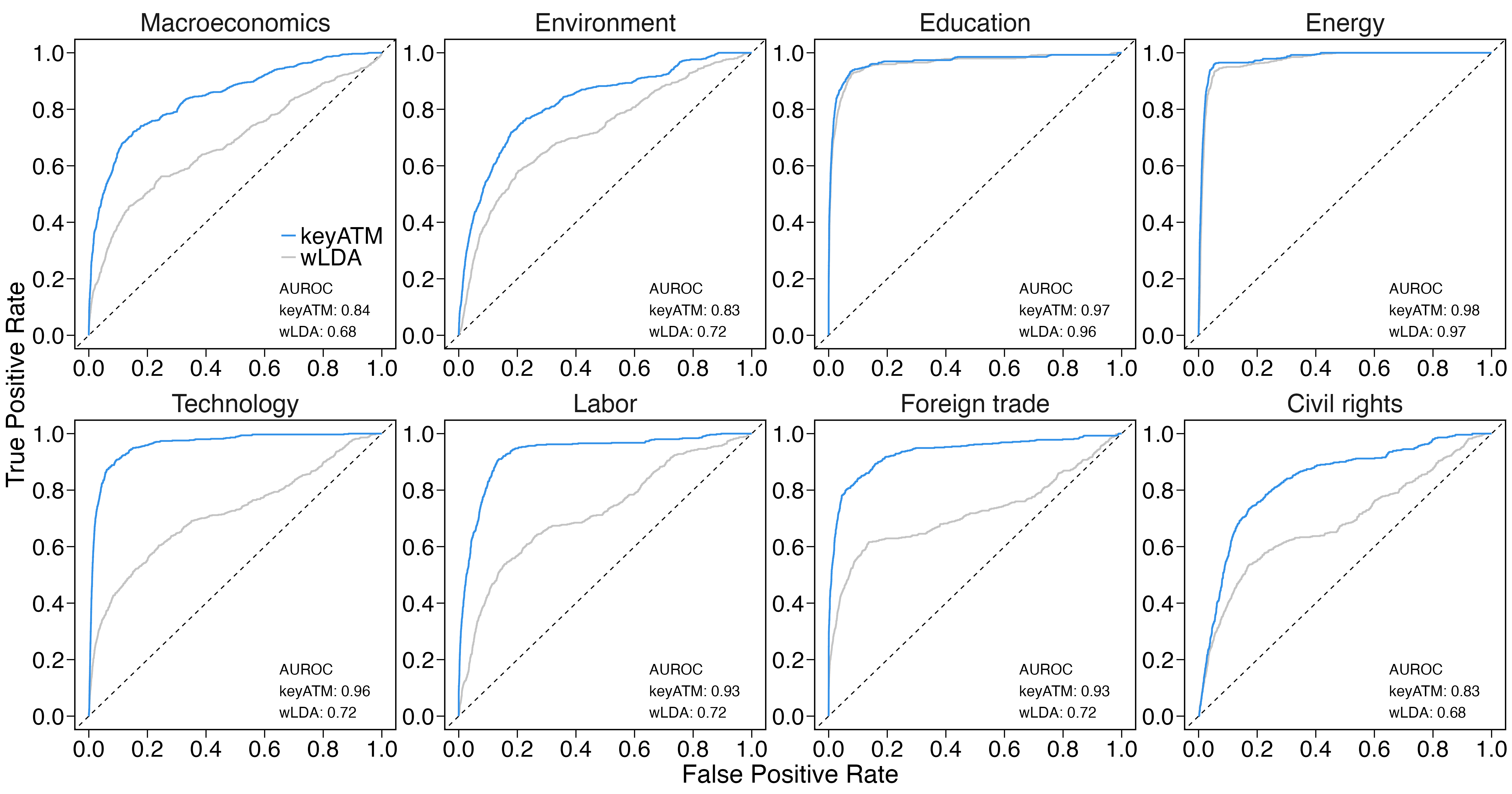}
  \centering
  \includegraphics[width=0.65 \linewidth]{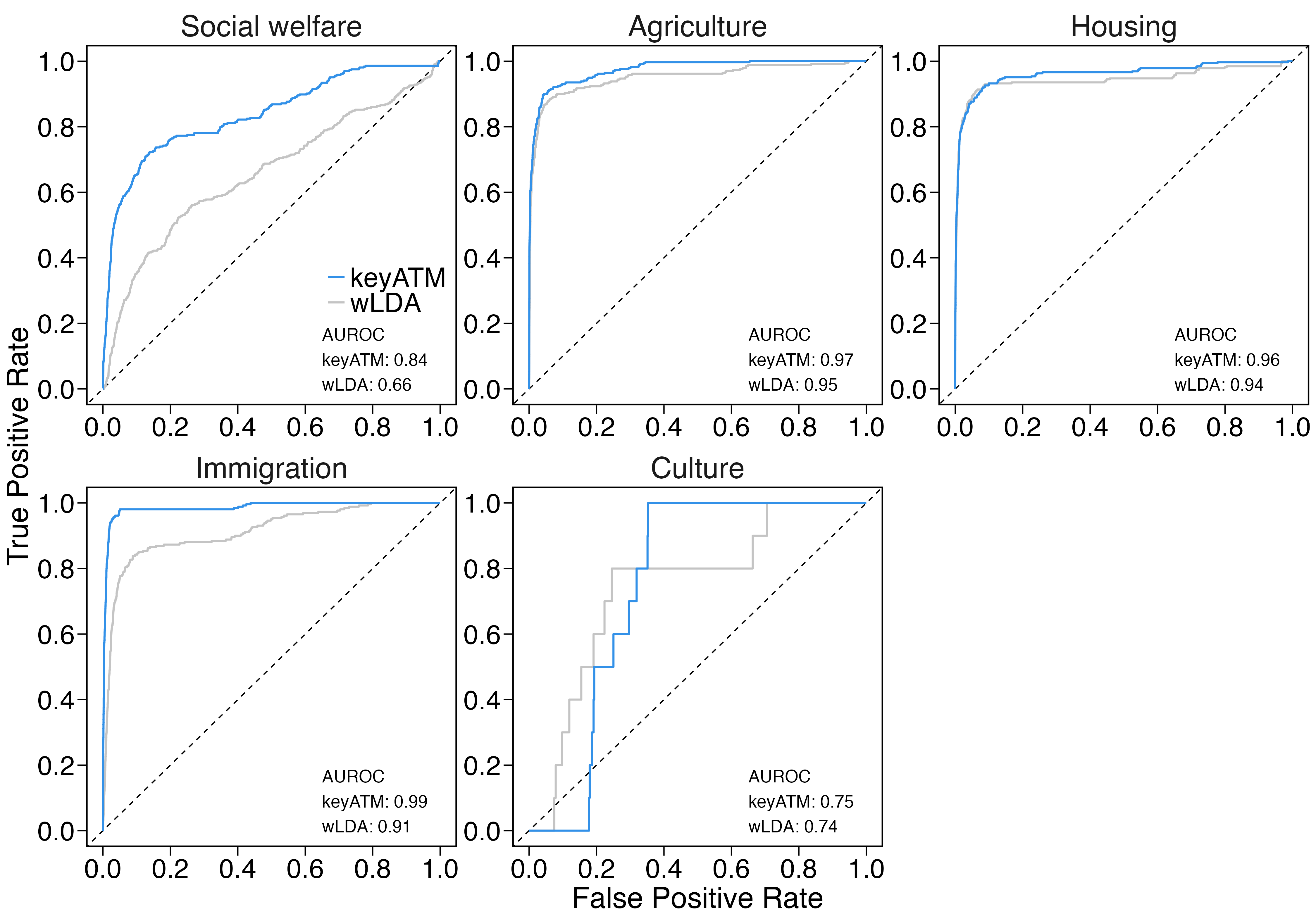}
  \caption{\textbf{Comparison of the ROC curve between \keyATM{} and
    \wLDA{} based on the Combined Chains.}}
  \label{app:fig-roc-pooled}
\end{figure}

Figure~\ref{app:fig-roc-pooled} presents the pooled results after
combining five Markov chains for \keyATM{} and \wLDA, respectively.
Each line represents the ROC curve based on pooled five
  Markov chains with different starting values for \keyATM{} (blue
  lines) and \wLDA{} (grey lines). The figures show that \keyATM{}
substantially outperforms \wLDA{} in terms of the AUROC.

\subsection{Quality of Keywords and Performance of the \keyATM}
\label{subsubsec:quality}

\begin{figure}[!t]
  \spacingset{1}
  \centering
  \vspace{-.25in}
\includegraphics[width= 0.7\linewidth]{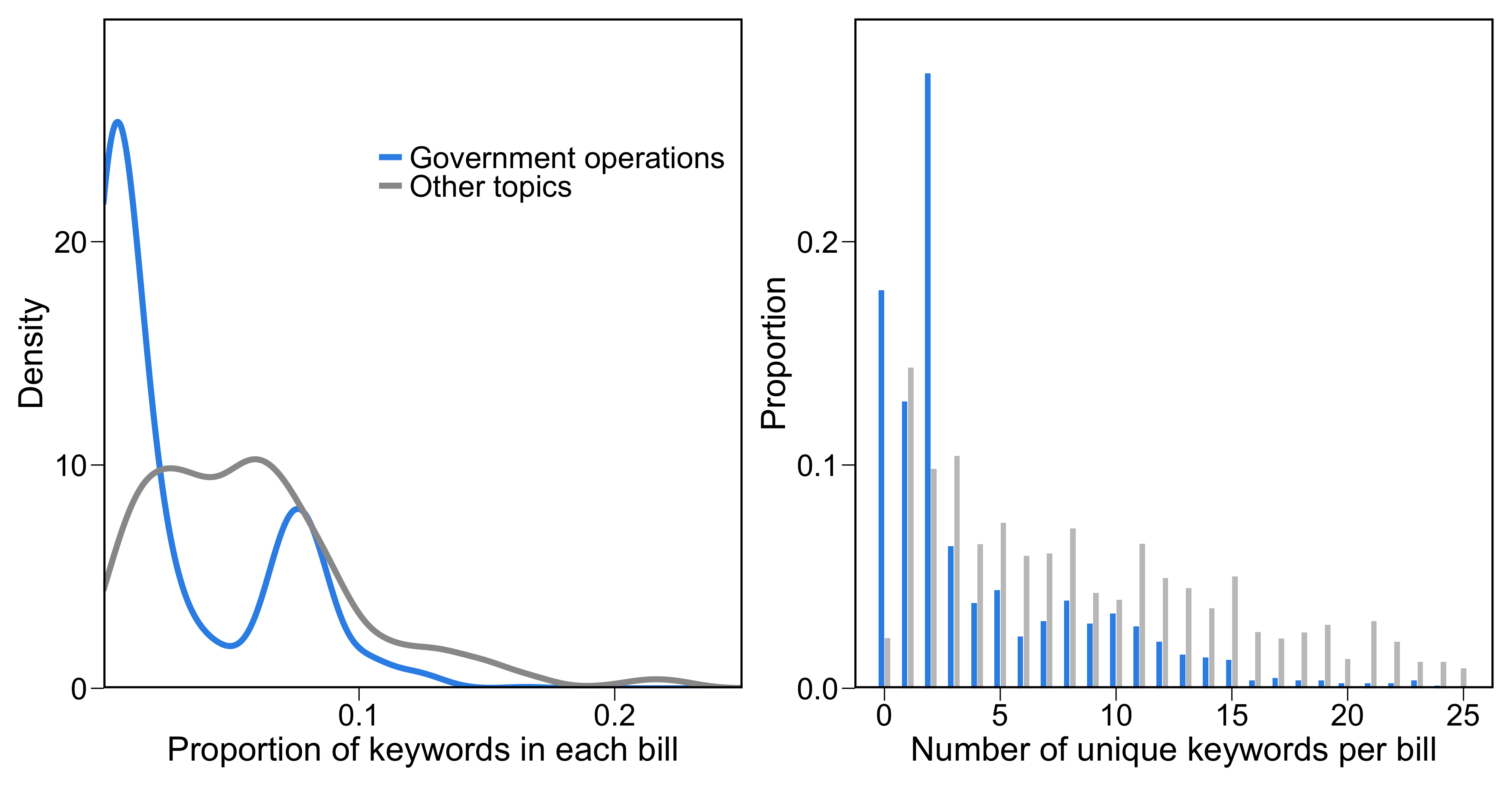}
\caption{\textbf{Poor quality of keywords for the \textit{Government
      operations} topic.}  The left panel presents the histogram for
  the proportion of keywords in each of the bills classified to the
  \textit{Government operations} topic (blue bars) by the
  Congressional Bills Project.  Compared to the average of other five topics
  (gray bars) from Table~\ref{tab:base-topwords}, the keywords appear
  less frequently in the \textit{Government operations} topic.  The
  right panel presents the histogram for the number of unique keywords
  that appear in each of the bills classified to the
  \textit{Government operations} topic.  Unique keywords appear less
  frequently in this topic than the other five topics.}
\label{fig:base-keys}
\end{figure}

The poor quality of keywords for the \textit{Government operations
  topic} appears to explain the failure of \keyATM{} in terms of both
topic interpretability and classification.  The left panel of
Figure~\ref{fig:base-keys} presents the histogram for the proportion
of keywords among all words in each of the bills classified to the
\textit{Government operations} topic by the CBP human coders.  When
compared to the average of the other five topics shown in
Table~\ref{tab:base-topwords} (gray bars), the keywords for this topic
(blue bars) appear much less frequently in the relevant bills.
Furthermore, the right panel shows the number of unique keywords
contained in each relevant bill.  Again, unique keywords appear in the
\textit{Government operations} bills less frequently than the
corresponding keywords do in the other five topics.  The results
suggest that selecting high-quality keywords is critical for
the successful application of \keyATM.

\clearpage

\subsection{The Results based on Different Keywords}\label{app:base-different-keywords}

This section provides the results based a different set of keywords.
As is the keyword set used for the main analysis, we derive the
keywords of each topic from the description provided by the CAP.
However, we do not prune any overlapping keywords and do not remove
words or phrases that have little to do with the substance of each
topic.  Therefore, the number of keywords assigned to each topic is
much larger than those used in the main analysis.

\subsubsection{The Full List of Different Keywords}\label{app:base-different-keywords-selection}

\begin{longtable}{p{2.4cm}p{13.1cm}} \hline
Label                 & Keywords\\  \hline
Government operations
& administration administrative advertising appointment appropriation attack auditing branch campaign capital census city civil claim coin collection commemorative compensation constitutional construction contractor corporation currency description efficiency elsewhere employee enforcement finance fraud government holiday individual intergovernmental issue local mail management medal mention mint multiple nomination observation operations oversight pension policy political post postal process procurement property provision reform registration regulation relate relation statistic substantive system tax terrorist victim voter without \\ \hdashline
Public lands
& affair assistance civil control cultural dependency description development fire flood forest grazing historic indigenous issue land livestock management natural parks people policy recreation relate research resource site staff territorial territorie water work \\      \hdashline
Defense
& activity affairs agreement aid air alliance appropriation arm armed assistance base capability civil civilian claim closing collateral compensation compliance construction contract contractor control conversion coordination country courts covert damage dependent description development direct disposal domestic employment environmental equipment espionage evaluation forces foreign fraud generally hazardous homeland industry injure installation intelligence issue land manpower material military modernization non nuclear old operation oversee oversight peacekeeping personnel policy population prisoner procurement proliferation readiness relate research reserve response sale sealift security services settlement stockpile strategic support system terrorism transfer veterans war waste weapon  \\      \hdashline
Domestic commerce
& abuse access accounting antitrust appropriation availability bank bankruptcy business card commerce commercial commodity consumer copyright corporate cost credit description development disaster domestic exchange facilitation finance financial fitness flood fraud gambling generally governance government health impact industry institution insurance intellectual investment issue management merger mortgage municipal natural non patent personal policy preparedness promote promotion property record reform regulate regulation relate relief research safety security small sport subsidize system tourism trading  \\      \hdashline
Law and Crime         &
abuse administration appropriation border child civil code combat control counterfeiting court crime criminal custom cyber description domestic drug effort enforcement exploitation family fine fraud guideline illegal impact international issue jail judiciary justice juvenile kidnapping launder legal mention money organize parental parole penalty police pornography pre prevention prison recidivism reduce relate release representation response security sexual special specialize system terrorism traffic violence welfare white  \\      \hdashline
Health                & abuse alcohol ambulance appropriation availability benefit broad care change child clinical comprehensive construction cost cover coverage description development device disease drug education effect facility fraud generally government health home ill illegal industry infant insurance insurer issue lab labor liability license long malpractice manpower medical mental multiple pay payment pharmaceutical policy practice prescription prevention promotion provider quality quantity reduce reform regulation rehabilitation relate research school specific supply system tobacco topic training treatment type unfair \\ \hdashline
International affairs &
abroad affairs agreement aid appropriation assessment associate bank citizen code combat committee conservation convention country court crime criminal cross debt description develop development diplomacy directly economic effort embassy europe european exploitation fight finance financial foreign genocide government human humanity implication increase individual institution international issue legal lending mechanism monetary nations ocean olympic organization passport piracy policy political red region regional related relation resource right sea see sovereign specific specifically target terrorism treaty union violation western world \\  \hdashline
Transportation  &
air airport appropriation availability aviation channel construction control deployment description development employment freight generally government highway infrastructure initiative inland issue maintenance maritime mass new pilot policy rail railroad regulate regulation relate research safety security ship technology traffic training transportation travel waterway work  \\
\hdashline
Macroeconomics        & bank budget budgeting central code control cost debt deficit description domestic effort emergency enforcement growth impact industrial inflation interest issue live macroeconomic manufacturing monetary policy price rate reduce relate revitalization tax taxis treasury unemployment wage \\       \hdashline
Environment       & air airline alternative animal appropriation asbestos change chemical climate conservation contamination control description development disposal domestic drinking endanger energy environment environmental forest government hazardous illicit indoor issue laboratory land noise performance policy pollution product protection recycling regulate regulation relate research resource reuse runoff safety solid specie substance supply technology toxic trade treatment waste wastewater water wildlife             \\      \hdashline
Education             & adult appropriation area bilingual child college description development education educational effort elementary excellence finance foreign generally government handicapped high impact improve increase initiative issue language literacy loan math mentally need outcome physically policy primary quality reform regulate regulation relate research rural safety school schools science secondary skill special specific standard student university vocational \\      \hdashline
Energy     & alternative appropriation biofuel clean coal commercial conservation description development disposal drilling efficiency electrical electricity energy gas gasification gasoline generally geothermal government home hydrogen hydropower issue natural nuclear oil policy power price production regulate regulation relate renewable research safety security shortage spill technology trade utility vehicle waste \\      \hdashline
Technology & agreement broadcast commercial communication computer cooperation description development effort encourage exploitation exploration forecast form generally geological government high industry infrastructure international internet issue mention military newspaper promotion publishing radio regulation relate research resource satellite science security space speed survey technology telecommunication telecommunications telephone television transfer weather \\      \hdashline
Labor                 & account adult appropriation bargaining benefit child collective compensation description development disease displace effort employee employer employment fair generally government guest injury insurance issue job labor migrant minimum overtime pension policy protection regulate relate relation retirement retrain safety seasonal standard training unemployment union wage work worker workforce youth  \\       \hdashline
Foreign trade&
agreement appropriation balance barrier business competitiveness control corporate description development dispute domestic exchange export foreign generally government impact import industry international investment issue negotiation payment policy private productivity promotion rate regulate regulation relate subsidy tariff tax trade treaty \\      \hdashline
Civil rights          & abortion access activity age anti civil communist contract description discrimination disease ethnic expand expression franchise freedom gender generally government group handicap information issue local mandatory minority orientation participation party policy privacy racial record relate religious retirement right rights sex sexual speech type voting \\      \hdashline
Social welfare        & aid alleviate assess assistance association care charity child credit dependency description direct disability disable domestic elderly family food generally government income issue leave low mental organization parental pension people physical policy poverty relate social tax volunteer welfare youth  \\       \hdashline
Agriculture           & agricultural agriculture animal appropriation commercial conservation consumer control crop description development disaster disease effort farmer fish fishery food foreign government impact information inspection insurance issue labeling market pest pesticide policy product promotion rancher regulation relate requirement research safety seafood subsidy trade welfare  \\      \hdashline
Housing  & affairs affordability assistance community description development economic effort elderly facility family generally handicap homeless homelessness housing income individual issue low military neighborhood non policy reduce relate research rural subsidy urban veteran \\      \hdashline
Immigration           & citizenship description immigration issue refugee related \\      \hdashline
Culture               & culture cultural \\ \hline
\caption{\normalsize \textbf{Different Keywords for the base \keyATM{}}: A different set of keywords used in the base \keyATM{} application} \label{table:app-base-keywords-diff}
\end{longtable}

\subsubsection{Top words}

Table~\ref{tab:app-base-diffkeywords-topwords} shows the
  top ten words with the highest estimated probabilities for each
  topic and for \keyATM{} and \wLDA. The results demonstrate that the
  advantages of \keyATM{} presented in the main text
  (Table~\ref{tab:base-topwords}) still hold with different sets of
  keywords: First, the \textit{Labor} topic of \keyATM{} lists many
  related words whereas the same topic for \wLDA{} does not include
  any related terms.  Second, \wLDA{} cannot find meaningful terms for
  the \textit{Foreign trade} topic and instead creates two topics
  whose top ten words are related to the \textit{Transportation} topic
  (under the label of \textit{Foreign trade} and
  \textit{Transportation}). The top words of the \textit{Foreign
    trade} topic for \keyATM{} contain words related to trade.  Third,
  \keyATM{} can distinguish \textit{Law \& crime} and
  \textit{Immigration} but \wLDA{} cannot.

\begin{longtable}{p{2.4cm}p{2.4cm}|p{2.4cm}p{2.4cm}|p{2.4cm}p{2.4cm}}
  \hline
  \hline
  \multicolumn{2}{c|}{Government operations} & \multicolumn{2}{c|}{Public lands} & \multicolumn{2}{c}{Defense}\\ \hline \keyATM{} & \wLDA{} & \keyATM{} & \wLDA{} & \keyATM{} & \wLDA{}\\ \hline
office & congress & \textbf{land} & land & \textbf{military} & military \\
  \textbf{employee} & house & \textbf{water} & management & force & force \\
  commission & senate & \textbf{management} & area & member & system \\
  committee$^*$ & office & area$^*$ & forest & \textbf{air} & support \\
  \textbf{administration} & committee & \textbf{resource} & indian & authority & authorization \\
  district & commission & river & interior & code$^*$ & security \\
  \textbf{administrative} & representative & \textbf{forest} & resource & authorization & operation \\
  pay$^*$ & congressional & cost$^*$ & park & \textbf{reserve} & army \\
  board & strike & authorize & conservation & \textbf{armed} & committee \\
  \textbf{management} & bill & stat & within & army & air \\
        \hline  \hline
  \multicolumn{2}{c|}{Domestic commerce} & \multicolumn{2}{c|}{Law \& crime} & \multicolumn{2}{c}{Health}\\ \hline \keyATM{} & \wLDA{} & \keyATM{} & \wLDA{} & \keyATM{} & \wLDA{}\\ \hline
\textbf{financial} & financial & \textbf{court} & security & \textbf{health} & health \\
  \textbf{insurance} & institution & attorney & information & \textbf{care} & care \\
  \textbf{institution} & bank & person & intelligence & individual$^*$ & individual \\
  \textbf{bank} & company & \textbf{code} & homeland & \textbf{medical} & drug \\
  company & insurance & \textbf{crime} & committee & part & payment \\
  board & corporation & grant & director & \textbf{payment} & medical \\
  commission & board & \textbf{enforcement} & system & \textbf{coverage} & describe \\
  \textbf{security} & security & offense & foreign & social$^*$ & respect \\
  corporation$^*$ & credit & \textbf{child} & government & describe & social \\
  \textbf{credit} & commission & victim$^*$ & office & respect & part \\
  \hline
  \hline   \multicolumn{2}{c|}{International affairs} & \multicolumn{2}{c|}{Transportation} & \multicolumn{2}{c}{Macroeconomics}\\ \hline \keyATM{} & \wLDA{} & \keyATM{} & \wLDA{} & \keyATM{} & \wLDA{}\\ \hline
assistance$^*$ & assistance & \textbf{transportation} & transportation & expense & transfer \\
  \textbf{foreign} & foreign & \textbf{highway} & highway & appropriation$^*$ & appropriation \\
  \textbf{international} & country & carrier & safety & remain & emergency \\
  \textbf{country} & international & \textbf{safety} & vehicle & necessary & operation \\
  \textbf{government} & president & system$^*$ & carrier & authorize & military \\
  president & government & code$^*$ & motor & transfer$^*$ & construction \\
  \textbf{committee} & committee & \textbf{air} & system & expend & procurement \\
  \textbf{development} & development & vehicle$^*$ & strike & exceed & remain \\
  \textbf{organization} & export & \textbf{airport} & rail & activity$^*$ & maintenance \\
  export$^*$ & organization & strike & code & pursuant & budget \\
    \hline \hline\multicolumn{2}{c|}{Environment} & \multicolumn{2}{c|}{Education} & \multicolumn{2}{c}{Energy}\\ \hline \keyATM{} & \wLDA{} & \keyATM{} & \wLDA{} & \keyATM{} & \wLDA{}\\ \hline
administrator & water & \textbf{education} & school & \textbf{energy} & energy \\
  \textbf{product} & river & \textbf{school} & education & fuel & fuel \\
  \textbf{regulation} & cost & \textbf{educational} & educational & \textbf{gas} & gas \\
  requirement$^*$ & stat & \textbf{student} & student & facility$^*$ & change \\
  review & authorize & \textbf{child} & child & \textbf{vehicle} & facility \\
  drug$^*$ & carry & grant & local & \textbf{oil} & new \\
  require & non & local$^*$ & grant & electric & vehicle \\
  fee & development & part & part & \textbf{power} & electric \\
  submit & resource & institution$^*$ & activity & \textbf{technology} & production \\
  application & study & activity$^*$ & eligible & \textbf{natural} & administrator \\
      \hline
  \hline   \multicolumn{2}{c|}{Technology} & \multicolumn{2}{c|}{Labor} & \multicolumn{2}{c}{Foreign trade}\\ \hline \keyATM{} & \wLDA{} & \keyATM{} & \wLDA{} & \keyATM{} & \wLDA{}\\ \hline
information$^*$ & research & \textbf{benefit} & apply & \textbf{trade} & air \\
  system$^*$ & technology & payment$^*$ & tax & change$^*$ & vessel \\
  \textbf{technology} & development & requirement$^*$ & amendment & \textbf{agreement} & airport \\
  \textbf{research} & establish & \textbf{employee} & end & good & transportation \\
  committee$^*$ & committee & period & taxable & head & aviation \\
  activity$^*$ & activity & rate$^*$ & respect & \textbf{import} & administrator \\
  director & administrator & pay$^*$ & period & article & aircraft \\
  \textbf{development} & system & determine & individual & chapter & carrier \\
  business$^*$ & information & code$^*$ & case & free & administration \\
  develop$^*$ & carry & provision$^*$ & relate & new$^*$ & coast \\
    \hline \hline
    \multicolumn{2}{c|}{Civil rights} & \multicolumn{2}{c|}{Social welfare} & \multicolumn{2}{c}{Agriculture}\\ \hline \keyATM{} & \wLDA{} & \keyATM{} & \wLDA{} & \keyATM{} & \wLDA{}\\ \hline
\textbf{information} & person & \textbf{tax} & expense & \textbf{food} & food \\
  action & action & apply & appropriation & \textbf{agricultural} & agricultural \\
  person & regulation & taxable & authorize & loan$^*$ & loan \\
  require & require & amendment & remain & \textbf{agriculture} & agriculture \\
  order & information & \textbf{relate} & necessary & farm & payment \\
  procedure & court & \textbf{income} & expend & rural$^*$ & farm \\
  provision$^*$ & order & end & office & \textbf{crop} & producer \\
  individual$^*$ & rule & strike & exceed & payment$^*$ & crop \\
  authority & commission & individual$^*$ & transfer & producer & rural \\
  request & review & respect & activity & commodity$^*$ & commodity \\  \hline
  \hline
  \multicolumn{2}{c|}{Housing} & \multicolumn{2}{c|}{Immigration} & \multicolumn{2}{c}{Culture}\\ \hline \keyATM{} & \wLDA{} & \keyATM{} & \wLDA{} & \keyATM{} & \wLDA{}\\ \hline
\textbf{housing} & housing & security$^*$ & alien & congress & member \\
  \textbf{assistance} & assistance & alien & attorney & house & strike \\
  loan$^*$ & loan & intelligence$^*$ & child & senate & code \\
  \textbf{development} & development & \textbf{immigration} & crime & committee$^*$ & force \\
  \textbf{family} & family & homeland$^*$ & immigration & strike & pay \\
  \textbf{community} & mortgage & foreign$^*$ & grant & office & military \\
  grant & community & information$^*$ & enforcement & representative & officer \\
  mortgage$^*$ & insurance & border$^*$ & person & congressional & authorize \\
  \textbf{income} & income & describe & court & bill & duty \\
  unit & unit & attorney & offense & veteran$^*$ & reserve \\
      \hline \hline
    \caption{\normalsize \textbf{Top words of the base \keyATM{} with a
  different sets of keywords.} The table shows the top ten words with the highest estimated probability for each topic under each model. For \keyATM, the pre-specified keywords for each topic appear in bold letters whereas the asterisks indicate the keywords specified for another topic.}
\label{tab:app-base-diffkeywords-topwords}
\end{longtable}


\subsubsection{ROC Curves}\label{app:base-diffkeywords-roc}
\begin{figure}[!htb]
\centering
\includegraphics[width = 0.9\linewidth]{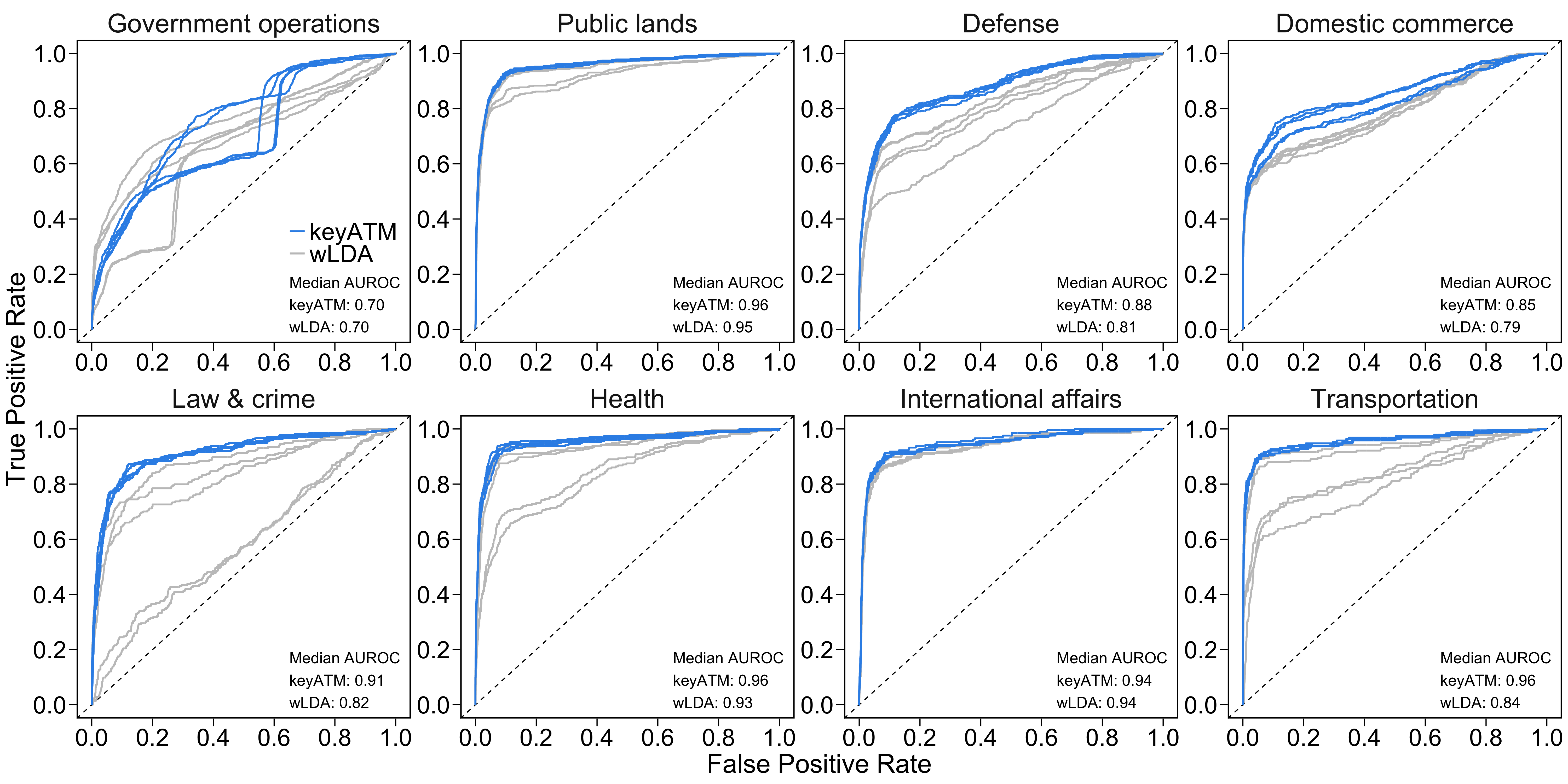}
%
\centering
\includegraphics[width = 0.9\linewidth]{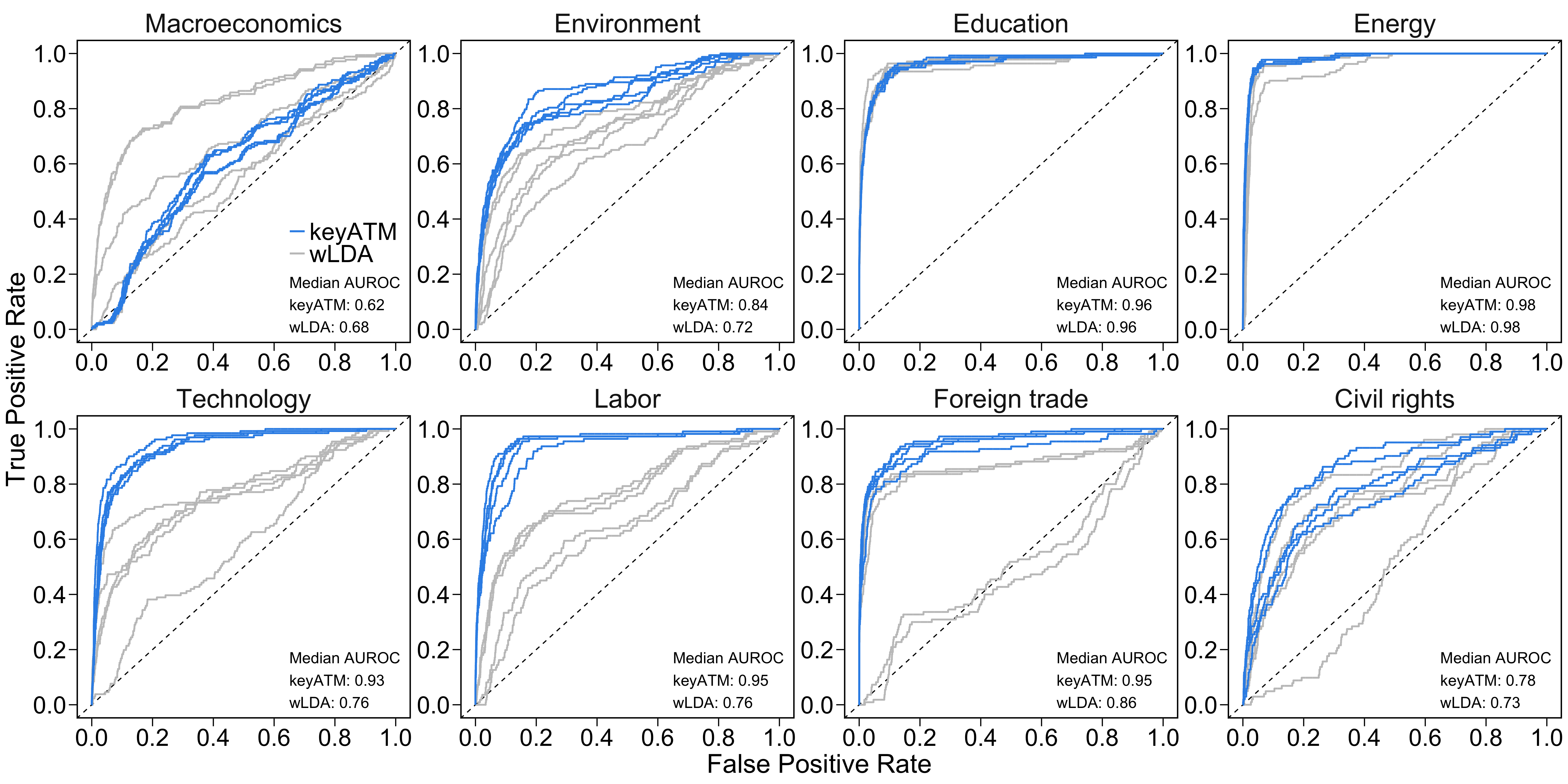}
\includegraphics[width = 0.65\linewidth]{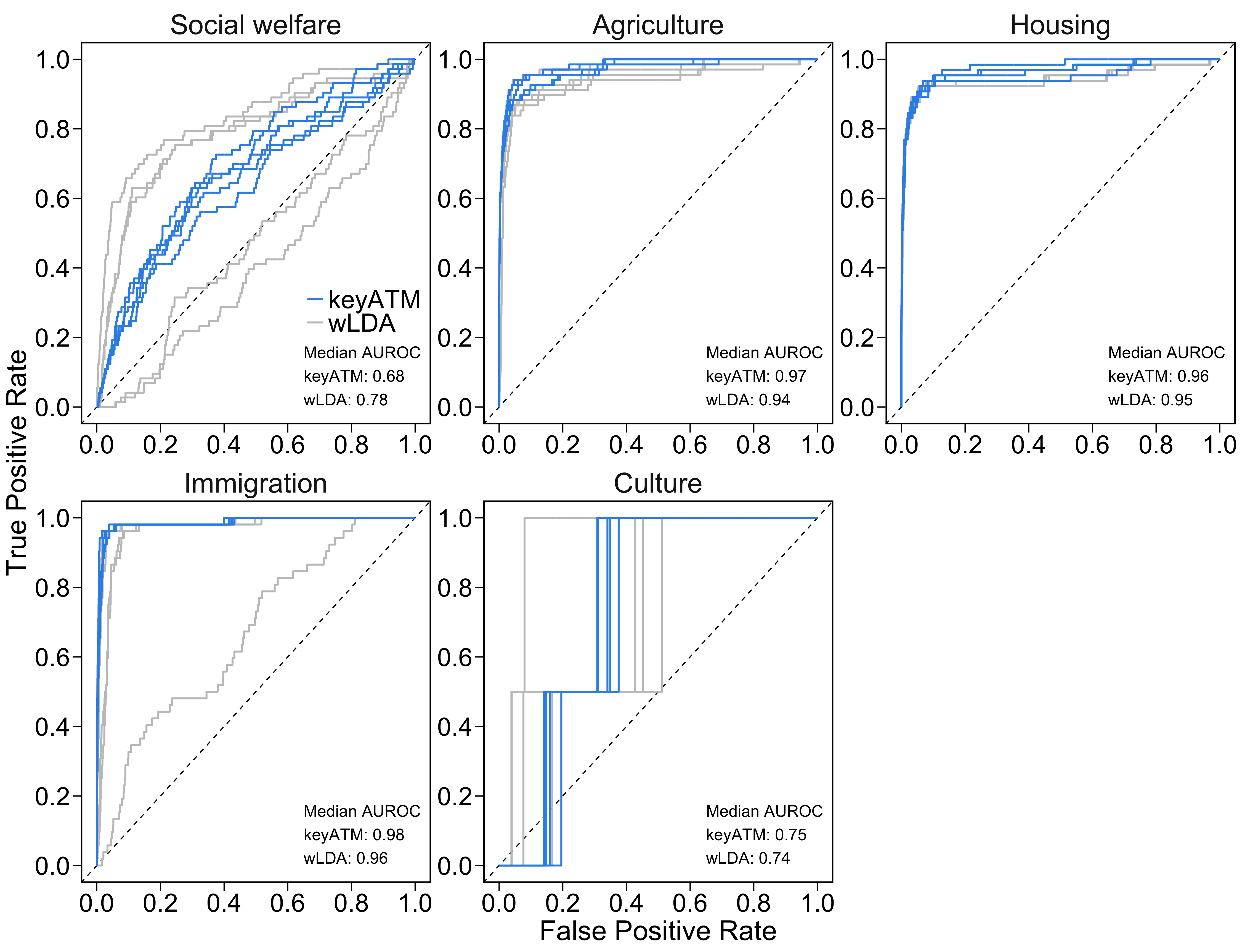}
\caption{\textbf{Comparison of the ROC curves between \keyATM{} and \wLDA{}
with a different set of keywords}}
\label{app:base-diff-keywords-roc-fig}
\end{figure}

Figure~\ref{app:base-diff-keywords-roc-fig} shows the ROC curves.
Each line represents the ROC curve from one of the five
  Markov chains with different starting values for \keyATM{} (blue
  lines) and \wLDA{} (grey lines).  The median AUROC indicates the
  median value of AUROC among five chains for each model.  \keyATM{}
  performs at least as well as \wLDA{} in 19 out of 21 topics.

\subsection{Randomly Removed Keywords}\label{app:base-random-keywords}

In this section, we present the base \keyATM{} results where we
randomly remove some keywords from the keyword sets shown in the
Section~\ref{app:base-keywords-selection}.  In particular, we randomly
remove 75\% or 50\% of keywords from the original keyword sets shown
in Table~\ref{table:app-base-keywords}.

The results are shown in Table~\ref{app:tab-random-keyword-auc}.
In most cases, the worst AUROC values from the base \keyATM{} are
still better than those of \wLDA{}.
Note that since the \textit{Immigration} topic only contains three keywords under
the original setting: ``citizenship'', ``immigration'', and
``refugee''.
We randomly select two out of these three keywords and fit the base \keyATM{}
for this topic.
Table~\ref{app:tab-random-keyword-auc}
  highlights the importance of keyword selection when the number of keywords
is extremely small.
For example, the \textit{Immigration} topic shows high variability
in the AUROC value.
The \textit{Immigration} topic performs poorly in models with the keyword sets 1, 3, and 6.
All of these sets do not contain ``immigration'' as a keyword.

\begin{table}[!ht]
\centering
\begin{tabular}{lclllllllll}
  \hline
  \hline & & \multicolumn{8}{c}{\keyATM} & \wLDA\\
    \cmidrule(lr){3-10} \cmidrule(lr){11-11}
  & \# of original  & \multicolumn{4}{c}{75\% of original} & \multicolumn{4}{c}{50\% of original} & \\
  \cmidrule(lr){3-6} \cmidrule(lr){7-10} \cmidrule(lr){11-11}
  Topic & keywords &  1 & 2 & 3 & 4 & 5 & 6 & 7 & 8 & N/A\\ \hline
  Government operations & 24 & 0.70 & 0.71 & 0.74 & 0.74 & 0.72 & 0.74 & 0.75 & 0.71 & 0.72 \\
  Public lands & 17 & 0.94 & 0.95 & 0.95 & 0.95 & 0.92 & 0.94 & 0.96 & 0.96 & 0.91 \\
  Defense & 23 & 0.88 & 0.88 & 0.87 & 0.87 & 0.86 & 0.87 & 0.87 & 0.88 & 0.73 \\
  Domestic commerce & 25 & 0.81 & 0.82 & 0.79 & 0.79 & 0.80 & 0.78 & 0.81 & 0.79 & 0.78 \\
  Law \& crime & 25 & 0.88 & 0.90 & 0.86 & 0.87 & 0.90 & 0.86 & 0.89 & 0.87 & 0.57 \\
  Health & 25 & 0.96 & 0.96 & 0.96 & 0.96 & 0.96 & 0.95 & 0.90 & 0.96 & 0.95 \\
  International affairs & 24 & 0.95 & 0.94 & 0.95 & 0.68 & 0.73 & 0.75 & 0.94 & 0.95 & 0.93 \\
  Transportation & 21 & 0.94 & 0.96 & 0.95 & 0.96 & 0.96 & 0.95 & 0.94 & 0.95 & 0.80 \\
  Macroeconomics & 17 & 0.83 & 0.64 & 0.81 & 0.81 & 0.57 & 0.78 & 0.62 & 0.84 & 0.54 \\
  Environment & 24 &  0.89 & 0.80 & 0.80 & 0.93 & 0.75 & 0.84 & 0.88 & 0.83 & 0.72 \\
  Education & 25 & 0.97 & 0.97 & 0.97 & 0.97 & 0.96 & 0.96 & 0.97 & 0.97 & 0.97 \\
  Energy & 25 & 0.98 & 0.98 & 0.97 & 0.97 & 0.98 & 0.97 & 0.98 & 0.98 & 0.95 \\
  Technology & 25 & 0.92 & 0.96 & 0.96 & 0.54 & 0.96 & 0.95 & 0.93 & 0.89 & 0.75 \\
  Labor & 25 & 0.93 & 0.85 & 0.94 & 0.93 & 0.90 & 0.94 & 0.79 & 0.83 & 0.76 \\
  Foreign trade & 17 & 0.94 & 0.92 & 0.94 & 0.92 & 0.89 & 0.72 & 0.91 & 0.94 & 0.47 \\
  Civil rights & 25 & 0.84 & 0.81 & 0.84 & 0.80 & 0.74 & 0.74 & 0.69 & 0.86 & 0.73 \\
  Social welfare & 21 & 0.81 & 0.83 & 0.81 & 0.85 & 0.71 & 0.84 & 0.81 & 0.88 & 0.47 \\
  Agriculture & 17 & 0.96 & 0.97 & 0.95 & 0.98 & 0.98 & 0.97 & 0.97 & 0.96 & 0.98 \\
  Housing & 13 & 0.95 & 0.94 & 0.93 & 0.94 & 0.96 & 0.95 & 0.96 & 0.95 & 0.95 \\
  Immigration & 3 & 0.51 & 0.98 & 0.49 & 0.99 & 0.99 & 0.50 & 0.99 & 0.99 & 0.96 \\
  Culture & 2 & 0.62 & 0.77 & 0.88 & 0.26 & 0.98 & 0.52 & 0.78 & 0.72 & 0.72 \\
    \hline
\end{tabular}
\caption{\normalsize \textbf{Comparison of AUROC with different keyword sets.} The table
presents AUROC values from the base \keyATM{} with 8 different-randomly-selected
keyword sets and \wLDA. We run five chains for each setting and
the results are based on the MCMC draws from one
of the five chains that has the median performance in terms of the overall AUROC.
In most cases, the worst AUROC values from the base \keyATM{} are
still better than those of \wLDA{}.}
\label{app:tab-random-keyword-auc}
\end{table}

\begin{sidewaysfigure}
  \includegraphics[width = \linewidth]{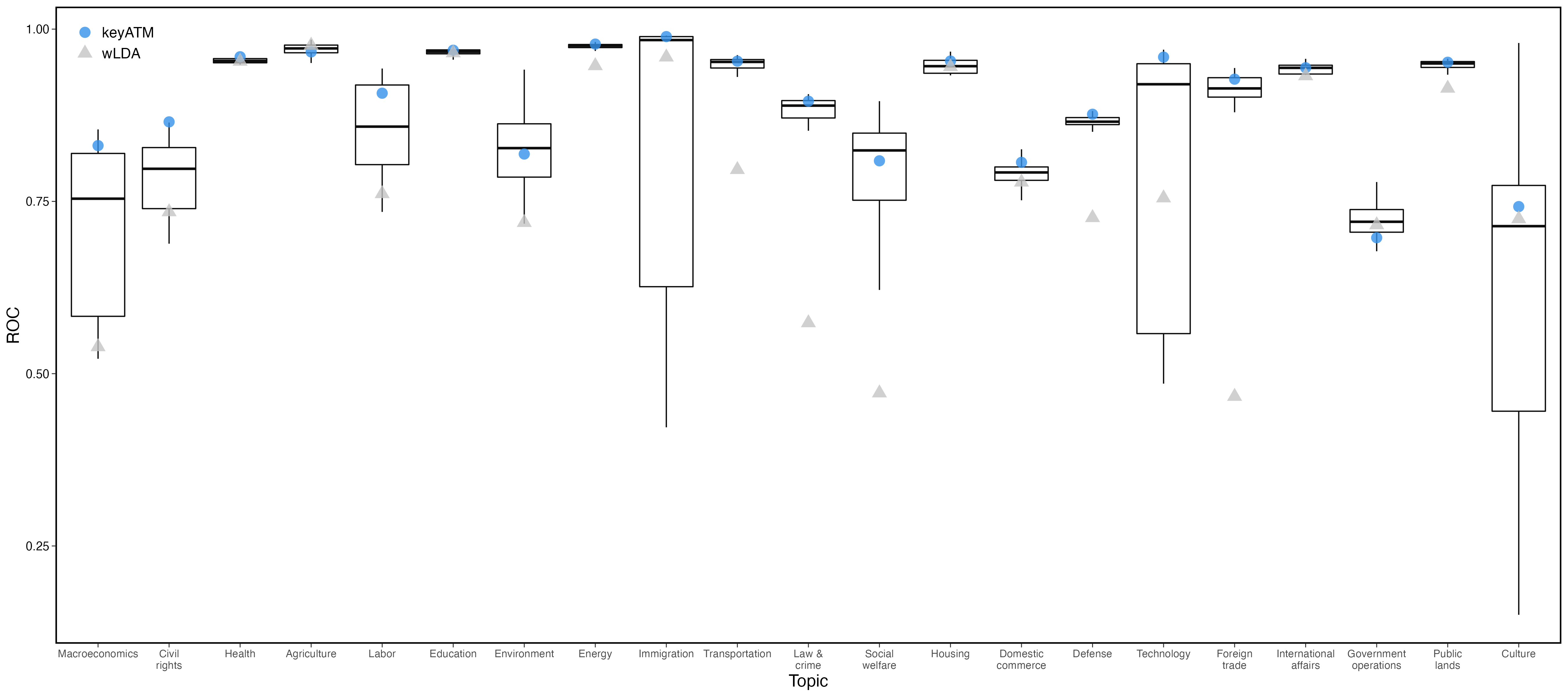}
  \caption{
    \textbf{AUROC with randomly selected keyword sets.} The boxplots
  present the AUROC with randomly selected 50 different keyword sets.
  Blue circle indicates the AUROC for \keyATM{} with all keywords and
  grey triangles represent the AUROC for \wLDA{} (those presented in the
  Figure~\ref{app:base-roc-fig}).
  }
  \label{app:fig-random-keyword-auc}
\end{sidewaysfigure}

Furthermore, we create 50 different keyword sets where
  we randomly remove 50\% of keywords from the keyword sets shown in
  Table~\ref{table:app-base-keywords}.  We compute AUROC for
  each topic and present the results in
  Figure~\ref{app:fig-random-keyword-auc}.  Again, in most cases, the
  worst AUROC values from the base \keyATM{} are still better than
  those of \wLDA{}, except for some topics such as
  \textit{Immigration}, which has only three keywords.  There is a
  large variation in AUROC for the \textit{Immigration} topic.

\begin{figure}[!ht]
  \centering
  \includegraphics[width = 0.8\linewidth]{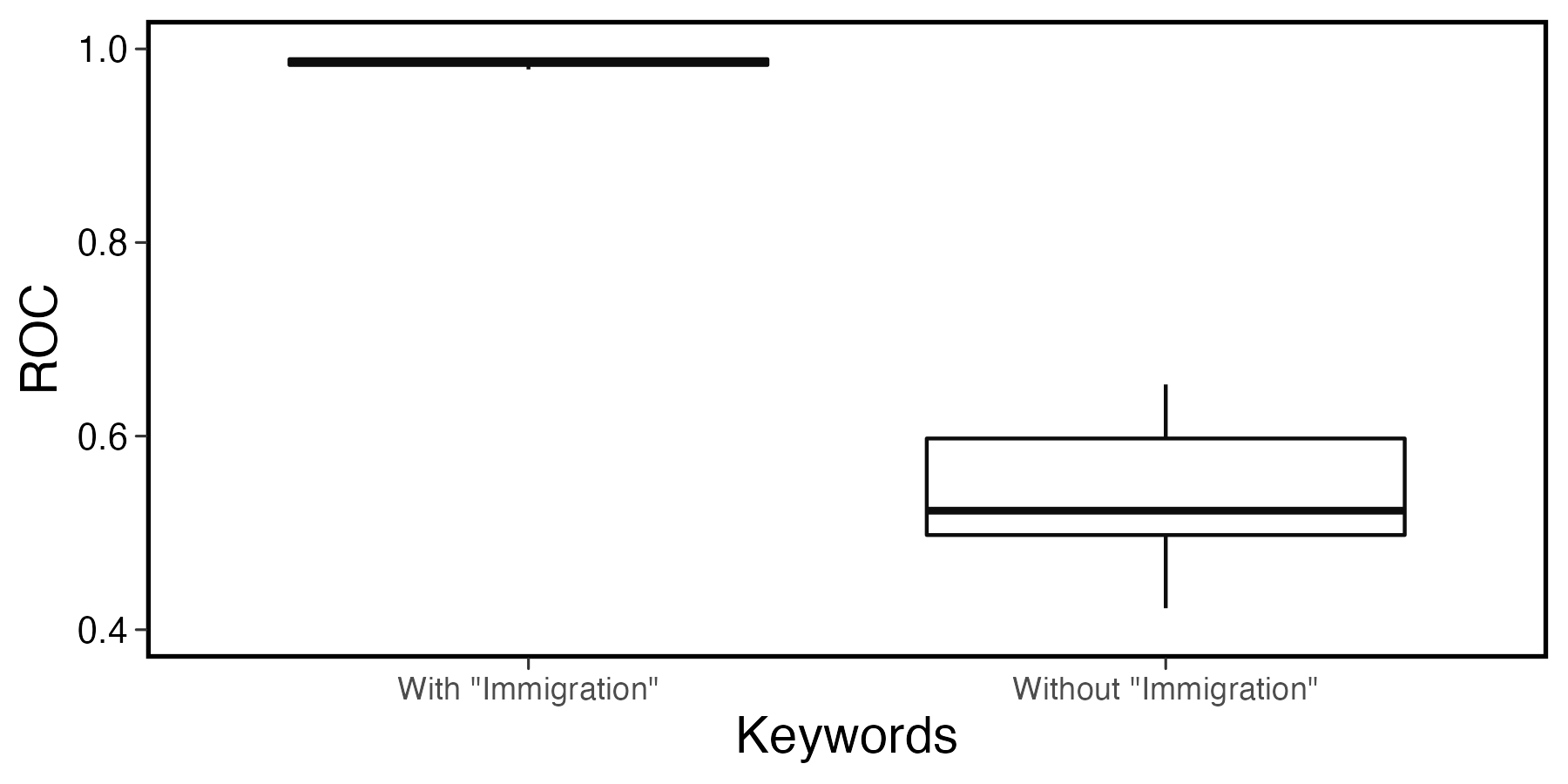}
  \caption{
    \textbf{Comparison of AUROC for the
  \textit{Immigration} topic between the keyword set with and without ``immigration''. }
  The figure indicates that the topic classification performance is much higher
  when the randomly selected keyword sets includes
  ``immigration'' as one of the keywords for the \textit{Immigration} topic.}
  \label{app:fig-random-keyword-topic9}
\end{figure}

  We further conduct a systematic investigation about when the
  \textit{Immigration} topic performs poorly.
  Figure~\ref{app:fig-random-keyword-topic9} shows the AUROC based on
  the previous simulation for the \textit{Immigration} topic with and
  without the keyword ``immigration'' in the keyword set.  The results
  show that the AUROC for the \textit{Immigration} topic is
  substantially lower when the word ``immigration'' is removed from
  the keyword set. This result indicates that whether ``immigration''
  is included in the \textit{Immigration} topic determines the quality
  of topic classification.

\subsection{Convergence}\label{app:base-convergence}

We present the Gelman-Rubin statistic for
perplexity of the base \keyATM{} with different number of
iterations to assess convergence.  We drop the first third of
iterations as burn-in.  The statistic monotonically decreases and
is reduced to 1.08 after 300,000 iterations.

\begin{figure}[!ht]
  \centering
  \includegraphics[width = 0.7\linewidth]{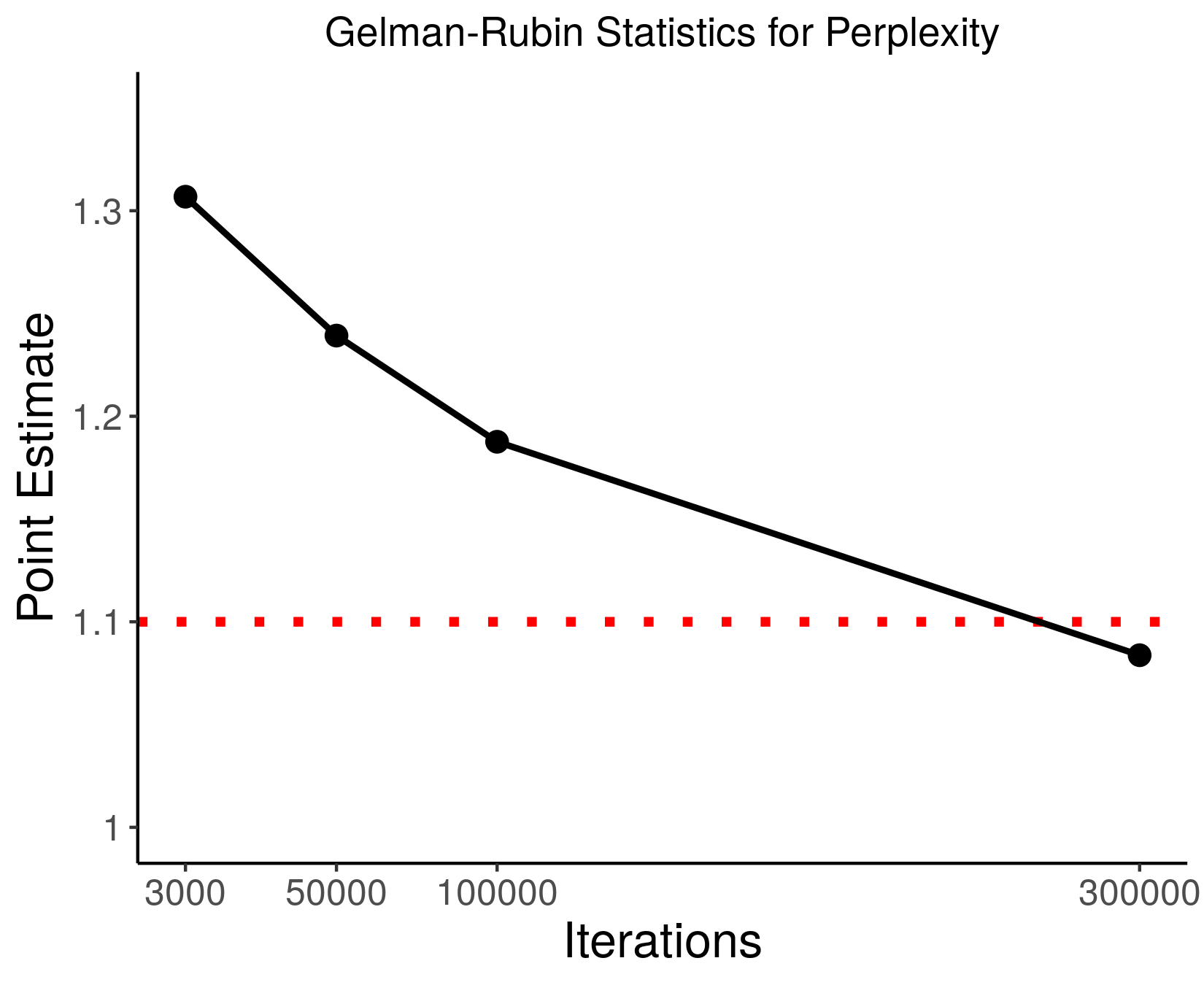}
  \caption{
    \textbf{Gelman-Rubin statistic for perplexity of the base \keyATM{}
  with different number of iterations. }
  The x-axis and y-axis indicate the number of iterations and Gelman-Rubin statistic respectively.
  The figure shows that the statistic decreases monotonically as the number of iterations increases.}
  \label{app:fig-base-convergence}
\end{figure}

\clearpage
\section{Additional Information for the Covariate \keyATM{}}

\subsection{Background}\label{app:CovBackground}
In 1994, the government reformed the
electoral system that lasted since 1947. The new electoral system,
which was first applied in the 1996 election, has two tiers:
single-member districts (SMD) and proportional representation (PR)
systems. The allocation of SMD seats is independent of that of PR
seats, making the Japanese electoral institution a mixed-member
majoritarian (MMM) system. \citet{Catalinac2015} finds that the
introduction of MMM changed the electoral strategies of LDP
candidates, focusing more on programmatic policies rather than pork
barrel.

\subsection{Keyword Construction Process}
\label{app:CovKeywords}

The application in Section~\ref{subsec:cov-empirical} does not have
human-coded topics. \citet{Catalinac2015} applies \wLDA{} and then
labels all topics. We do not use the most frequent words from the
original output because that would lead to analyze the same data as
the one used to derive keywords.  Instead, we independently construct
keywords using the survey questionnaires of the UTokyo-Asahi Surveys
(UTAS). This section explains the three steps of obtaining keywords
from UTAS: categorizing all labels in \citet{Catalinac2015} into 21
policy areas, matching 21 policy areas with UTAS questions, and
selecting keywords from the question.

Prior to every Upper and Lower
House election since 2003, the UTAS has fielded a survey to all
candidates (both incumbents and challengers) and published its results
in the newspaper.  Because of its publicity, the survey has a
remarkably high average response rate of approximately 85 \%
\citep{hirano:etal:11}.  Because the UTAS is designed to measure policy
issues relevant across multiple elections, it serves as an ideal
source of keywords.

\citet{Catalinac2015} fits \wLDA{} with 69 topics and uses 66 of
them as pork barrel or programmatic policy topics, excluding three
topics as credit claiming topics. Since the most of the topic labels
are granular and contain overlapping political issues, we first
categorize 66 topics into 21 policy areas to find corresponding
questions in the UTAS. Tables~\ref{table:app-cov-pork}
and~\ref{table:app-cov-programmatic} show all 66 topic labels taken
from \citet{Catalinac2015} in the middle column and 21 policy areas in
the left column.  Next, we find UTAS questions that represent each
policy area. This process results in the removal of five policy areas
(sightseeing, regional revitalization, policy vision, political
position, and investing more on human capital) that do not appear in
the UTAS.

Finally, we select keywords that represent each policy area from the
corresponding UTAS questionnaires.  Since most questions are fairly
short, we choose nouns so that they match with the preprocessed texts
(the preprocessing steps remove the Japanese conjugation).  The full
list of keywords is presented in Table~\ref{tab:cov-keywords}. The
original Japanese keywords are all single words, although some of them
become two English words after translation.

\begin{longtable}{p{2.4cm}p{5cm}p{8cm}}
\hline
  Policy Area                & Original Labels                         &  UTAS Question (Year)  \\ \hline
  Public Works               &  Appropriator for the district          &  \multirow{4}{\hsize}{Subsidies to the local government should be abolished in general (2003), It is necessary to secure employment by public works (2009)}    \\
                            &  Fixer-upper for the district           &    \\
                            &  Statesperson and appropriator          &    \\  
                            &                                         &    \\ \hline
  Road Construction          &  Transportation                         &   \multirow{5}{\hsize}{We should reduce the number of new highways and make existing ones free (2003), We should privatize four public highway corporations (2003), We should keep the road budget (2009)}         \\
                            &                                         &    \\
                            &                                         &    \\
                            &                                         &    \\ \hline
  Sightseeing                &  Primary industries and tourism         &   No corresponding questions \\
                            &  Health and leisure infrastructure      &    \\ \hline
  Regional                   &  Building a spiritually-rich community  &  No corresponding questions  \\
  Revitalization             &  Hometown development                   &         \\
                            &  Revitalizing the local community       &         \\
                            &  Local facilities and infrastructure    &         \\
                            &  Building a safe, reassuring community  &         \\
                            &  Benefits for organized groups          &    \\
                            &  Love of my hometown                    &         \\
                            &  Catching up with the rest of Japan     &         \\
                            &  Agriculture, forestry and fisheries    &    \\ \hline 
  & & \\ [-7pt]
  \caption{\normalsize \textbf{Pork barrel labels and UTAS questions}: The left
  column lists the policy areas created by the authors, whereas the middle
  column shows the original topic labels taken form
  \citet{Catalinac2015}. The right column presents the questions from the UTAS,
  translated from Japanese to English by
  authors.} \label{table:app-cov-pork}
\end{longtable}

\begin{longtable}{p{2.4cm}p{6cm}p{7cm}}
\hline
  Policy Area                 & Original Labels                              &  UTAS Question (Year)  \\ \hline

  Regional devolution         &  Regional devolution                            &  \multirow{5}{\hsize}{We should reduce the state subsidies and transfer tax revenue resources to municipalities (2003), We should merge municipalities (2003)} \\
                              &                                                 &    \\
                              &                                                 &    \\
                            &                                         &    \\  \hline
  Tax                         &  No more unfair taxes, peace constitution       & \multirow{6}{\hsize}{We should increase the consumption tax for stable pension system (2003), Do you agree or disagree with increasing the consumption tax for social security and financial reconstruction? (2005), It is inevitable to increase the consumption tax within five years (2009)}  \\
                              &  No tax increases, no U.S.-Japan alliance       & \\
                              &  Consumption tax is to fund the military        & \\
                              &  No tax increase, no constitutional revision    & \\
                              &  Tax cuts for everyone                          & \\
                              &  No consumption tax, no constitutional revision & \\ \hline
  Economic                    &  Economic recovery                              & \multirow{5}{\hsize}{It is urgent to finish the deflation, so instead of cutting budget for fiscal reconstruction, we should implement a fiscal stimulus program to boost the economy (2003, 2005)} \\
  recovery                    &  Economic stimulus                              & \\
                              &  Fiscal reconstruction                          & \\
                              &                                                 &    \\
                              &                                                 &    \\ \hline
  Global economy              &  Japan in global economy                        & \multirow{4}{\hsize}{We should protect the domestic industry (2009), We should promote trade and investment liberalization (2009)} \\
                              &                                                 &    \\
                              &                                                 &    \\ \hline
  Alternation                 & Political and administrative reform             & \multirow{3}{\hsize}{What would be the ideal political framework (2005),  In general alternation of government brings better politics (2009)}  \\
  of                          &  No more big business-favoritism                & \\
  government                  &  Political reform                               & \\
                              &  Reforming japan                                & \\
                              &  No more iron triangle                          & \\
                              &  Problems facing Japan                          & \\
                              &  Small government                               & \\
                              &  Alternation of government                      & \\\hline
  Constitution                &  Political reform, protect the constitution     & \multirow{2}{\hsize}{Do you think we should change the constitution? (2005)} \\
                              &                                                 &    \\ \hline
  Party                       &  No more LDP, no more public works              & \multirow{2}{\hsize}{Do you think it is good for your party to join the government coalition? (2009)} \\
                              &  Doing away with decayed LDP politics           & \\
                              &  No other party can be trusted                  & \\ \hline
  Postal                      &  Postal privatization                           & \multirow{8}{\hsize}{There are politicians who could not get an authorization from LDP because of the rebellion against postal privatization. If they form a new party, what this new party should do after the election? (2005), Do you support the postal privatization law, which was rejected in the Upper House? (2005)}  \\
  privatization               &  No postal privatization                        & \\
                              &  Post offices                                   & \\
                              &                                                 &    \\
                              &                                                 &    \\
                              &                                                 &    \\
                              &                                                 &    \\
                              &                                                 &     \\ \hline
  Inclusive                   &  Building a society kind to women               & \multirow{8}{\hsize}{The government should implement special policies to increase the number of women who have the higher positions and better jobs (2003), The basic form of the family consists of a couple and their children (2009), The forms of family should be diverse such as a single mother and DINKS (2009)}  \\
  society                    &                                                 &    \\
                              &                                                 &    \\
                              &                                                 &    \\
                              &                                                 &    \\
                              &                                                 &    \\
                              &                                                 &    \\
                              &                                                 &    \\ \hline
  Social                      &  Welfare and medical care                       & \multirow{3}{\hsize}{Even if it end up with lowering the quality of social welfare, the smaller government is better (2003, 2005, 2009)} \\
  welfare                     &  No reform of medical care                      & \\
                              &  Free medical care, no military spending        & \\
                              &  Nursing care                                   & \\ \hline
  Pension                     &  Protecting people                              & \multirow{6}{\hsize}{What do you think about merging the national pension, employees' pension, and the mutual aid pension systems? (2005), We should use tax for the universal pension (2009)} \\
                              &  Pensions and child allowance                   & \\
                              &  From roads to pension                          & \\
                              &                                                 &    \\
                              &                                                 &    \\
                              &                                                 &    \\ \hline
  Education                   &  Investing in young people                      & \multirow{3}{\hsize}{Education should respect the precedent methods rather than cultivating child's individuality (2009) }\\
                              &  Social security and child support              & \\
                              &  Better education and child-care facilities     & \\
                              &  Pensions and child allowance                   & \\ \hline
  Environment                 &  Saving the natural environment                 & \multirow{6}{\hsize}{We need to sacrifice the standard of living to protect the environment (2009), Environmental issues are not as important as to sacrifice the standard of living (2009)} \\
                              &  Earthquakes and nuclear accidents              & \\
                              &                                                 &    \\
                              &                                                 &    \\
                              &                                                 &    \\
                              &                                                 &    \\ \hline
  Security                    &  Foreign and national security policy           & \multirow{17}{\hsize}{Japan should strengthen its defense capacity (2003, 2005), Japan should not have nuclear weapons (2003), Japan-US alliance should be strengthened (2003, 2005), We should not hesitate a preemptive attack if there is an expected attack from (2003, 2005), Japan should join the United Nations Security Council to actively play an international role (2003, 2005), We should use dialogue rather than pressures to North Korea (2005, 2009), The government should change the interpretation of the constitution to exert the right to collective defense (2005), Japan-US alliance is the basis of the Japanese diplomacy (2009), Japanese diplomacy should be centered around the United Nations (2009)}  \\
                              &  No American bases                              & \\
                              &  Opposition to military spending                & \\
                              &  Stubbornly for peace and human rights          & \\
                              &  Security and reassurance                       & \\
                              &  Not a strong military but a kind society       & \\
                              &                                                 &    \\
                              &                                                 &    \\
                              &                                                 &    \\
                              &                                                 &    \\
                              &                                                 &    \\
                              &                                                 &    \\
                              &                                                 &    \\
                              &                                                 &    \\
                              &                                                 &    \\
                              &                                                 &    \\
                              &                                                 &    \\ \hline
  Policy vision               &  Vision for Japan                               & No corresponding questions \\
                              &  Politics for the civilian, not for bureaucrats & \\ \hline
  Political position          &  Liberal democracy is best!                     & No corresponding questions \\ \hline
  \multirow{3}{\hsize}{Investing more on human capital}  &  From concrete to people                        & No corresponding questions  \\
                              &                                                 &    \\
                              &                                                 &    \\ \hline
  \multicolumn{3}{c}{} \\[-7pt]
  \caption{\normalsize \textbf{Programmatic policy labels and UTAS questions}: The
  left column lists the policy areas created by the authors, whereas the middle
  column shows the original topic labels taken form
  \citet{Catalinac2015}. The right column presents the questions from the
  UTAS, translated from  Japanese to English by the
  authors.} \label{table:app-cov-programmatic}
\end{longtable}

\subsection{Top Words}\label{app:cov-topwords}

\begin{longtable}{p{2.1cm}p{2.3cm}|p{2.6cm}p{2.1cm}|p{3.1cm}p{2.3cm}}
  \hline
  \hline
  \multicolumn{2}{c|}{Public works} & \multicolumn{2}{c|}{Road construction} & \multicolumn{2}{c}{Regional devolution}\\ \hline \keyATM{} & \STM{} & \keyATM{} & \STM{} & \keyATM{} & \STM{}\\ \hline
  politic        & politic      & development     & tax          & reform              & reform   \\
  Japan          & Japan        & \textbf{road}   & reduced tax  & \textbf{rural area} & administration    \\
  society$^*$    & society      & city            & yen          & administration      & tax    \\
  citizen        & livelihood   & construction    & housing      & tax$^*$             & civilian    \\
  protect        & citizen      & tracks          & realize      & Japan               & consumption    \\
  livelihood     & protect      & \textbf{budget} & daily life   & politic             & rural area    \\
  secure         & secure       & realize         & move forward & citizen             & fiscal policy    \\
  LDP            & budget       & promote         & city         & country             & devolve    \\
  \textbf{works} & constitution & move forward    & education    & society$^*$         & nursing    \\
  \textbf{public}& LDP          & early           & measure      & consumption$^*$     & bureaucrat    \\
        \hline  \hline
  \multicolumn{2}{c|}{Tax} & \multicolumn{2}{c|}{Economic recovery} & \multicolumn{2}{c}{Global economy}\\ \hline \keyATM{} & \STM{} & \keyATM{} & \STM{} & \keyATM{} & \STM{}\\ \hline
  Japan                 & Japan        & reform                    & reform     & development              & development \\
  \textbf{tax}          & citizen      & \textbf{measure}          & postal     & \textbf{industry}        & community \\
  citizen               & JCP          & society$^*$               & privatize  & promote                  & road \\
  JCP                   & politic      & Japan                     & Japan      & prefecture               & promote \\
  \textbf{consumption}  & tax          & \textbf{economic climate} & rural area & agriculture              & industry \\
  politic               & consumption  & reassure                  & country    & plan                     & street \\
  \textbf{tax increase} & tax increase & economy                   & citizen    & community                & agriculture \\
  oppose                & oppose       & institution               & safe       & agriculture and forestry & prefecture \\
  business              & business     & safe                      & government & fishery                  & promote \\
  protect               & protect      & support                   & pension    & enrich                   & plan \\
  \hline
  \hline   \multicolumn{2}{c|}{Alternation of government} & \multicolumn{2}{c|}{Constitution} & \multicolumn{2}{c}{Party}\\ \hline \keyATM{} & \STM{} & \keyATM{} & \STM{} & \keyATM{} & \STM{}\\ \hline
  \textbf{government}  & yen         & \textbf{constitution} & consumption & Japan          & politic \\
  \textbf{alternation} & citizen     & consumption$^*$        & tax         & politic        & business \\
  yen                  & politic     & tax$^*$                & politic     & JCP            & LDP \\
  citizen              & Japan       & protect                & abolish     & LDP            & citizen \\
  politic              & medical     & tax increase$^*$       & citizen     & citizen        & Japan \\
  medical              & cost        & politic                & LDP         & business       & reform \\
  Japan                & government  & Japan                  & liberty     & protect        & donation \\
  trillion             & tax         & livelihood             & Japan       & \textbf{party} & JCP \\
  elderly              & trillion    & society$^*$            & rice        & donation       & plutocracy \\
  cost                 & consumption & peace                  & agriculture  & reform        & party \\
    \hline \hline\multicolumn{2}{c|}{Postal privatization} & \multicolumn{2}{c|}{Inclusive society} & \multicolumn{2}{c}{Social welfare}\\ \hline \keyATM{} & \STM{} & \keyATM{} & \STM{} & \keyATM{} & \STM{}\\ \hline
  \textbf{privatize} & tax increase  & politic              & politic    & politic          & politic \\
  tax increase$^*$   & constitution  & \textbf{civilian}    & reform     & \textbf{society} & rich \\
  \textbf{postal}    & tax           & society$^*$          & new        & Japan            & society \\
  yen                & consumption   & \textbf{participate} & realize    & reform           & hometown \\
  LDP                & protect       & peace                & citizen    & education$^*$    & building \\
  post               & oppose        & welfare$^*$          & government & rich             & welfare \\
  Japan              & Japan         & aim                  & daily life & building         & make effort \\
  oppose             & LDP           & human rights         & rural area & realize          & heart \\
  protect            & deterioration & realize              & corruption & century          & move forward \\
  ordinary people    & yen           & consumption$^*$      & change     & \textbf{welfare} & plan \\
  \hline \hline   \multicolumn{2}{c|}{Pension} & \multicolumn{2}{c|}{Education} & \multicolumn{2}{c}{Environment}\\ \hline \keyATM{} & \STM{} & \keyATM{} & \STM{} & \keyATM{} & \STM{}\\ \hline
  \textbf{pension}  & pension           & politic            & Japan     & \textbf{environment} & society \\
  yen               & institution       & Japan              & person    & society$^*$          & reassure \\
  institution       & yen               & person             & country   & education$^*$        & community \\
  wasteful spending & medical           & \textbf{children}  & politic   & realize              & building \\
  medical           & community         & \textbf{education} & necessary & community            & education \\
  community         & parenting         & country            & problem   & institution          & environment \\
  money             & wasteful spending & make               & children  & reassure             & support \\
  person            & support           & force              & force     & aim                  & measure \\
  abolish           & daily life        & have               & have      & move forward         & economic decline \\
  realize           & money             & problem            & future    & proceed              & employment \\
    \hline \hline
\end{longtable}
\vspace{-8.5mm}
\begin{longtable}{p{2.1cm}p{2.3cm}|p{2.6cm}p{2.1cm}|p{3.1cm}p{2.3cm}}
    \multicolumn{2}{c|}{Security} & \multicolumn{2}{c}{} & \multicolumn{2}{c}{}\\ \cline{1-2} \keyATM{} & \STM{} &  &  &  & \\ \cline{1-2}
  Japan                   & society & & & & \\
  \textbf{foreign policy} & Japan & & & & \\
  peace                   & world & & & & \\
  world                   & economy & & & & \\
  economy                 & environment & & & & \\
  country                 & international & & & & \\
  citizen                 & education & & & & \\
  \textbf{defense}        & country & & & & \\
  safe                    & peace & & & & \\
  international           & aim & & & & \\
      \hline \hline
\caption{\normalsize \textbf{Top words of the covariate \keyATM{}.}
  The table shows the top ten words with the highest estimated probability for each topic under each model. For \keyATM, the pre-specified keywords for each topic appear in bold letters whereas the asterisks indicate the keywords specified for another topic.}
\label{tab:app-cov-topwords}
\end{longtable}

\section{Additional Information for the Dynamic \keyATM{}}

\subsection{Introducing Time Dynamics with HMM}\label{app:dynamic-literature}
HMM has been used to introduce time dynamic components in various applications.
For example, \citet{zhai2014} extends LDA to a dynamic
setting with the HMM to analyze dialogues. Unlike this model, the
dynamic \keyATM{} allows multiple documents to share one state.
\citet{quinn2010} proposes an HMM-based dynamic topic model but only
allows for a single topic for each document and no keyword.  Although
the pioneering dynamic topic model of \citet{Blei2006} uses the Kalman
filter, this modeling strategy does not exploit conjugacy, and hence
the authors use an approximation to the posterior. Others such as
\citet{Wang2006} model a distribution of ``time stamps'' for
documents, but this approach is not ideal for social science research
because it does not directly model time trends.  Indeed, social
scientists have effectively used HMM in other settings
\citep[e.g.,][]{park:12,Knox2021,Olivella2022}.

\subsection{Keyword Construction Process}\label{app:dynamic-keywords}

The procedure is similar to the one for the base \keyATM{}
application.  We obtain keywords from the Supreme Court Database
project issue description available at their
website.\footnote{\url{http://www.supremecourtdatabase.org/documentation.php?var=issue}. Last
  accessed on March 10, 2020.}  After scraping the description from
the website, we lemmatize each word using the \textsf{Python} library
\texttt{NLTK} and remove stopwords via the \textsf{R} package
\texttt{quanteda} \citep{Benoit2018}.  We then remove words and
phrases that have little to do with the substance of each topic.  For
example, some topic description includes ``Note: '', which explains
the background information that does not relate to the substance of
topics or specifies the content that should be excluded from the
topic.  We do not include such descriptions when constructing
keywords.  Third, we keep the same keywords for multiple topics only
if their inclusion can be substantively justified.  Lastly, we limit
the number of keywords to 25 per topic.  We remove terms based on the
proportion of keywords among all terms in the corpus if the topic
contains more than 25 keywords.

\subsection{The Full List of Keywords}\label{app:dynamic-keywords-list}

\begin{longtable}{p{2.5cm}p{13.5cm}} \hline  \hline
Label & Keywords \\ \hline
Criminal procedure
& bank call constitutional construction counsel crime
criminal death evidence immunity jury justice line obtain
procedure prosecution remedy review right rule search
sentence statement trial witness\\ \hline
Civil rights         &
benefit civil constitutional cost counsel discrimination
duty employment equal file liability national party plan
political protection provision public requirement right school security subject suit \\ \hline
First amendment
& amendment bar benefit concern cost election employee failure first
form free legislative light material official party political private
public regulation religious requirement school security speech\\ \hline
  Due process
  & constitutional defendant due employee hear hearing impartial
  litigant maker notice prisoner process property resident right statutorily \\ \hline
Privacy
& abortion contraceptive die freedom information privacy regulation right \\ \hline
  Attorneys
  & admission attorney bar commercial compensation disbarment discipline employee
  fee license official speech \\ \hline
  Unions
  & activity agency antitrust bargaining discharge dispute election employee
  employer fair fund health injunction labor litigation member relation
  representative right safety standard strike trust union work  \\\hline
  Economic activity
& business civil claim company contract damage defense determination
employee employer evidence federal land liability local official
power process property protection public regulation remedy right tax\\ \hline
Judicial power
& agency appeal authority circuit claim district evidence federal
file grant ground judicial jurisdiction order part party power
present private procedure question review right rule suit \\ \hline
  Federalism
  & air child commerce conflict dispute enforcement family federal ground
  interpretation interstate land legislation national natural obligation
  primary property regulation relationship resource rest support tax water \\ \hline
Interstate relations
& boundary conflict dispute foreign incorporation interstate property relation territory\\ \hline
  Federal taxation
  & business claim entity expense federal fiscal gift internal personal priority
  private professional provision revenue supremacy tax taxation \\  \hline
Miscellaneous
& authority congress executive legislative veto\\ \hline
  Private action
  & civil commercial contract evidence personal procedure property real tort transaction trust\\ \hline \hline
\caption{\normalsize \textbf{Keywords for the dynamic \keyATM{}}: Keywords used in the dynamic \keyATM{} application}\label{table:app-dynamic-keywords}
\end{longtable}

\subsection{Top Words}\label{app:dynamic-topwords}

Table~\ref{tab:app-dynamic-topwords} presents the top
  ten words with the highest estimated probabilities defined in
  Equation~\eqref{eq:base-phi}.  The results show that \keyATM{}
  performs better than \wLDA{} in most topics.

\begin{longtable}{p{2.2cm}p{2.2cm}|p{2.2cm}p{2.2cm}|p{2.2cm}p{2.2cm}} \hline
  \hline
  \multicolumn{2}{c|}{Criminal procedure} & \multicolumn{2}{c|}{Economic activity} & \multicolumn{2}{c}{Civil rights}\\ \hline
  \keyATM{} & \wLDA{} & \keyATM{} & \wLDA{} & \keyATM{} & \wLDA{}\\ \hline
\textbf{trial} & trial & \textbf{federal} & commission & district$^*$ & school \\
  \textbf{jury} & jury & action & commerce & \textbf{school} & district \\
  defendant$^*$ & petitioner & \textbf{claim} & price & {\small \textbf{discrimination}} & religious \\
  \textbf{evidence} & evidence & \textbf{damage} & rate & election$^*$ & discrimination \\
  \textbf{criminal} & defendant & suit$^*$ & interstate & \textbf{equal} & county \\
  \textbf{sentence} & counsel & statute & market & county & election \\
  petitioner & right & \textbf{right} & carrier & vote & vote \\
  judge & rule & rule$^*$ & sale & \textbf{plan} & equal \\
  conviction & make & plaintiff & use & one & education \\
  \textbf{counsel} & judge & jurisdiction$^*$ & service & race & student \\
    \hline
  \hline  \multicolumn{2}{c|}{Judicial power} & \multicolumn{2}{c|}{First amendment} & \multicolumn{2}{c}{Due process}\\ \hline
  \keyATM{} & \wLDA{} & \keyATM{} & \wLDA{} & \keyATM{} & \wLDA{}\\ \hline
\textbf{appeal} & federal & \textbf{public} & public & child$^*$ & sentence \\
  \textbf{district} & claim & \textbf{amendment} & first & interest & offense \\
  \textbf{order} & district & \textbf{first} & speech & \textbf{right} & death \\
  petitioner & appeal & government & amendment & \textbf{process} & jury \\
  \textbf{federal} & action & may & interest & may & defendant \\
  \textbf{claim} & judgment & interest & party & \textbf{due} & crime \\
  \textbf{rule} & rule & \textbf{speech} & may & statute & criminal \\
  judgment & petitioner & right$^*$ & right & prison & penalty \\
  issue & jurisdiction & can & political & person & punishment \\
  proceeding & order & \textbf{religious} & government & parent & sentencing \\
      \hline
\hline    \multicolumn{2}{c|}{Federalism} & \multicolumn{2}{c|}{Unions} & \multicolumn{2}{c}{Federal taxation}\\ \hline
\keyATM{} & \wLDA{} & \keyATM{} & \wLDA{} & \keyATM{} & \wLDA{}\\ \hline
\textbf{land} & search & \textbf{employee} & employee & \textbf{tax} & tax \\
  \textbf{water} & officer & \textbf{union} & union & property$^*$ & property \\
  indian & police & board & labor & pay & income \\
  tribe & amendment & \textbf{labor} & employer & income & pay \\
  right$^*$ & arrest & \textbf{employer} & board & payment & bank \\
  \textbf{property} & warrant & agreement & agreement & interest & interest \\
  use & fourth & employment$^*$ & contract & benefit$^*$ & corporation \\
  reservation & person & contract$^*$ & employment & amount & payment \\
  \textbf{federal} & cause & \textbf{work} & bargaining & plan$^*$ & amount \\
  indians & evidence & \textbf{bargaining} & work & fund$^*$ & business \\
    \hline
  \hline  \multicolumn{2}{c|}{Privacy} & \multicolumn{2}{c|}{Attorneys} & \multicolumn{2}{c}{Interstate relations}\\ \hline
    \keyATM{} & \wLDA{} & \keyATM{} & \wLDA{} & \keyATM{} & \wLDA{}\\ \hline
search$^*$ & child & agency$^*$ & ante & commission & land \\
  officer & benefit & regulation$^*$ & rule & commerce$^*$ & water \\
  police & interest & use & can & \textbf{interstate} & indian \\
  amendment$^*$ & medical & standard$^*$ & standard & rate & tribe \\
  arrest & plan & congress$^*$ & whether & carrier & right \\
  warrant & provide & require & even & railroad & reservation \\
  fourth & parent & provide & opinion & service & property \\
  evidence$^*$ & woman & \textbf{fee} & decision & new & use \\
  person & may & rule$^*$ & dissent & gas & indians \\
  use & statute & secretary & apply & line$^*$ & river \\
      \hline
  \hline \multicolumn{2}{c|}{Miscellaneous} & \multicolumn{2}{c|}{Private action}\\ \hline
  \keyATM{} & \wLDA{} & \keyATM{} & \wLDA{} \\ \hline
\textbf{congress} & congress & price & power  \\
  statute & provision & market & right \\
  power$^*$ & section & sale & government \\
  federal$^*$ & agency & business$^*$ & congress\\
  government & provide & company$^*$ & federal \\
  provision$^*$ & federal & sell & amendment \\
  shall & shall & product & constitution \\
  section & regulation & competition & statute \\
  cong & statute & antitrust$^*$ & clause  \\
  committee & committee & patent & may  \\ \hline \hline
  \multicolumn{6}{c}{} \\[-7pt]
  \caption{\normalsize \textbf{Top words of the dynamic \keyATM{}.}
  The table shows the top ten words with the highest estimated probability for each topic under each model. For \keyATM, the pre-specified keywords for each topic appear in bold letters whereas the asterisks indicate the keywords specified for another topic.}
\label{tab:app-dynamic-topwords}
\end{longtable}

\subsection{ROC Curves}\label{app:dynamic-roc}

\begin{figure}[!htb]
\centering
  \includegraphics[width= 0.9\linewidth]{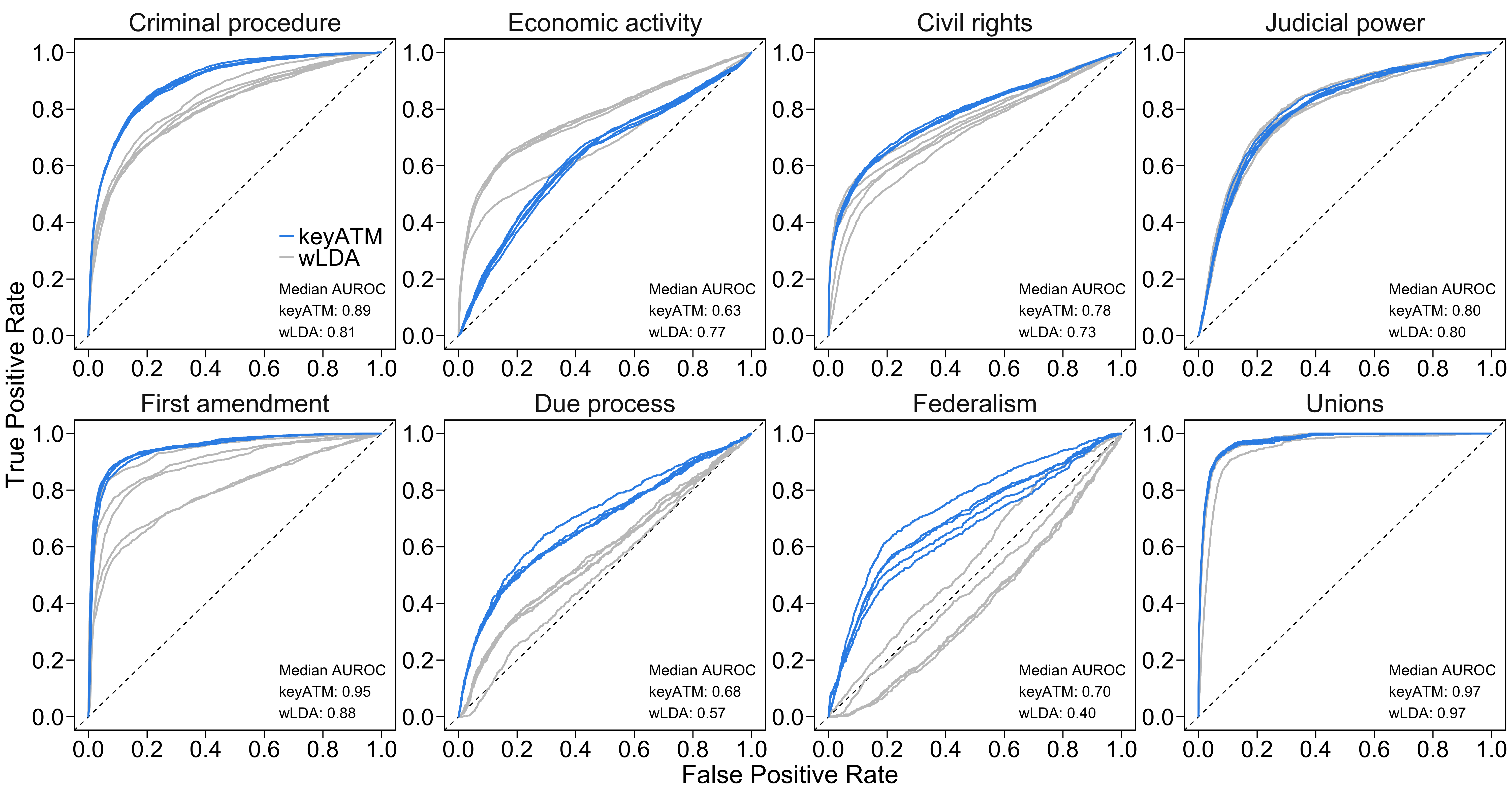}

  \includegraphics[width= 0.65 \linewidth]{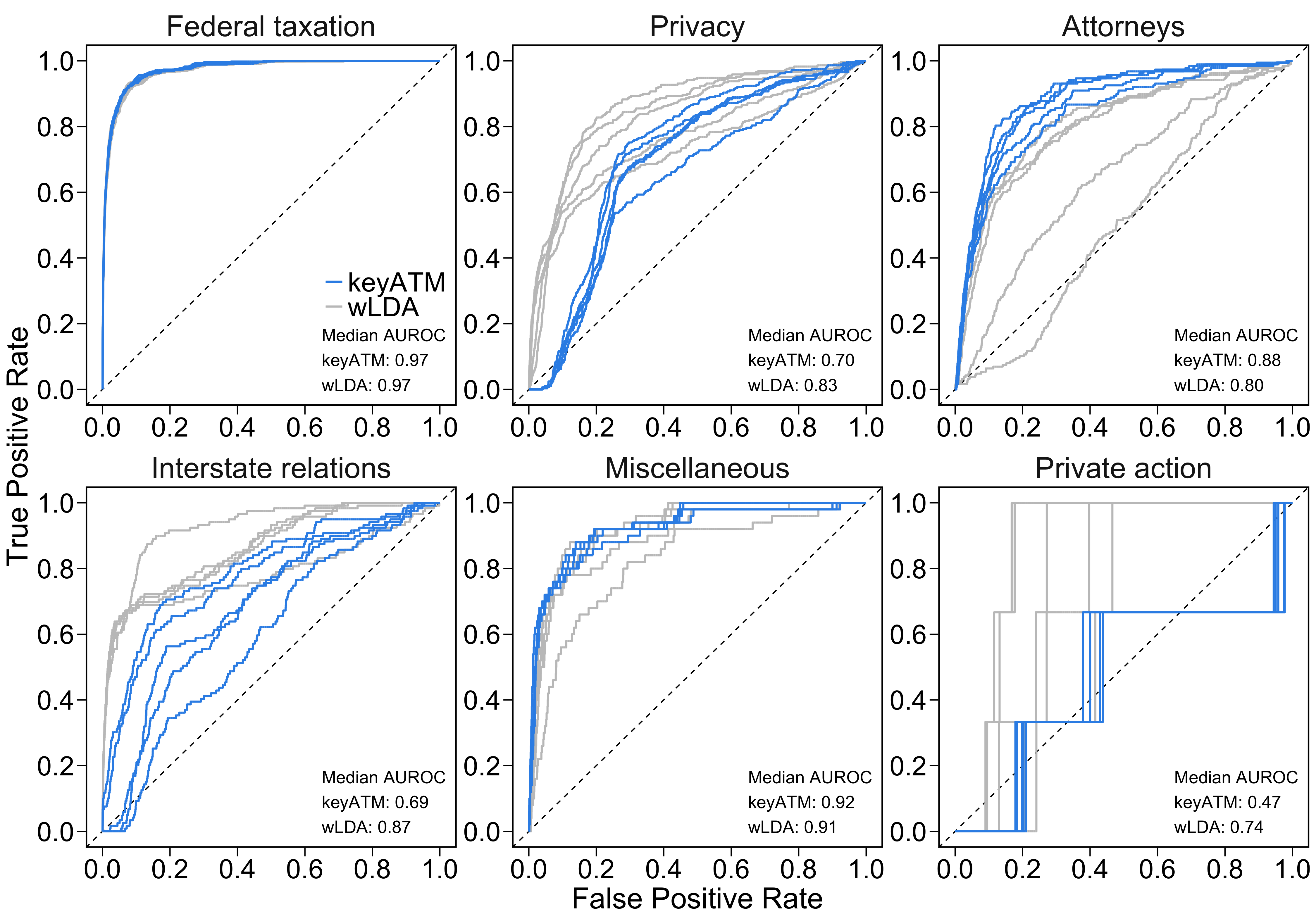}
\caption{\textbf{Comparison of the ROC curves between the dynamic keyATM and \wLDA{}}}
\label{app:fig:dynamic-roc}
\end{figure}

Figure~\ref{app:fig:dynamic-roc} shows ROC curves.  Each
  line represents the ROC curve from one of the five Markov chains
  with different starting values for \keyATM{} (blue lines) and
  \wLDA{} (grey lines).  The median AUROC indicates the median value
  of AUROC among five chains for each model.  \keyATM{} performs at
  least as well as \wLDA{} in 11 out of 14 topics.

  \subsection{Correlation of topic prevalence}\label{app:dynamic-cor}
  Table~\ref{app:dynamic-correlation-table} summarizes the results of
  the the correlation between the estimated topic prevalence and the
  SCD human coding.  To compare the estimated topic prevalence and the
  SCD human coding, we use the standardized measure that subtracts its
  mean from each data point and then divide it by its standard
  deviation.  The results show that \keyATM{} exhibits a higher
  correlation with the human coding for most topics (9 out of 13) than
  \wLDA.  Note that the result for \textit{Private action} is not
  shown because, according to the SCD codebook, all documents
  associated with this topic are from the same year (2012).

\begin{table}[!ht]
  \centering
  \begin{tabular}{lcc}
    \hline
    Topic & Dynamic \keyATM & Dynamic \wLDA \\
    \hline
    Criminal procedure & 0.83 & 0.07 \\
    Economic activity & 0.46 & 0.44 \\
    Civil rights & 0.76 & 0.70 \\
    Judicial power & 0.24 & 0.40 \\
    First amendment & 0.82 & 0.64 \\
    Due process & 0.37 & 0.11 \\
    Federalism & 0.11 & 0.04 \\
    Unions & 0.78 & 0.74 \\
    Federal taxation & 0.80 & 0.79 \\
    Privacy & 0.07 & 0.18 \\
    Attorneys & 0.18 & 0.25 \\
    Interstate relations & 0.24 & 0.34 \\
    Miscellaneous & 0.12 & -0.02 \\
      \hline
  \end{tabular}
  \caption{\textbf{Comparison of the time trends of
        topic prevalence between the dynamic \keyATM{} / \wLDA{} and
        the SCD human coding.} The table shows the correlation between
      the estimated topic prevalence and the SCD human coding. To
      compare the estimated topic prevalence and the SCD human coding,
      we use the standardized measure that subtracts its mean from
      each data point and then divide it by its standard deviation.} \label{app:dynamic-correlation-table}
\end{table}

\subsection{Quality of Keywords}\label{app:dynamic-quality}
\begin{figure}[!h]
  \spacingset{1}
  \centering
\includegraphics[width= 0.7\linewidth]{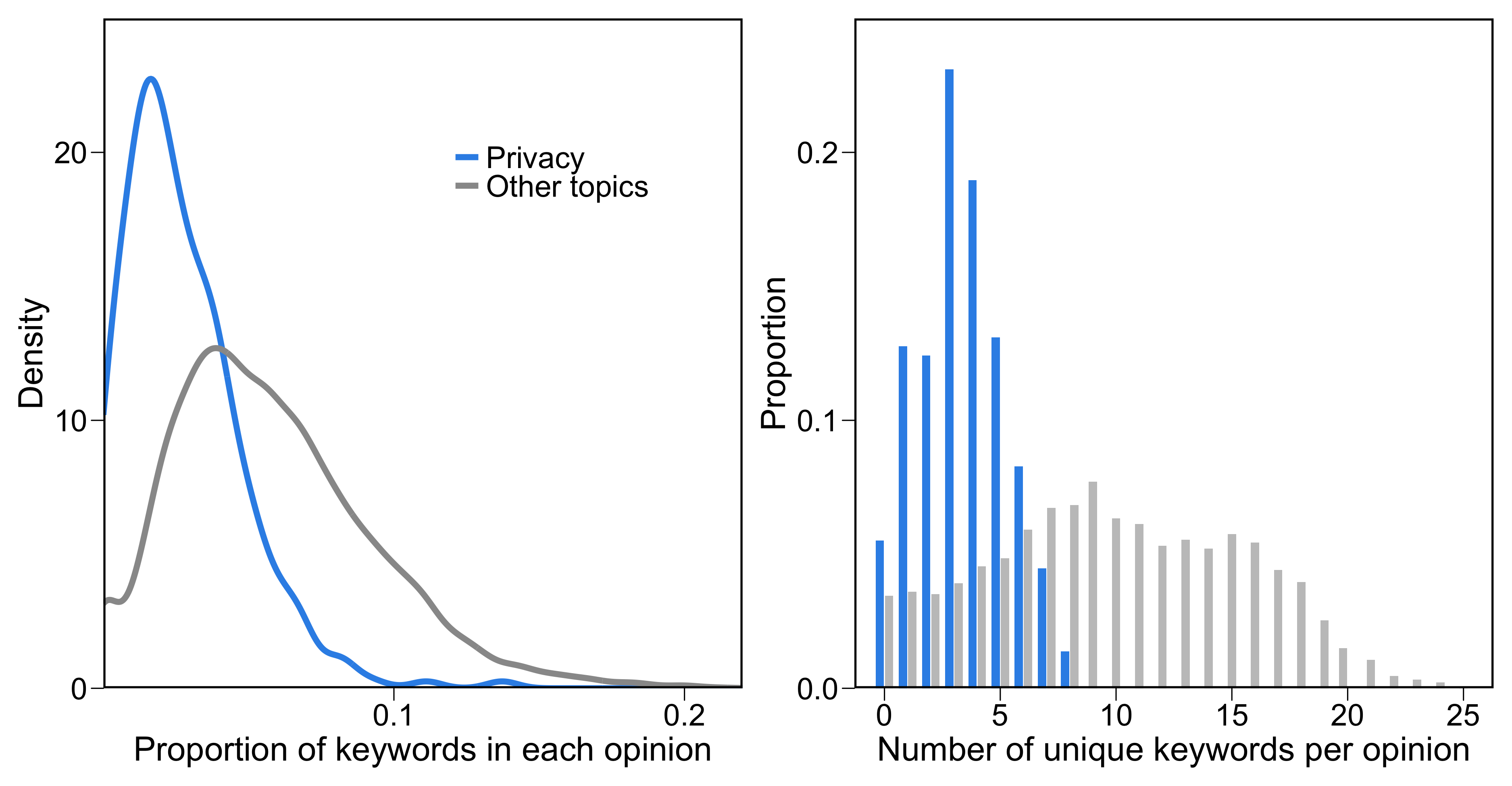}
\caption{\textbf{Poor quality of keywords for the \textit{Privacy}
    topic.}  The left panel presents the histogram for the proportion
  of keywords among all words in each of the opinions that are
  classified to the {\it Privacy} topic (blue bars) by the Supreme
  Court Database (SCD) human coders.  Compared to the other five
  topics (grey bars) from Table~\ref{tab:dynamic-topwords}, the
  keywords appear less frequently in the {\it Privacy} topic.  The
  right panel shows the histogram for the number of unique keywords
  that appear in each of the opinions associated with the {\it
    Privacy} topic by SCD.  The number of unique keywords are smaller
  in the documents classified to the {\it Privacy} topic than the
  other five topics.}
\label{fig:dynamic-keys}
\end{figure}

Similar to the result of Section~\ref{subsubsec:quality}, the quality
of keywords seems to matter for the poor results of the
interpretability and classification by the dynamic \keyATM{} for the
{\it Privacy} topic.  The left panel of Figure~\ref{fig:dynamic-keys}
shows the histogram for the proportion of keywords in each of the
opinions that are classified to the {\it Privacy} topic by the SCD
human coding.  The keywords for the {\it Privacy} topic appear much
less frequently than the other selected topics from
Table~\ref{tab:dynamic-topwords} (grey bars).  The right panel of the
figure presents the histogram for the number of unique keywords that
are contained in each opinion associated with the {\it Privacy} topic
and other five topics by SCD.  The figure indicates that on average
fewer keywords appear in the {\it Privacy} related opinions compared
to the corresponding keywords for the other five topics.  As similar
to the application of the base \keyATM, these results suggest that
researchers need to pay a careful attention to the selection of
keywords when fitting \keyATM.

%


\section{Additional Information for the Topic Matching of \wLDA{}}\label{app:topicmatch}
For \keyATM, there is no need to label topics with pre-specified
keywords after model fitting.  In contrast, \wLDA{} requires the
post-hoc labeling of the resulting topics.  Here, we determine the
topic labels such that the document classification performance of
\wLDA{} is maximized.

We use the Hungarian algorithm to match
the topic labels to the resulting topics by maximizing the area
under the receiver operating characteristics (AUROC).  To account
for the multi-label classification structure, we calculate the
harmonic mean of binary classification measures across each label
with the \textsf{R} package \texttt{multiROC} \citep{Wei2018}.

\section{Additional Information for the Validation}\label{app:validation}

\subsection{Design}\label{app:validation-design}

We recruited participants via Amazon Mechanical Turk (for English
tasks) and CrowdWorks (for Japanese tasks). Participants from both
platforms conducted tasks on Qualtrics, which makes recruiting
platform the only difference between English and Japanese tasks.

\begin{figure}[!h]
  \spacingset{1}
  \centering
\includegraphics[width=0.7\linewidth]{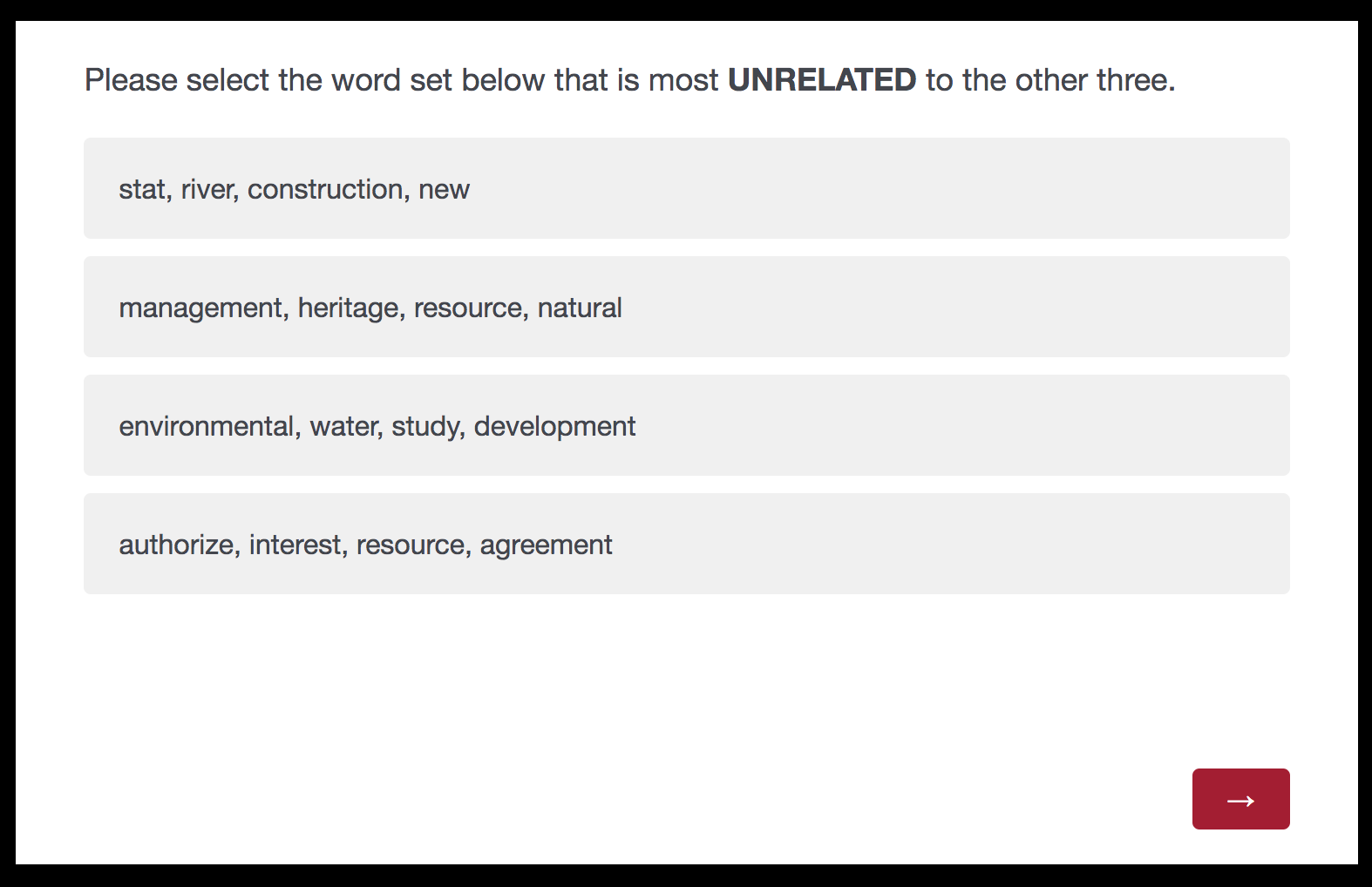}
\caption{\textbf{Screen shot of a task from the coherency validation exercise.}}
\label{fig:validation-screenshot-R4WSI}
\end{figure}

\begin{figure}[!h]
  \spacingset{1}
  \centering
\includegraphics[width=0.7\linewidth]{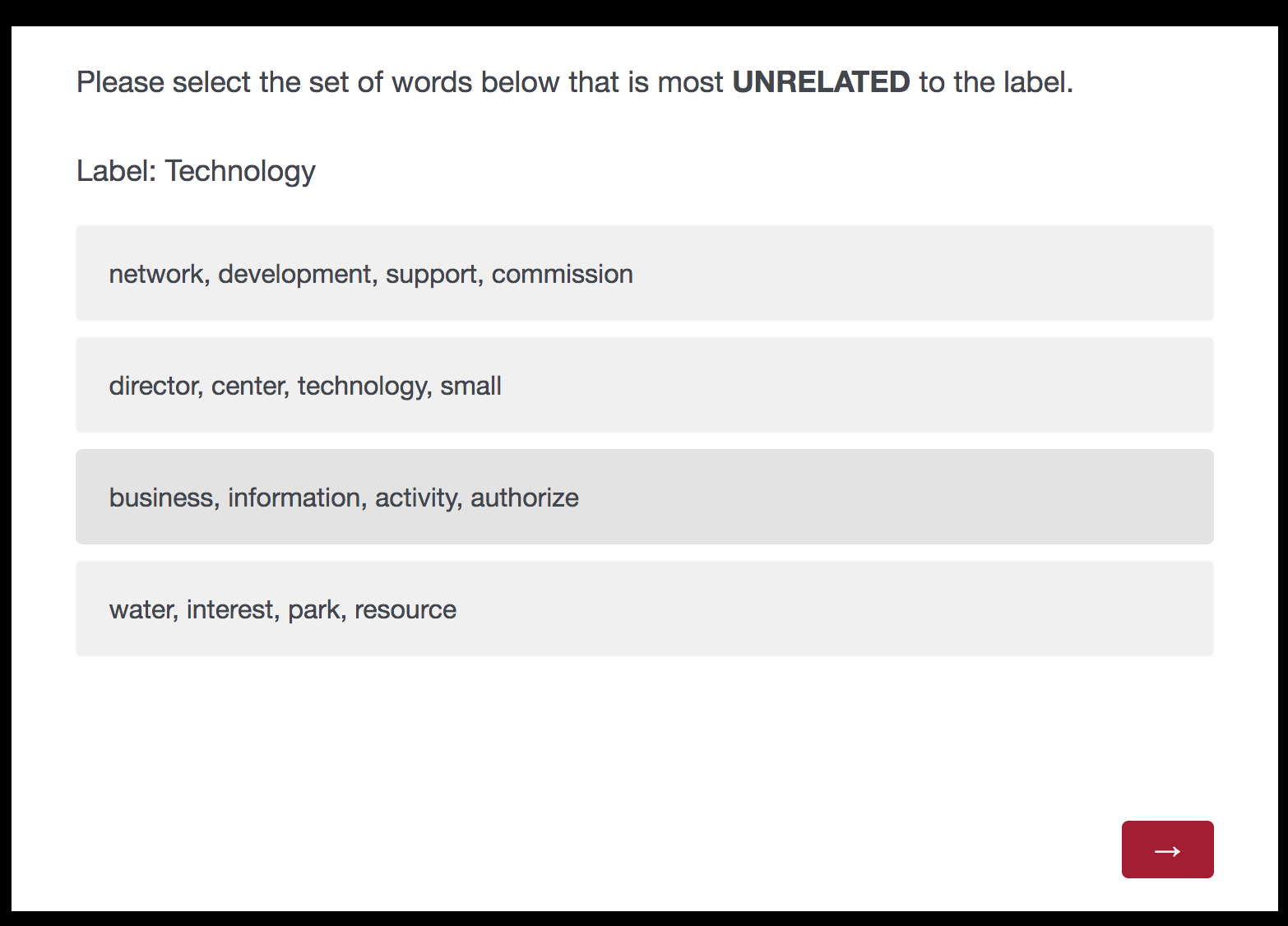}
\caption{\textbf{Screen shot of a task from the coherency-and-label validation exercise.}}
\label{fig:validation-screenshot-R4WSIwL}
\end{figure}

Once they agree on the consent form, participants read the instruction
and start answering them.  We modify methods proposed by
\citet{Ying2021} to implement our validation exercises.  The package
to implement their method is available at
\url{https://github.com/Luwei-Ying/validateIt}. In both
  coherency design and coherency-and-label design, participants choose
  one option per each task.  For each participant, we randomize tasks
  for \keyATM{} and baseline models as well as the order of choice
  options in order to deal with the possible ordering effect.  Each
  participant works on 11 tasks of the same design from a single
  empirical application, five for the output of \keyATM, another five
  for the output of \wLDA, and one gold-standard task to assess the
  worker quality.  Examples of coherency design and coherency-and-label
  design are shown in
  Figures~\ref{fig:validation-screenshot-R4WSI}~and~\ref{fig:validation-screenshot-R4WSIwL},
  respectively.

  We track the quality of workers by randomly inserting a
  gold-standard task for each exercise. We prepare the gold-standard
  tasks for each of three applications separately so that the
  gold-standard tasks mimic texts used in three applications.  We
  carefully chose the gold-standard tasks so that they are simple and
  unambiguous.  Examples of gold-standard tasks are shown in
  Table~\ref{tab:validation-gold}.  In each task, we drop all
  responses given by workers who fail to provide the correct answers
  to any of gold-standard tasks they conduct.

  In addition to the coherency and coherency-and-label design,
    we also tried topic intrusion design (T8WSI) using the
    covariate \keyATM{} application.  In this task, each worker is
    shown an actual document and four sets of words.  Each of the four
    word sets contains the eight highest probability words for a topic
    (i.e., four sets of eight words).  Three of these topics
    correspond to the highest probability topics for the displayed
    document, while one is a low probability for that document.  The
    covariate \keyATM{} application is the most suitable for T8WSI
    because reading the candidate manifestos does not require expert
    knowledge.  Unfortunately, the number of workers who provided the
    correct answer to our gold-standard task is much lower for T8WSI
    compared to the coherency task (R4WSI) and the coherency-and-label
    task (modified R4WSI).  Indeed, 18.7\% of workers (17 out 91)
    failed to provide the correct answer in topic intrusion task
    whereas the number is much lower for coherency task (6.7\%, 6 out
    of 90) and coherency-and-label task (8.0\%, 7 out of 87).
    This finding is consistent with
    the observation made by \citet{Ying2021} who wrote, ``In our initial
    testing, we found that the WI and T8WSI tasks were often too
    difficult for coders, reducing their power to discriminate.
    Further, T8WSI is sensitive to the words included in the `top
    eight,' making the results more arbitrary and again less
    informative.''

\begin{table}[!h]
\spacingset{1}
\centering
\begin{tabular}{lll}
\hline
                  & Coherency & Coherency-and-label \\ \hline
Base model (Legislative bills) &                 105 &                                 109 \\
Covariate model (Manifestos)        &                 80 &                                 84 \\
Dynamic model (Court opinions)    &                 125 &                                 110 \\ \hline
\end{tabular}
  \caption{\textbf{Number of workers assigned to each design}. Note that we
  ask workers to complete 11 tasks of the same kind from
  a single empirical application: five tasks for \keyATM, five tasks
  for the baseline model, and one gold-standard task.
  }
\end{table}

\begin{table}[!h]
  \spacingset{1}
  \centering
  \begin{tabular}{llllll}
    \hline
    Method & Label & Option 1 & Option 2 & Option 3 & Correct choice \\ \hline
    Coherency & NA & health, & coverage, & hospital, & energy, \\
    & & period, & insurance, & cost, & gas, \\
    & & medical,  & part,  & provider,  & vehicle, \\
    & & medicare & care & payment & oil \\\hline
    Coherency-and-Label & Transportation & airport, & safety, & transit, & college\\
    & & car, & traffic, & air, & education, \\
    & & rail, & aviation, & motor, & excellence, \\
    & & ship & carry & passenger & language \\ \hline

  \end{tabular}
  \caption{\textbf{Examples of the gold-standard tasks.} For both methods,
  first three options are from the same topic and the last choice is from another one.
  Note that we randomize the order of options in the actual tasks.
  }
  \label{tab:validation-gold}
\end{table}

\subsection{Alternative design}\label{app:validation-alternative}
We do not include Word Intrusion (WI) and Top 8 Word Set
Intrusion (T8WSI) in our validation exercise because, as
\citet{Ying2021} note, both are difficult tasks, and the latter is
particularly known to be sensitive to the choice of displayed
words.

We also tried T8WSI using the
    covariate \keyATM{} application.  In this task, each worker is
    shown an actual document and four sets of words.  Each of the four
    word sets contains the eight highest probability words for a topic
    (i.e., four sets of eight words).  Three of these topics
    correspond to the highest probability topics for the displayed
    document, while one is a low probability for that document.  The
    covariate \keyATM{} application is the most suitable for T8WSI
    because reading the candidate manifestos does not require expert
    knowledge.  Unfortunately, the number of workers who provided the
    correct answer to our gold-standard task is much lower for T8WSI
    compared to the coherency task (R4WSI) and the coherency-and-label
    task (modified R4WSI).  Indeed, 18.7\% of workers (17 out 91)
    failed to provide the correct answer in topic intrusion task
    whereas the number is much lower for coherency task (6.7\%, 6 out
    of 90) and coherency-and-label task (8.0\%, 7 out of 87).

\subsection{Descriptive Statistics}\label{app:validation-desc-stats}
This section lists descriptive statistics by model and exercise.
Each topic has about twenty to thirty tasks.  The number
  of tasks is not fixed because we randomly chose tasks to present.
  In
  Figures~\ref{fig:validation-main}~and~\ref{app:fig:validation-appendix},
  we show both pooled and topic-by-topic results.

\begin{table}[!ht]
\centering
\begin{tabular}{lrrrr}
  \hline
  \hline & \multicolumn{2}{c}{\keyATM} & \multicolumn{2}{c}{\wLDA}\\ \hline Topic &  \# of correct tasks & \# of tasks & \# of correct tasks & \# of tasks\\ \hline
Agriculture &  18 &  24 &  18 &  24 \\
  Civil rights &  13 &  23 &   9 &  23 \\
  Culture &  21 &  27 &  16 &  27 \\
  Defense &  19 &  23 &  13 &  23 \\
  Domestic commerce &  22 &  30 &  22 &  30 \\
  Education &  19 &  21 &  16 &  21 \\
  Energy &  40 &  42 &  32 &  42 \\
  Environment &  21 &  31 &  21 &  31 \\
  Foreign trade &  13 &  25 &  21 &  25 \\
  Government operations &  17 &  29 &  25 &  29 \\
  Health &  17 &  20 &  17 &  20 \\
  Housing &  11 &  17 &  13 &  17 \\
  Immigration &  12 &  16 &   8 &  16 \\
  International affairs &  23 &  26 &  23 &  26 \\
  Labor &  16 &  24 &  12 &  24 \\
  Law and crime &  30 &  33 &  24 &  33 \\
  Macroeconomics &   8 &  19 &   5 &  19 \\
  Public lands &  13 &  17 &  14 &  17 \\
  Social welfare &  14 &  23 &  15 &  23 \\
  Technology &  13 &  18 &   8 &  18 \\
  Transportation &  30 &  37 &  29 &  37 \\ \hline
  Total & 390 & 525 & 361 & 525 \\ \hline
    \hline
\end{tabular}
\caption{\textbf{Summary statistics for the coherency validation
exercises with base model (legislative bills).}
This table shows the number of correct tasks and the number of total tasks
for each model and for each topic.}
\end{table}

\begin{table}[!ht]
\centering
\begin{tabular}{lrrrr}
  \hline
  \hline & \multicolumn{2}{c}{\keyATM} & \multicolumn{2}{c}{\wLDA}\\ \hline Topic &  \# of correct tasks & \# of tasks & \# of correct tasks & \# of tasks\\ \hline
Agriculture &  15 &  18 &  16 &  18 \\
  Civil rights &  13 &  23 &  12 &  23 \\
  Culture &  11 &  26 &   4 &  26 \\
  Defense &  20 &  24 &  20 &  24 \\
  Domestic commerce &  24 &  31 &  19 &  31 \\
  Education &  30 &  32 &  29 &  32 \\
  Energy &  20 &  25 &  16 &  25 \\
  Environment &  10 &  27 &  17 &  27 \\
  Foreign trade &  12 &  26 &  11 &  26 \\
  Government operations &   6 &  25 &  18 &  25 \\
  Health &  25 &  27 &  25 &  27 \\
  Housing &  19 &  22 &  17 &  22 \\
  Immigration &  23 &  29 &  19 &  29 \\
  International affairs &  25 &  29 &  25 &  29 \\
  Labor &  18 &  26 &  13 &  26 \\
  Law and crime &  28 &  30 &  15 &  30 \\
  Macroeconomics &  15 &  28 &  10 &  28 \\
  Public lands &  15 &  19 &  16 &  19 \\
  Social welfare &  14 &  25 &   9 &  25 \\
  Technology &  20 &  28 &  16 &  28 \\
  Transportation &  22 &  25 &  18 &  25 \\ \hline
  Total & 385 & 545 & 345 & 545 \\ \hline
    \hline
\end{tabular}
\caption{\textbf{Summary statistics for the coherency-and-label
validation exercise with base model (legislative bills).}
This table shows the number of correct tasks and the number of total tasks
for each model and for each topic.}
\end{table}

\begin{table}[!ht]
\centering
\begin{tabular}{lrrrr}
  \hline
  \hline & \multicolumn{2}{c}{\keyATM} & \multicolumn{2}{c}{\STM}\\ \hline
  Topic &  \# of correct tasks & \# of tasks & \# of correct tasks & \# of tasks\\ \hline
Alternation of government &   9 &  30 &   7 &  23 \\
  Constitution &   4 &  21 &  14 &  18 \\
  Economic recovery &  14 &  26 &  12 &  27 \\
  Education &  10 &  23 &  11 &  21 \\
  Environment &  13 &  23 &  12 &  25 \\
  Inclusive society &  12 &  32 &   6 &  25 \\
  Party &  10 &  31 &   5 &  19 \\
  Pension &   6 &  31 &  12 &  34 \\
  Postal privatization &  11 &  32 &  18 &  35 \\
  Public works &   8 &  19 &   7 &  18 \\
  Regional devolution &  13 &  26 &   9 &  28 \\
  Road construction &  19 &  26 &  14 &  27 \\
  Security &   8 &  17 &   7 &  18 \\
  Social welfare &   5 &  22 &  11 &  28 \\
  Tax &   6 &  19 &   7 &  25 \\
  Trade &  13 &  22 &  11 &  29 \\  \hline
  Total & 161 & 400 & 163 & 400 \\
    \hline
\end{tabular}
\caption{\textbf{Summary statistics for the coherency
validation exercise with covariate model (manifestos).}
This table shows the number of correct tasks and the number of total tasks
for each model and for each topic.}
\end{table}

\begin{table}[!ht]
\centering
\begin{tabular}{lrrrr}
  \hline
  \hline & \multicolumn{2}{c}{\keyATM} & \multicolumn{2}{c}{\STM}\\ \hline
  Topic &  \# of correct tasks & \# of tasks & \# of correct tasks & \# of tasks\\ \hline
Alternation of government &  11 &  31 &  12 &  20 \\
  Constitution &  10 &  28 &   7 &  25 \\
  Economic recovery &  15 &  30 &  16 &  33 \\
  Education &  11 &  23 &  16 &  28 \\
  Environment &   6 &  18 &  11 &  27 \\
  Inclusive society &  10 &  18 &   8 &  23 \\
  Party &  16 &  39 &  12 &  29 \\
  Pension &  12 &  22 &   9 &  20 \\
  Postal privatization &  18 &  29 &  10 &  37 \\
  Public works &   7 &  26 &   7 &  22 \\
  Regional devolution &  15 &  26 &  13 &  26 \\
  Road construction &  26 &  33 &   5 &  19 \\
  Security &  25 &  28 &  13 &  30 \\
  Social welfare &  11 &  27 &   8 &  20 \\
  Tax &   4 &  24 &   8 &  30 \\
  Trade &  10 &  18 &  12 &  31 \\ \hline
  Total & 207 & 420 & 167 & 420 \\
    \hline
\end{tabular}\caption{\textbf{Summary statistics for the coherency-and-label
validation exercise with covariate model (manifestos).}
This table shows the number of correct tasks and the number of total tasks
for each model and for each topic.}
\end{table}

\begin{table}[!ht]
\centering
\begin{tabular}{lrrrr}
  \hline
  \hline & \multicolumn{2}{c}{\keyATM} & \multicolumn{2}{c}{\wLDA}\\ \hline Topic &  \# of correct tasks & \# of tasks & \# of correct tasks & \# of tasks\\ \hline
Attorneys &  16 &  44 &  15 &  44 \\
  Civil rights &  19 &  39 &  24 &  39 \\
  Criminal procedure &  47 &  53 &  44 &  53 \\
  Due process &  20 &  40 &  31 &  40 \\
  Economic activity &  28 &  47 &  32 &  47 \\
  Federal taxation &  28 &  39 &  34 &  39 \\
  Federalism &  42 &  54 &  40 &  54 \\
  First amendment &  24 &  42 &  24 &  42 \\
  Interstate relations &  31 &  37 &  32 &  37 \\
  Judicial power &  34 &  48 &  38 &  48 \\
  Miscellaneous &  34 &  41 &  27 &  41 \\
  Privacy &  21 &  30 &  20 &  30 \\
  Private action &  46 &  63 &  47 &  63 \\
  Unions &  44 &  48 &  45 &  48 \\ \hline
  Total & 434 & 625 & 453 & 625 \\ \hline
    \hline
\end{tabular}
\caption{\textbf{Summary statistics for the coherency
validation exercise with dynamic model (court opinions).}
This table shows the number of correct tasks and the number of total tasks
for each model and for each topic.}
\end{table}

\begin{table}[!ht]
\centering
\begin{tabular}{lrrrr}
  \hline
  \hline & \multicolumn{2}{c}{\keyATM} & \multicolumn{2}{c}{\wLDA}\\ \hline Topic &  \# of correct tasks & \# of tasks & \# of correct tasks & \# of tasks\\ \hline
Attorneys &  10 &  32 &  13 &  32 \\
  Civil rights &  15 &  35 &  20 &  35 \\
  Criminal procedure &  29 &  33 &  27 &  33 \\
  Due process &   9 &  31 &  28 &  31 \\
  Economic activity &  14 &  46 &  36 &  46 \\
  Federal taxation &  38 &  48 &  39 &  48 \\
  Federalism &  22 &  46 &  15 &  46 \\
  First amendment &  37 &  42 &  27 &  42 \\
  Interstate relations &  14 &  35 &  21 &  35 \\
  Judicial power &  30 &  40 &  24 &  40 \\
  Miscellaneous &  27 &  51 &  21 &  51 \\
  Privacy &  22 &  37 &  16 &  37 \\
  Private action &  17 &  33 &  14 &  33 \\
  Unions &  38 &  41 &  37 &  41 \\ \hline
  Total & 322 & 550 & 338 & 550 \\ \hline
    \hline
\end{tabular}
\caption{\textbf{Summary statistics for the coherency-and-label
validation exercise with dynamic model (court opinions).}
This table shows the number of correct tasks and the number of total tasks
for each model and for each topic.}
\end{table}

\subsection{Topic-by-topic result}
\label{app:validation-topic-by-topic}

\begin{figure}[!h]
    \centering (a) Base Model: Legislative Bills\par\medskip
    \includegraphics[width= \linewidth]{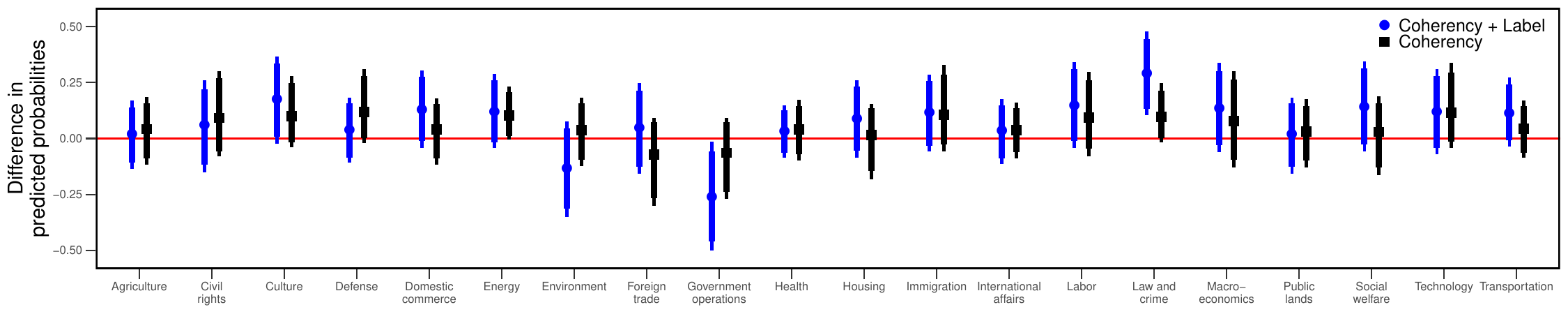}
    \centering (b) Covariate Model: Manifestos\par\medskip
    \includegraphics[width= \linewidth]{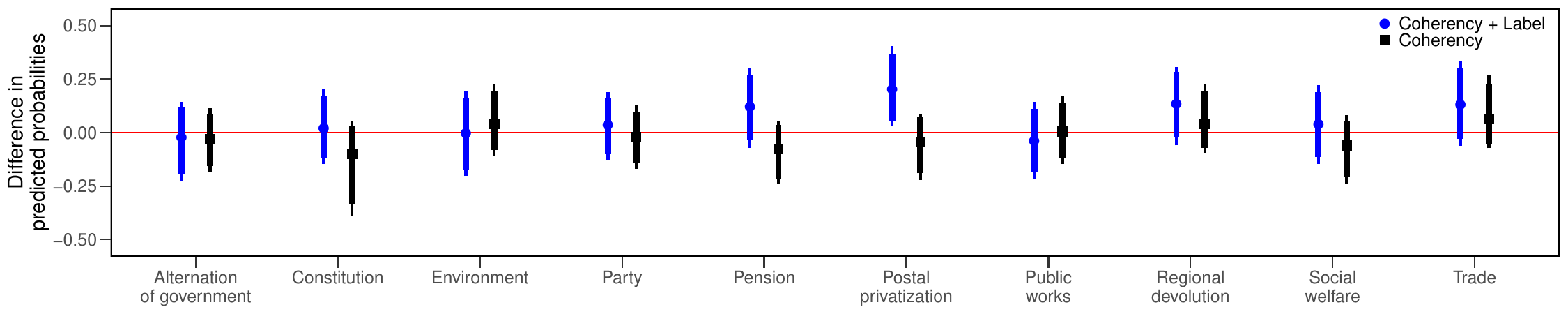}
    \centering (c) Dynamic Model: Court Opinions\par\medskip
    \includegraphics[width= \linewidth]{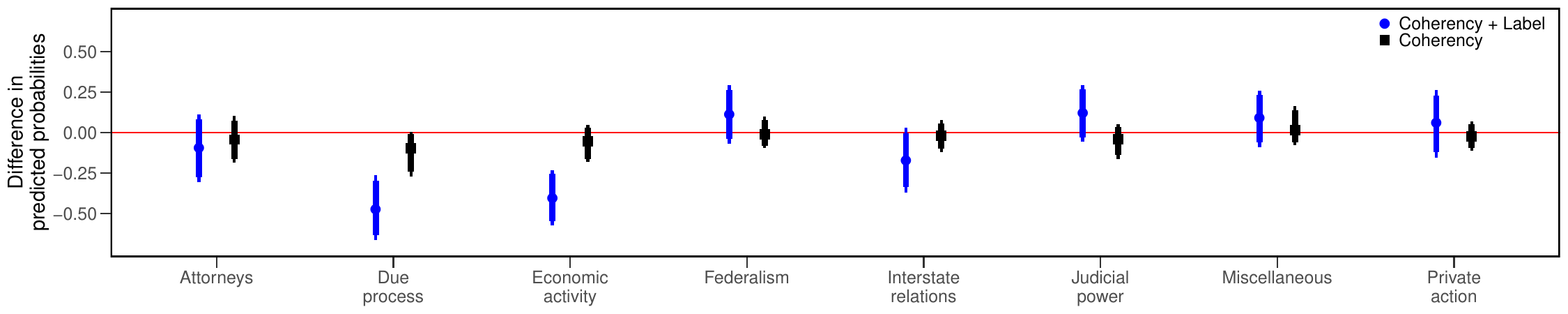}
    \caption{\textbf{Comparison of the performance of validation results
      between \keyATM{} and its baseline counterparts.}
      Each point represents the difference in the predicted probabilities
      derived from the three different regression.
      The thick and thin vertical lines indicate the 95\% and 90\% credible intervals respectively.
      Black lines and blue lines show results from the coherency task and
      the coherency-and-label task respectively.}
    \label{app:fig:validation-appendix}
\end{figure}

Figure~\ref{app:fig:validation-appendix} shows the topic-by-topic
results.  \keyATM{} performs better than \wLDA{} in most
  topics.  The performance of \keyATM{} is worse than \wLDA{} for some
  topics in the Supreme Court opinion application. The quality of
  these topics is lower for \keyATM{} than \wLDA{}. For example,
  \textit{Economic activity} does not perform well in \keyATM{} and
  the AUROC curve for this topic is lower for \keyATM{}
  than \wLDA{} (Figure~\ref{app:fig:dynamic-roc}). Also, the top ten
  frequent words for these topics are less interpretable for
  \keyATM. As shown in Table~\ref{tab:app-dynamic-topwords}, \wLDA{}
  includes more meaningful words in \textit{Economic activity} than
  \keyATM{}.

\end{appendices}

\end{document}